\documentclass[10pt,twocolumn,letterpaper]{article}

\usepackage{iccv}              

\newcommand{\myparagraph}[1]{\paragraph{#1}}

\newcommand{\themethod}[0]{\textsc{GaSLight}\xspace}

\definecolor{iccvblue}{rgb}{0.21,0.49,0.74}
\usepackage[pagebackref,breaklinks,colorlinks,allcolors=iccvblue]{hyperref}

\usepackage{nicefrac}
\usepackage{array}
\usepackage{color}
\usepackage{pifont}
\usepackage{caption}
\usepackage{float}
\usepackage{relsize}
\usepackage{siunitx}
\usepackage{graphicx}
\usepackage{multirow}
\usepackage{rotating}
\usepackage{soul}

\newcommand{\rot}[1]{\begin{turn}{90}#1\enspace\end{turn}}

\usepackage{xcolor}
\newcommand{\modif}[1]{#1}

\usepackage{colortbl}
\def\paperID{14636} 
\def\confName{ICCV}
\def\confYear{2025}

\title{\themethod: Gaussian Splats for Spatially-Varying Lighting in HDR}

\author{Christophe Bolduc$^1$ \quad
Yannick Hold-Geoffroy$^2$ \quad
Zhixin Shu$^2$ \quad
Jean-François Lalonde$^1$ \\
$^1$Université Laval, $^2$Adobe
}

\begin{document}

\makeatletter
\g@addto@macro\@maketitle{
  \vspace{-2em}
  \begin{figure}[H]
  \setlength{\linewidth}{\textwidth}
  \setlength{\hsize}{\textwidth}
  \centering
    \footnotesize
    \begin{tabular}{@{}c@{}c@{}c@{}c@{}}
    \includegraphics[width=0.25\linewidth]{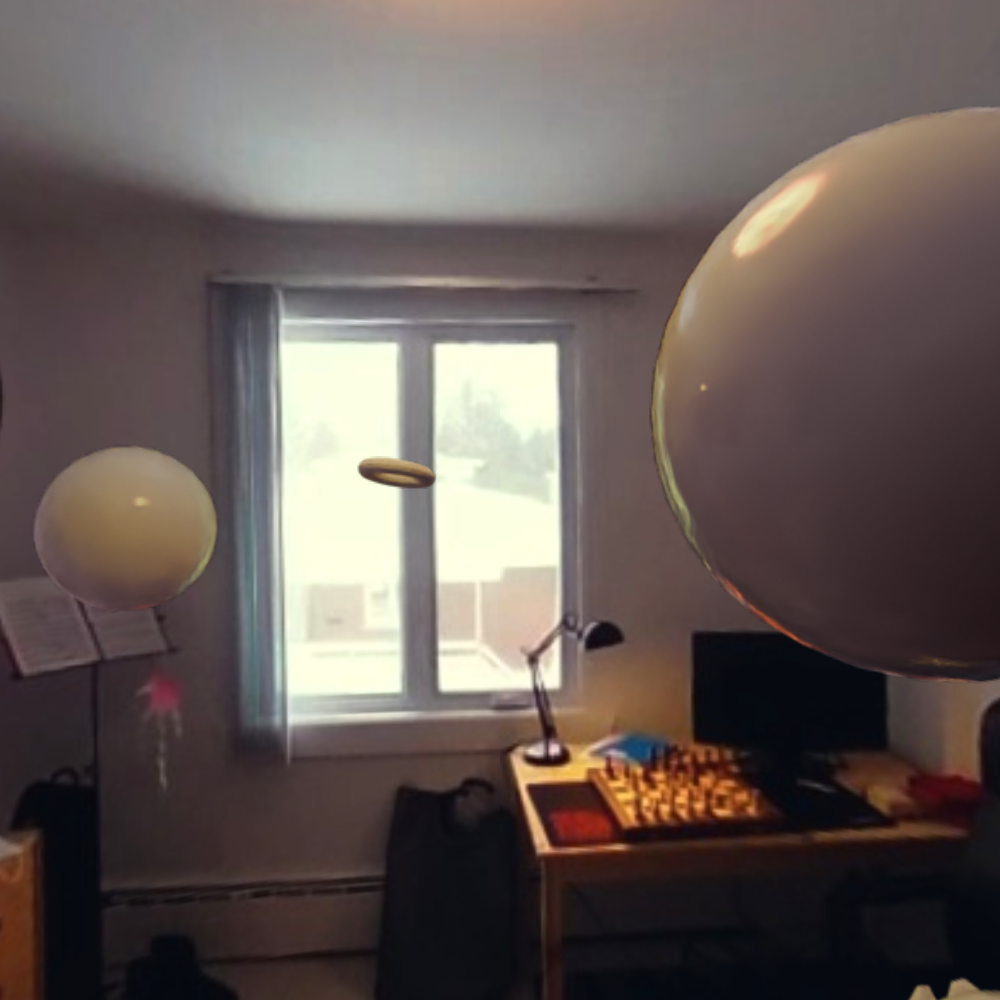} &
    \includegraphics[width=0.25\linewidth]{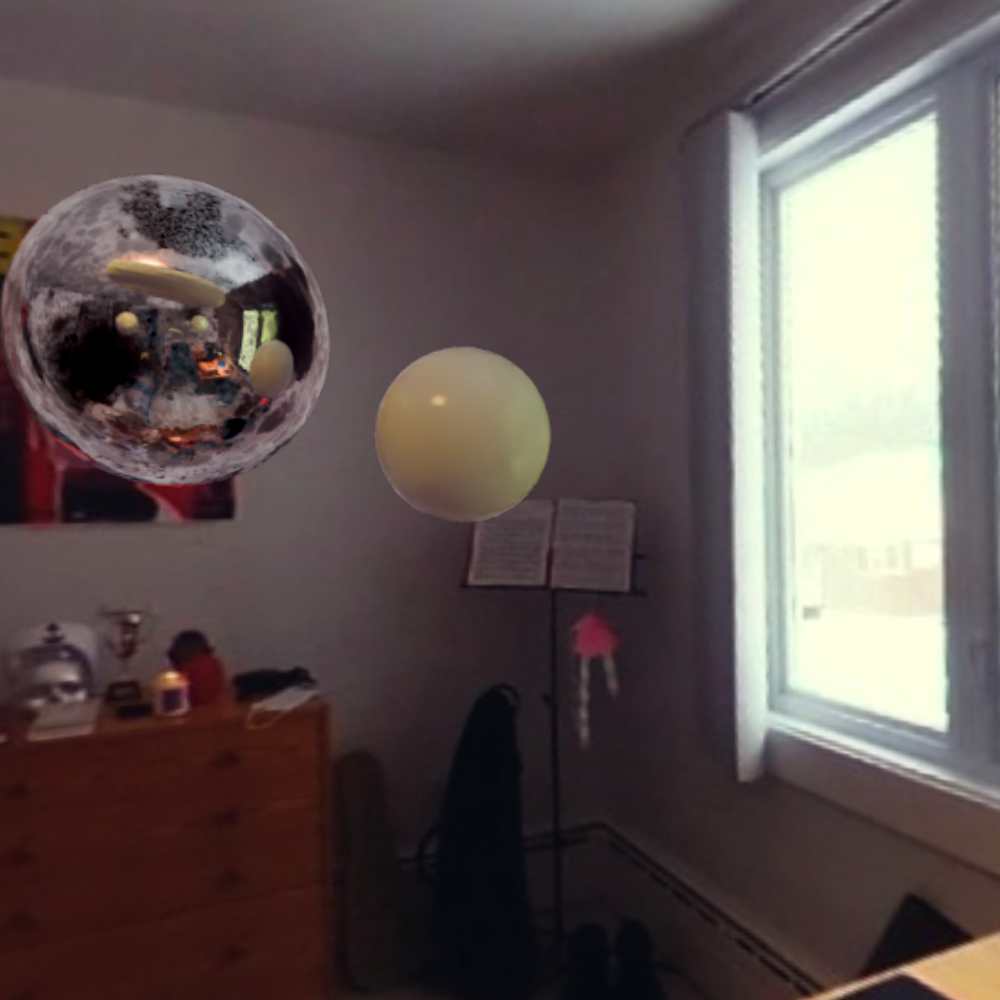} &
    \includegraphics[width=0.25\linewidth]{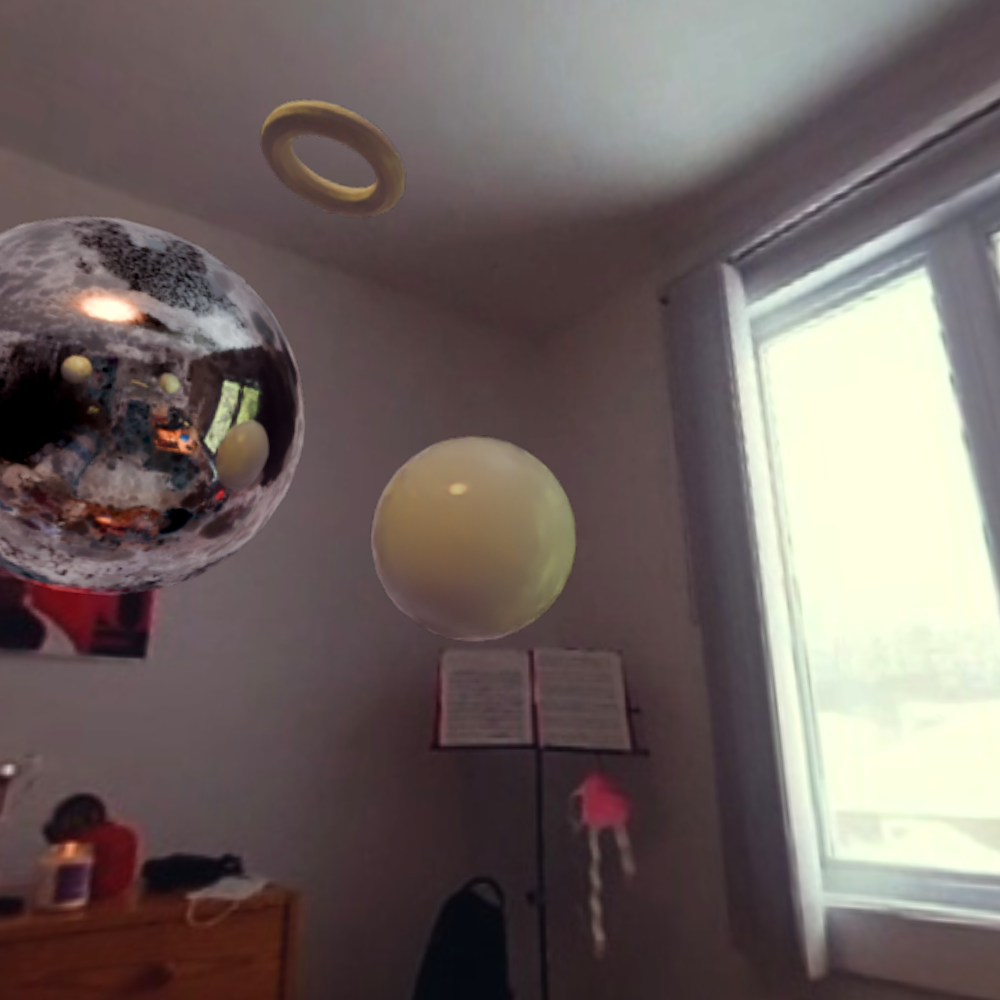} &
    \includegraphics[width=0.25\linewidth]{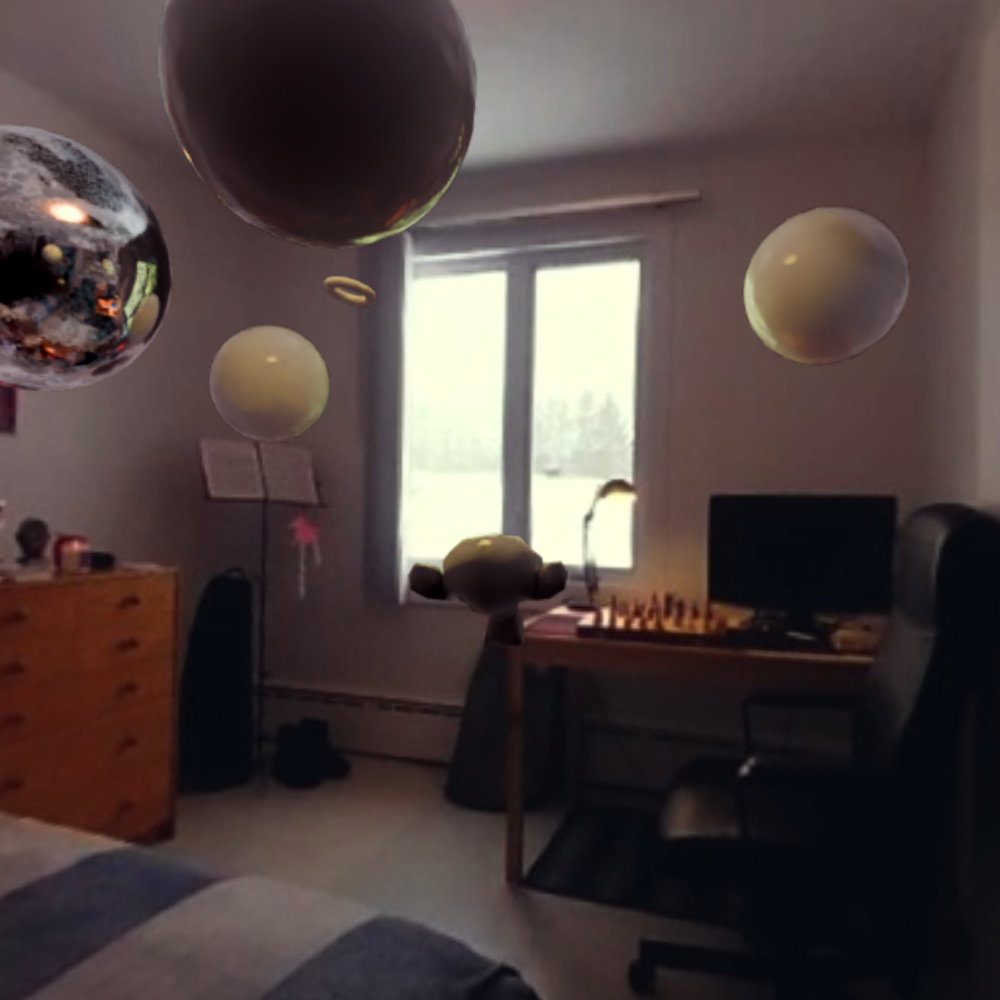} \\
    \end{tabular}
  \caption{We introduce \themethod{}, a framework that captures environment lighting as HDR Gaussian Splats. From a sequence of regular photographs of a scene (backgrounds), \themethod{} generates plausible illumination and can readily be applied as lighting in a 3D renderer to composite virtual objects onto images. Our representation inherently provides both high-frequency texture for reflections, as shown in the mirror spheres, and plausible light intensity, as seen in the shading of the inserted spheres, tori, and Blender's Suzanne.}
  \label{fig:teaser}
  \end{figure}
}
\makeatother

\maketitle

\begin{abstract}
We present \themethod, a method that generates spatially-varying lighting from regular images. Our method proposes using HDR Gaussian Splats as light source representation, marking the first time regular images can serve as light sources in a 3D renderer. Our two-stage process first enhances the dynamic range of images plausibly and accurately by leveraging the priors embedded in diffusion models. Next, we employ Gaussian Splats to model 3D lighting, achieving spatially variant lighting. Our approach yields state-of-the-art results on HDR estimations and their applications in illuminating virtual objects and scenes. To facilitate the benchmarking of images as light sources, we introduce a novel dataset of calibrated and unsaturated HDR captures to evaluate images as light sources. We assess our method using a combination of our dataset and an existing dataset from the literature. The code to reproduce our method is available at {\small \url{https://lvsn.github.io/gaslight/}}.
\end{abstract}

\section{Introduction}

Lighting is crucial not only for shaping an image, but also for ensuring realism and aesthetic appeal when inserting virtual content. Estimating accurate lighting representations from one or many photographs is vital for creating lifelike visuals that convincingly blend within their surroundings and convey realistic surface appearance. This is essential in numerous applications including multimedia productions, virtual reality, forensics, cultural displays, and more. 

Despite its importance, current methods face significant limitations in capturing and representing lighting accurately. One reason for this is the difficulty in capturing high dynamic range (HDR), which requires multiple exposures~\cite{debevec1997recovering} or specialized apparatus~\cite{stumpfel2006direct,nayar2000high}. As a result, most approaches \cite{yu2021luminance_lanet,liu2020singlehdr,eilertsen2017hdrcnn} attempt to convert regular, low-dynamic range (LDR) images to HDR. However, as will be demonstrated later, they primarily focus on extending the dynamic range for tonemapping and visualization purposes \cite{wang2024lediff,gharbi2017deep,hasinoff2016burst}, rather than extrapolating the intensity of bright light sources. 

Another reason is the difficulty of simultaneously estimating HDR lighting that is both spatially-varying and high frequency.
\Cref{tab:lighting_representation_comparison} provides an overview of existing lighting representations that have been proposed in the literature. 
Existing approaches either focus on accurate energy estimation, using parametric light sources~\cite{gardner2019deepparametric}, which allow for near-field effects, such as positioning a discrete lamp in the scene. Alternatively, per-pixel spherical harmonics~\cite{garon2019fast} tend to struggle in modeling high-frequency lighting and cannot generate strong cast shadows. Volumetric representations~\cite{li2020inverse,srinivasan2020lighthouse,wang2021learning,li2023spatiotemporally} do not accurately model high-frequency lighting variations, as needed for reflections on glossy surfaces, and do not readily integrate within existing rendering engines. Another prevalent representation used by lighting estimation methods is the environment map \cite{gardner2017learning,song2019neural,somanath2021hdr,dastjerdi2023everlight,phongthawee2024diffusionlight} (commonly referred to as IBL, for image-based lighting), which models infinitely-distant lighting as a spherical image. This representation cannot yield near-field lighting effects, such as a lamp in the scene, and model lighting as an infinitely-distant sphere. These limitations reduce the versatility of current lighting estimation methods, affecting their ability to achieve realistic lighting in scenes. 

\begin{table}[t]
    \centering
    \footnotesize
    \newcommand{\badrep}{\textcolor{red}{\ding{55}}}
    \newcommand{\goodrep}{\textcolor{green}{\ding{51}}}
    \begin{tabular}{@{}l@{}c@{}c@{}c@{}c@{}c@{}}
        & Param. & SH & Vol. & IBL & Ours \\
        \hline
        Near-field & \goodrep & \badrep & \goodrep & \badrep & \goodrep \\
        Spatially-varying & \goodrep & \goodrep & \goodrep & \badrep & \goodrep \\
        HF reflections & \badrep & \badrep & \badrep & \goodrep & \goodrep \\
        HF lighting & \goodrep & \badrep & \goodrep & \goodrep & \goodrep \\
        Easy & \goodrep & \goodrep & \badrep & \goodrep & \goodrep \\
        Schematic &
        \includegraphics[width=1.3cm]{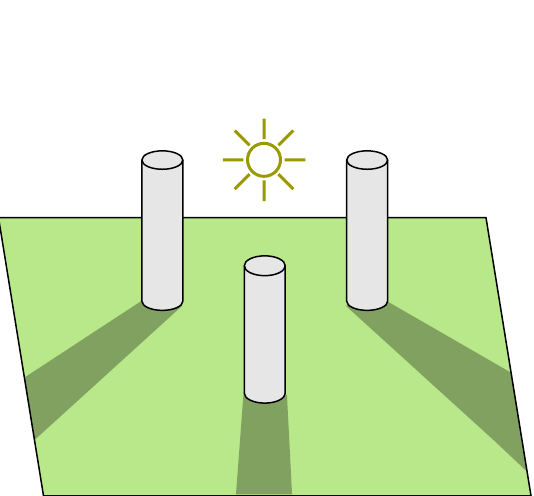} &
        \includegraphics[width=1.3cm]{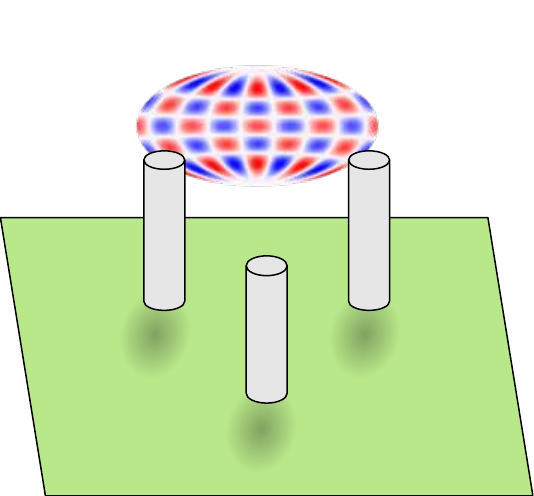} &
        \includegraphics[width=1.3cm]{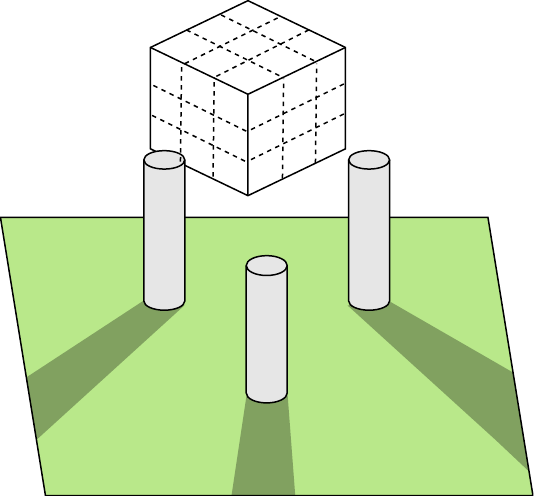} &
        \includegraphics[width=1.3cm]{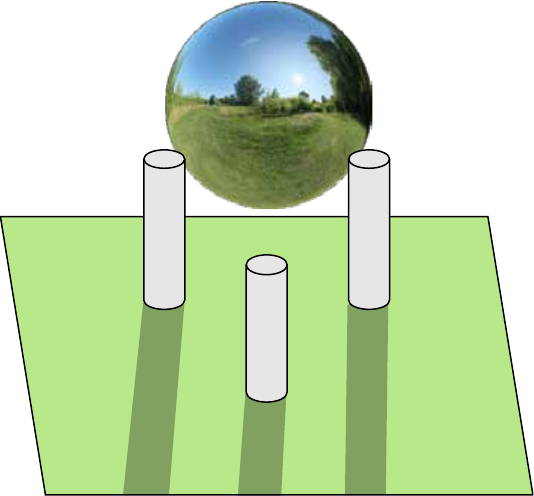} &
        \includegraphics[width=1.3cm]{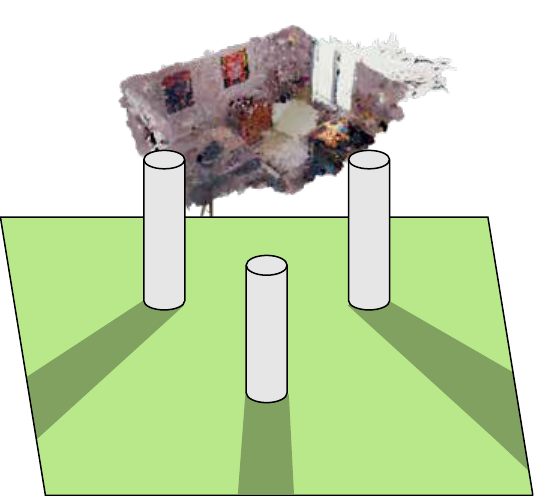}\\
    \end{tabular}
    \caption{Comparison of different lighting representations. We include parametric (point or directional light sources, ``param.''), Spherical Harmonics (``SH''), volumetric (``Vol.''), environment maps (``IBL''), and our proposed HDR GS representation. Aspects compared are near-field effects, such as the ability to represent a light source \emph{within} the scene, High-Frequency reflections allowing for high-resolution details in reflections, High-Frequency lighting allowing strong shadows, and its overall ease of use (\eg, integration in existing rendering engines).}
    \label{tab:lighting_representation_comparison}
\end{table}

In this work, we introduce \themethod, a novel method that addresses these challenges by capturing the high-frequency and spatially-varying lighting of a scene. \themethod employs HDR Gaussian Splats (GS) as light representation, which are reconstructed from a set of LDR images. We argue that GS are particularly well-suited to representing lighting, as they allow continuous, spatially-varying intensity representation with native support for high-frequency detail and efficient integration in renderers. As opposed to HDRGS~\cite{cai2025hdrgs} (or HDR-NeRF~\cite{huang2022hdr}), our approach does not require alternating exposures to recover the HDR---rather, we rely on a custom, specially-designed LDR-to-HDR network based on powerful diffusion models. Compared to previous NeRF-based HDR estimation approaches~\cite{gera2022panohdrnerf}, our method is more accurate, does not require per-scene finetuning, and is much faster. 

In summary, our contributions are threefold. First, we present a framework for using Gaussian Splats as a spatially-varying HDR light representation that can be estimated from a few images of a scene. This light representation can readily be used in a 3D renderer, enabling accurate and versatile lighting effects. Second, at the heart of our framework is a novel LDR-to-HDR estimation model, which leverages the strong priors in diffusion models to plausibly estimate the intensity of light sources. Finally, we compare our method with existing HDR datasets from the literature and demonstrate that it best predicts the intensity of bright light sources. We also demonstrate its applicability for virtual object insertion, where virtual objects can be inserted at multiple locations in the scene and relit appropriately in a multi-view setting.

\section{Related work}

\myparagraph{Inverse tonemapping}
has been an active area of research for several decades, with early methods focusing on inverting the camera function through various ad-hoc extrapolation techniques \cite{hayden2002production,banterle2006inverse}. 
Later, this topic was revisited using deep learning, first using end-to-end approaches \cite{eilertsen2017hdrcnn,endo2017drtmo,marnerides2018expandnet}, then by explicitly inverting the camera pipeline~\cite{liu2020singlehdr}. Inverse tonemapping has also been applied to spherical 360\textdegree{} panoramas, both outdoors~\cite{zhang2017learning} and indoors~\cite{yu2021luminance_lanet}. 

More recently, generative imaging has been utilized in inverse tonemapping, with GlowGAN~\cite{wang2023glowgan} being the first to recover multi-exposure bracketing, akin to traditional HDR capture~\cite{debevec1997recovering}. Exposure Diffusion~\cite{bemana2024exposurediffusion} builds on this by incorporating exposure constraints into a pretrained diffusion model to generate relative exposures, although it is limited to pixel-space diffusion. Subsequently, LeDiff~\cite{wang2024lediff} finetunes a latent diffusion model~\cite{rombach2022high} to determine relative exposures and merge them into an extended dynamic range image. This method shows promising results with latent-space diffusion, but remains constrained by the decoder's limited dynamic range, which does not fully represent the richness of natural lighting. 
Our method proposes an iterative dynamic range expansion scheme inspired by diffusion-based HDR estimation, enabling a significantly larger increase in dynamic range than current methods allow.

\myparagraph{3D scene representations}
Traditional methods often relied on textured 3D meshes \cite{foley1996computer} as the primary means of representing scene geometry and appearance. 
Implicit radiance functions, such as those introduced in Neural Radiance Fields (NeRFs)~\cite{mildenhall2021nerf}, have revolutionized 3D scene representations by leveraging neural networks to represent detailed geometries and appearances, successfully modeling complex, view-dependent effects. HDR-NeRF~\cite{huang2022hdr}, HDR-HexPlane~\cite{wu2024fast}, and others \cite{jun2022hdrplenoxels,mildenhall2022nerf,wang2024bilateral} have extended this framework to handle HDR content, allowing for the reconstruction of scenes with an extended dynamic range. PanoHDR-NeRF~\cite{gera2022panohdrnerf} combines an LDR-to-HDR network trained for 360\textdegree{} panoramas with an omnidirectional NeRF representation to capture the spatially-varying HDR light field of an indoor scene. 

Later, explicit radiance functions based on 3D Gaussian Splatting~\cite{kerbl2023gaussiansplatting} were proposed. 
HDR-GS~\cite{cai2025hdrgs} and CineGR~\cite{wang2024cinematic}, leveraged multi-exposure input photographs to capture an extended dynamic range, similarly to their predecessor HDR-NeRF. 
In contrast to these methods, we focus on the extrapolation of bright light sources present in the scene, doing so without the requirement of multi-exposure inputs. We therefore create an explicit GS-based HDR representation, allowing them to function effectively as light sources for 3D renderers.



%

\myparagraph{Lighting representations}
have evolved significantly over time, each addressing different aspects of light modeling and scene illumination. Parametric lights, such as point lights or directional lights, have been foundational since the inception of computer graphics. Methods that learn to predict parametric lights from images have been proposed \cite{gardner2019deepparametric,zhan2021emlight,griffiths2022outcast}, providing intuitive control over lighting, but often struggle to accurately represent complex environments. 

Spherical harmonics offer an efficient representation for lighting in real-time applications. They have been used to model both irradiance \cite{ramamoorthi2001efficient,green2003spherical} and radiance \cite{sloan2023precomputed}. Garon et al.~\cite{garon2019fast} estimate spatially-varying spherical harmonics directly from a single image. \modif{Similarly, spherical Gaussians are an efficient basis to produce low dimensionality environment lighting~\cite{physg2021,sf3d2024}. }

Environment maps \cite{blinn1976texture} provide a non-parametric alternative for capturing complex lighting environments, as they represent an infinitely distant light source modeled as a spherical image. They have been employed in many lighting estimation methods \cite{gardner2017learning,song2019neural,hold2019deep,somanath2021hdr,gera2022panohdrnerf,dastjerdi2023everlight,phongthawee2024diffusionlight}. Implicit volumetric lighting representations were also devised to model both indoor \cite{srinivasan2020lighthouse,li2023spatiotemporally} and outdoor \cite{wang2023neural} lights, or for inverse rendering purposes \cite{ling2024nerf} \modif{\cite{Wu_2023_CVPR, boss2021neuralpil}}. Their opaqueness thwarts their powerful modeling capabilities, making them challenging to interpret and edit. \modif{Explicit spatial light representations have also been introduced ~\cite{du2024gsidilluminationdecompositiongaussian}, but the number of sources is limited and used only to represent indirect lighting.} In contrast, our method uses Gaussian splats as lighting representation, which combines spatially-varying lighting capabilities while maintaining the flexibility and interpretability of parametric representations.





\section{Method}

Our method takes a set of regular LDR images as input and produces lighting as an HDR 3D Gaussian Splats (3DGS) scene. The 3DGS can then be queried to obtain the amount of incoming radiance from a specific direction for rendering purposes. 
We break down our method into two steps. First, we uplift the LDR image to unsaturated HDR, providing plausible intensities for the observed light sources. Then, we fit the 3DGS scene to the estimated HDR content. 
It is important to note that our method does not perform lighting estimation from unseen parts of the scene---rather, it models only the lights visible within at least one image. 
We first describe our per-image light intensity estimation method, followed by the 3D scene reconstruction we employ. 

\subsection{HDR light intensity estimation}
\label{sec:light_intensity_estimation}

\begin{figure}[t]
    \centering
    \footnotesize
      \includegraphics[width=\linewidth]{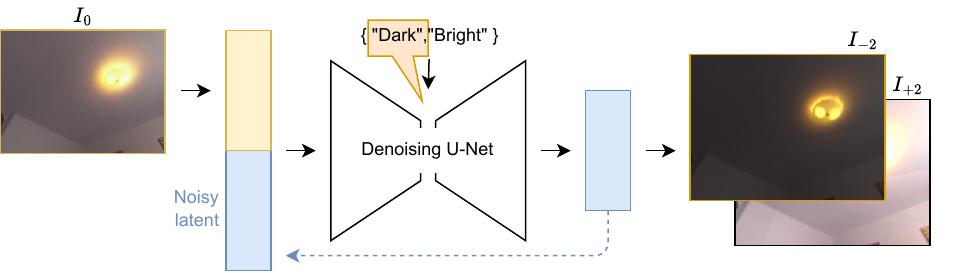} \\
      \vspace{-4pt}(a) training \vspace{4pt} \\
      \includegraphics[width=\linewidth]{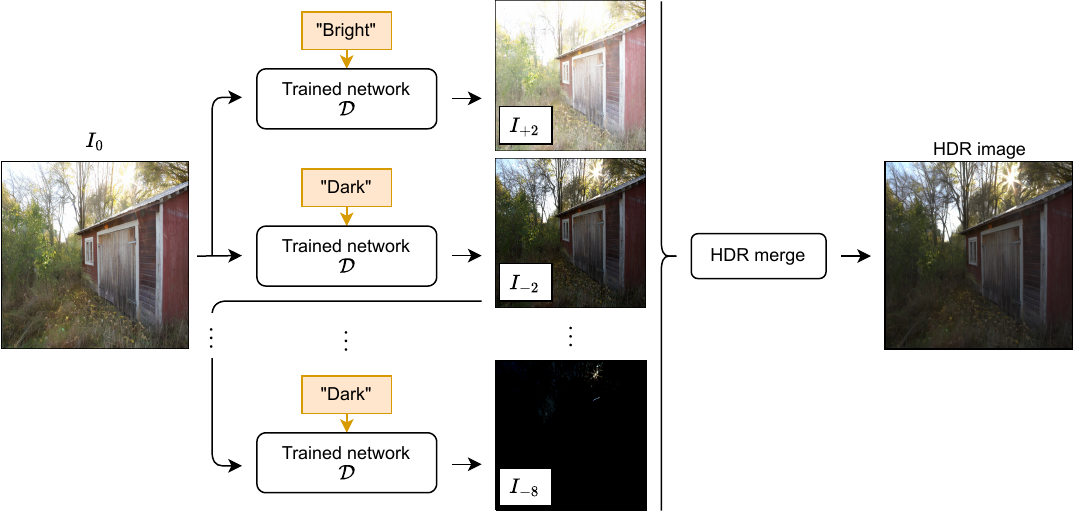} \\
      (b) inference \\
    \caption{Overview of our HDR estimation pipeline. (a) We train a diffusion model to predict two exposures (brighter or darker) given an input image. (b) At inference time, we recursively apply the same network to obtain an exposure stack, which is subsequently merged to HDR.}
    \label{fig:pipeline_HDR}
\end{figure}

Our HDR estimation method aims to produce a plausible hallucination of the intensities of the bright light sources present in an LDR image. As illustrated in \cref{fig:pipeline_HDR}, our approach relies on a diffusion model $\mathcal{D}$, which accepts as input an image at a certain exposure $I_0$, and is trained (\cref{fig:pipeline_HDR}a) to predict two versions of the input image at different exposures: $I_{-2} = \mathcal{D}_{-}(I_0)$, and $I_{+2} = \mathcal{D}_+(I_0)$. Here, the $I_X$ notation means applying a $2^X$ factor to the exposure of the input image $I_0$. The resulting exposures (\cref{fig:pipeline_HDR}b) are then merged into a single HDR image using a conventional HDR merging algorithm~\cite{debevec1997recovering}.

\myparagraph{Multi-exposure prediction network}
The backbone of our light intensity estimation is ICLight \cite{zhang2025scaling}, a diffusion-based model fine-tuned for relighting, since it demonstrated strong performance in understanding lighting conditions. We further fine-tune it for HDR estimation as shown in \cref{fig:pipeline_HDR}a.
To do so, the input image is encoded into the latent space and concatenated with the initial noise. As in RGB$\leftrightarrow$X~\cite{zeng2024rgb}, we employ text-based conditioning to indicate whether the network should output a darker $I_{-2} = \mathcal{D}_{-}(I_0)$ (conditioning text: ``dark'') or brighter $I_{+2} = \mathcal{D}_+(I_0)$ (``bright'') exposure. The network accepts an LDR image in sRGB space (with $\gamma=2.2$~\cite{eilertsen2017hdrcnn}), and is trained to produce a re-exposed version of the input, also in sRGB space. Exposure $I_0$ is therefore simply the input image.

At inference time, the network can be applied recursively to the predicted exposures to further expand the dynamic range (\ie, $I_{+4} = \mathcal{D}_+(\mathcal{D}_+(I_0))$ as illustrated in \cref{fig:pipeline_HDR}b. In practice, we repeat this process until images no longer contain (under- or over-)saturated pixels\modif{, which takes on average roughly 5 iterations}.

\myparagraph{HDR merging}
We generate HDR images from exposure-bracketed sequences $\{I_{-2n}, ..., I_{+2m}\}$ 
\cite{reinhard2020high}, where $n$ and $m$ represent the number of dark and bright exposures estimated by the network, respectively. We first linearize the images by assuming a gamma $\gamma\!=\!2.2$ tonemapping, followed by the Debevec HDR fusion method \cite{debevec1997recovering} to combine the sequence into a single HDR image. To preserve unsaturated pixels from $I_0$ and reduce blurring caused by averaging diffusion model outputs across exposures, we assign weight $1$ to pixels with intensity $\in \! [0.2, 0.8]$. Remaining pixels use the standard weighting function, $w \! = \! -2 \left\lvert z \! - \! 0.5 \right\rvert \! + \! 1$, producing a linear unbounded color space.

\myparagraph{360\textdegree{} prediction}
Our method can also be applied to 360\textdegree{} panoramas by using a cube map representation and processing each face independently. While this does not explicitly ensure consistency for light sources crossing the edges of the cube's face, we observe that results are usable and produce plausible lighting and reflections, effectively converting 360\textdegree{} into HDRI maps.

\myparagraph{Training data and implementation details}
We employ a dataset of 12,611 HDR 360\textdegree{} panoramas to train our light intensity estimation model, containing a mix of publicly available \cite{polyhaven,bolduc2023beyond} and captured data. At training time, we project the spherical panorama to an image plane, using random parameters drawn according to the distributions shown in \cref{tab:parameters-sampling}. Note that images are first re-exposed using \cite{reinhard2002photographic} such that the of the log-luma channel is 0.17 prior to applying the exposure augmentation from \cref{tab:parameters-sampling}. 
We also randomly apply a tone-mapping operator between gamma, Filmic, or Hybrid-Log-Gamma to ensure the model is robust to various camera functions. We normalize the output of the tonemapping operators to $[\text{-}1, 1]$ to fit Stable Diffusion's original training range. We obtain the prompts from each image using LLaVA~\cite{liu2023llava}.

\begin{table}[t]
\centering
\footnotesize
\begin{tabular}{ll}
\toprule
Parameter & Distribution \\
\midrule
Azimuth & $\mathcal{U}[0, 2\pi]$ \\
Elevation & $\mathcal{U}[-\nicefrac{\pi}{2}, \nicefrac{\pi}{2}]$ \\
Roll & $\mathcal{U}[-\nicefrac{\pi}{16}, \nicefrac{\pi}{16}]$ \\
\midrule
Vert. FoV & $\mathcal{U}[60^\circ, 120^\circ]$ \\
\midrule
Exposure & $2^{e}, e \sim \mathcal{U}[-8,4]$\\
\bottomrule
\end{tabular}
\caption{Sampling parameters to generate training data for our HDR light intensity estimation network. Here, $\mathcal{U}[a,b]$ denotes a uniform distribution in the $[a,b]$ interval.}
\label{tab:parameters-sampling}
\end{table}

Our captured data was acquired in static scenes using a Ricoh Theta Z1 with exposure bracketing of 10 captures spanning from 1s to 1/25000s to obtain unsaturated captures. The camera was set to $f/2.1$, ISO 80, and fixed 5000K white balance for all captures. We stitched the RAW stereo spherical captures using PTGui \cite{ptgui}. 

We initialize our Stable Diffusion 1.5 backbone \cite{rombach2022high} with the IC-Light \cite{zhang2025scaling} checkpoint and fine-tune it for 415k iterations. We employ an initial learning rate of $10^{\text{-}5}$. \modif{Training took 10 days using 8 A100-40GB GPUs. At inference, we perform 20 diffusion steps, taking $\sim{}\!4$ seconds per relative EV on an Nvidia Titan V GPU. }

\subsection{Gaussian Splat lighting representation}

\begin{figure*}[t]
    \centering
    \includegraphics[width=0.75\linewidth]{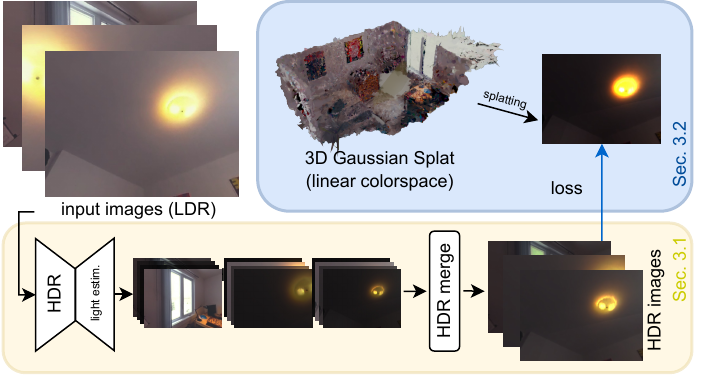}
    \caption{Overview of our 3D reconstruction pipeline. From a collection of LDR input images, we first convert them to HDR by using our HDR light intensity estimation network (bottom, see \cref{sec:light_intensity_estimation}). A 3D Gaussian Splatting (3DGS) representation is subsequently trained on the resulting HDR images (top). }
    \label{fig:pipeline_gs}
\end{figure*}

We utilize the 3D Gaussian Splatting framework (3DGS) 
\cite{kerbl2023gaussiansplatting} as our HDR lighting representation. This representation offers multiple benefits, establishing it as a superior option over other lighting representations, as detailed in \cref{tab:lighting_representation_comparison}. 

As illustrated in \cref{fig:pipeline_gs}, we first process each image in the set with our HDR extrapolation network (c.f. \cref{sec:light_intensity_estimation}) to obtain (linear) HDR images. Then, we apply the training process proposed by InstantSplat~\cite{fan2024instantsplat} to recover an HDR 3DGS representation of the scene (in linear space). \modif{We optimize the 3DGS using an $\ell_1$ loss, along with a SSIM loss on LDR pixels. }

\section{Evaluation}

We first discuss the evaluation datasets, then compare our light source intensity estimation model to existing LDR-to-HDR methods. Finally, we show that our 3D lighting framework better represents natural scene lighting conditions than prior work. 

\subsection{Data}

\begin{figure}
    \centering
    \footnotesize
    \begin{tabular}{cc}
    \includegraphics[width=0.3\linewidth]{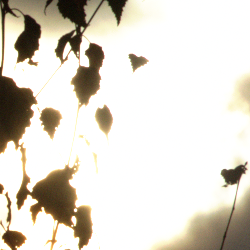} &
    \includegraphics[width=0.3\linewidth]{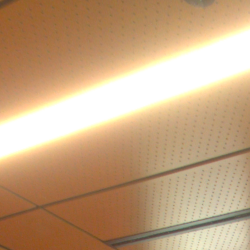} \\ 
    \includegraphics[width=0.4\linewidth]{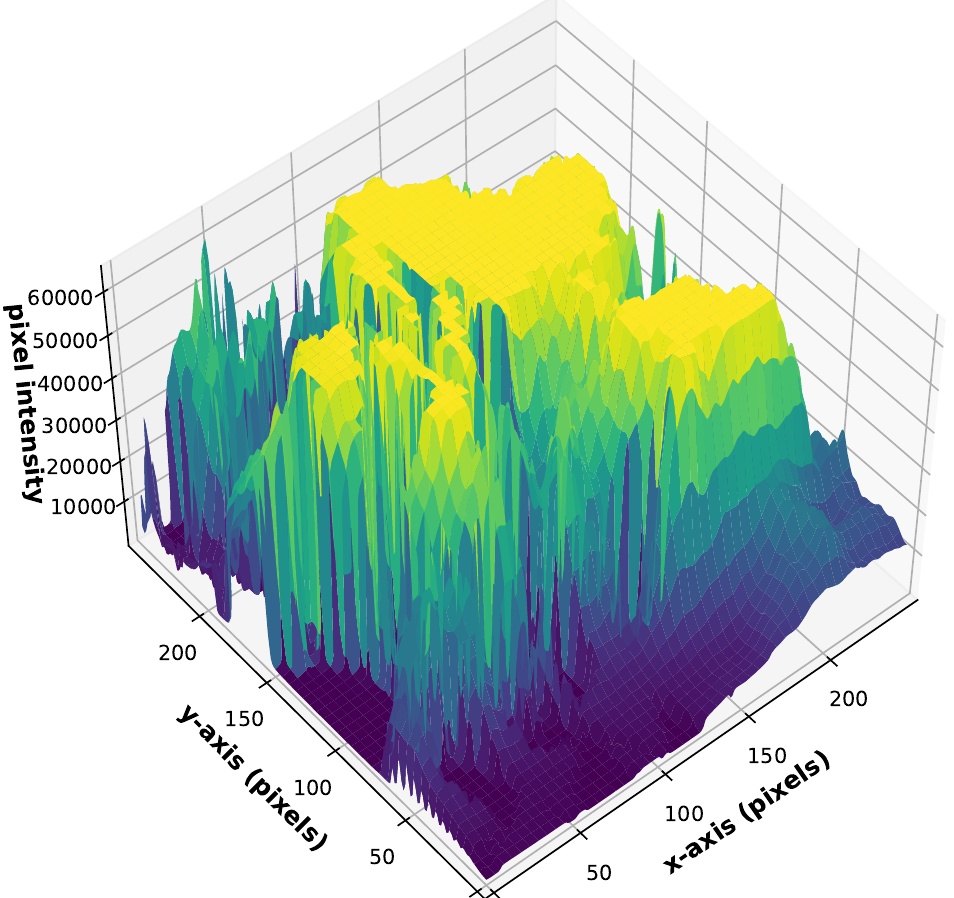} &
    \includegraphics[width=0.4\linewidth]{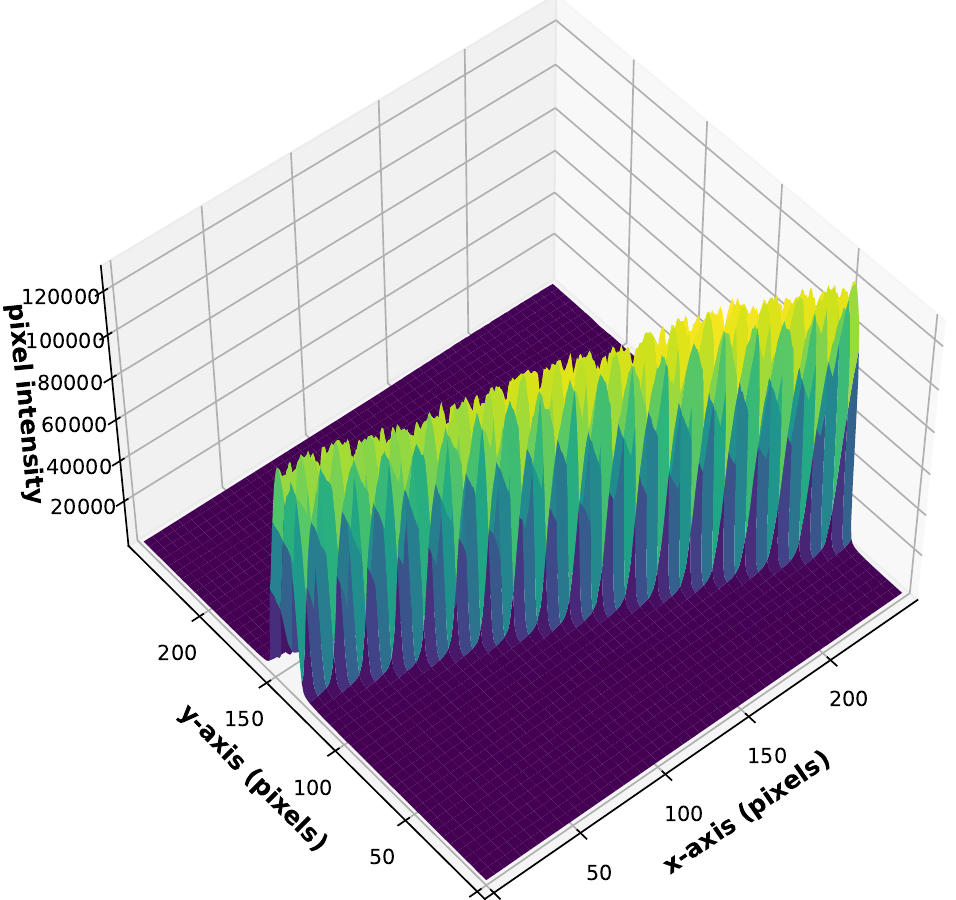} \\
    (a) Si-HDR & BtP-HDR
    \end{tabular}
    \caption{Comparison of images from the (a) Si-HDR~\cite{hanji2022comparison} and (b) BtP-HDR~\cite{bolduc2023beyond} datasets. Image crops are shown on the top row, while the bottom row shows the corresponding plot of the luminance, showing (a) a saturated sample from the Si-HDR dataset; and (b) an unsaturated sample from BtP-HDR. The large flat regions of maximum intensity in (a) correspond to saturated pixel values. The BtP-HDR dataset does not contain such saturation. }
    \label{fig:saturated3D}
\end{figure}

To evaluate our light estimation model from \cref{sec:light_intensity_estimation}, we employ the following two datasets from the literature. 

\myparagraph{Si-HDR~\cite{hanji2022comparison}}
is comprised of 195 captures, each composed of 5 RAW exposures merged into HDR stacks and a reference LDR image for each capture, which is obtained by applying a custom CRF and adding realistic camera noise modeled from the Canon 5D Mark III. The dataset depicts various indoor and outdoor scenes with light sources and saturated highlights in the frame. 

Analyzing the dataset reveals that 82 captures contain saturated pixels, even at the fastest shutter speed, an example of this is shown in \cref{fig:saturated3D}. While these captures allow for expanding the dynamic range of LDR images for purposes such as re-tonemapping, the saturated captures cannot faithfully represent the dynamic range of light sources. 
In our evaluation, we eliminate the saturated captures, which results in 113 HDR images.

\myparagraph{Beyond the Pixel~\cite{bolduc2023beyond} (BtP-HDR)}
The unsaturated HDR dataset proposed for evaluation in \cite{bolduc2023beyond} was expressly conceived to evaluate the presence of bright light sources in images. 
Contrarily to Si-HDR, the LDR reference image of this dataset is an actual capture from a camera, instead of a simulated re-exposure from the HDR merged image. We refer to this dataset BtP-HDR henceforth.

\subsection{Light source estimation}
\label{sec:eval_light_estimation}

\begin{table*}
\centering
\footnotesize
\setlength{\tabcolsep}{3pt}
\begin{tabular}{lccccccccccccc}
& \multicolumn{6}{c}{Si-HDR} & \;\; & \multicolumn{6}{c}{BtP-HDR} \\
\midrule
& \multicolumn{3}{c}{Relight} & \; & \multicolumn{2}{c}{HDR}  & &  \multicolumn{3}{c}{Relight} & \; & \multicolumn{2}{c}{HDR} \\
\midrule
 Method      & MSE$_\downarrow$ & PSNR$_\uparrow$ & Ang err$_\downarrow$ & & H-VDP3$_\uparrow$  & P-PIQE$_\downarrow$ & & MSE$_\downarrow$ & PSNR$_\uparrow$ & Ang err$_\downarrow$ & & H-VDP3$_\uparrow$  & P-PIQE$_\downarrow$ \\
 \midrule
 LDR     & \num{0.0715} & \num{13.0605} & \num{2.77} & & \num{6.86}  & \num{88.38} & & \num{0.0282} & \num{17.5467} & \num{3.15} & & \num{6.68}  & \num{97.21}  \\
 HDRCNN~\cite{eilertsen2017hdrcnn}    & \num{0.0246} & \num{23.6138} & \num{2.59} & & \num{8.47}  & \num{59.13} & & \cellcolor{orange!25}\num{0.0060} & \cellcolor{orange!25}\num{30.6715} & \cellcolor{orange!25}\num{2.02} & & \cellcolor{orange!25}\num{8.611}   & \cellcolor{yellow!25}\num{52.55} \\
 MaskHDR~\cite{santos2020maskHDR} & \num{0.0243} & \num{23.5503} & \num{2.52} & & \cellcolor{orange!25}\num{8.54}  & \num{58.33} & & \cellcolor{yellow!25}\num{0.0061} & \cellcolor{red!25}\num{30.9510} & \cellcolor{red!25}\num{1.86} & & \cellcolor{red!25}\num{8.62}   & \num{54.83} \\
 SingleHDR~\cite{liu2020singlehdr}  & \cellcolor{orange!25}\num{0.0178} & \cellcolor{yellow!25}\num{24.0680} & \cellcolor{yellow!25}\num{3.32} & & \cellcolor{red!25}\num{8.57}  & \cellcolor{orange!25}\num{40.34} & & \num{0.0092} & \num{23.6101} & \num{3.25} & & \num{7.91}   & \cellcolor{red!25}\num{40.58} \\
 ExpandNet~\cite{marnerides2018expandnet}  & \cellcolor{yellow!25}\num{0.0187} & \cellcolor{orange!25}\num{24.7158} & \cellcolor{red!25}\num{2.10} & & \num{8.48}  & \num{58.11} & & \num{0.0148} & \num{22.5799} & \num{4.56} & & \num{8.06}   & \num{56.46} \\
 Exposure Diff~\cite{bemana2024exposurediffusion}    & \num{0.0264} & \num{23.7384} & \num{8.23} & & \num{7.88}  & \cellcolor{yellow!25}\num{47.94} & & \cellcolor{yellow!25}\num{0.0061} & \num{27.4963} & \num{2.51} & & \num{8.38}  & \num{63.51} \\
 Ours       & \cellcolor{red!25}\num{0.0131} & \cellcolor{red!25}\num{25.3895} & \cellcolor{orange!25}\num{2.17} & & \cellcolor{yellow!25}\num{8.49}  & \cellcolor{red!25}\num{40.17} & & \cellcolor{red!25}\num{0.0041} & \cellcolor{yellow!25}\num{28.7022} & \cellcolor{yellow!25}\num{2.39} & &  \cellcolor{yellow!25}\num{8.52}  & \cellcolor{orange!25}\num{52.41}\\
\end{tabular}
\caption{Quantitative evaluation of renderings lit by each method's HDR image output. Results are color coded by \colorbox{red!25}{best}, \colorbox{orange!25}{second} and \colorbox{yellow!25}{third} best. ``P-PIQE'' refers to PU21-PIQE and ``H-VPD3'' to HDR-VDP-3.}
\label{tab:quant_relighting}
\end{table*}

\begin{table}[]
    \centering
    \footnotesize
    \setlength{\tabcolsep}{3pt}
    \begin{tabular}{lcccccc}
     & \multicolumn{3}{c}{Relight} & \; & \multicolumn{2}{c}{HDR} \\
     \midrule
     Method      & MSE$_\downarrow$ & PSNR$_\uparrow$ & Ang. err$_\downarrow$ & & H-VDP3$_\uparrow$ & P-PIQE$_\downarrow$ \\
     \midrule
     LDR     & \cellcolor{yellow!25}\num{0.049} & \cellcolor{yellow!25}\num{14.04} & \cellcolor{yellow!25}\num{1.83} & & \cellcolor{yellow!25}\num{7.30} & \cellcolor{yellow!25}\num{90.00} \\
     GlowGAN~\cite{wang2023glowgan}     & \cellcolor{red!25}\num{0.012} & \cellcolor{orange!25}\num{23.82} & \cellcolor{red!25}\num{1.09} & & \cellcolor{orange!25}\num{5.67} & \cellcolor{orange!25}\num{56.82} \\
     Ours     & \cellcolor{orange!25}\num{0.013} & \cellcolor{red!25}\num{25.05} & \cellcolor{orange!25}\num{1.14} & & \cellcolor{red!25}\num{8.49} & \cellcolor{red!25}\num{37.56} \\
    \end{tabular}
    \caption{\modif{Quantitative comparison to GlowGAN~\cite{wang2023glowgan}, trained on the ``landscape'' category, on selected landscape images from the Si-HDR dataset. Note that, unlike GlowGAN, our network is not category-specific. Results are color-coded by \colorbox{red!25}{best}, \colorbox{orange!25}{second} and \colorbox{yellow!25}{third} best.} ``P-PIQE'' refers to PU21-PIQE and ``H-VDP3'' to HDR-VDP-3.}
    \label{tab:quant_glowgan}
\end{table}

We evaluate our novel light intensity estimation method (\cref{sec:light_intensity_estimation}) on the task of predicting bright light sources. We first show qualitative results on the filtered Si-HDR dataset in \cref{fig:relight_glossy_sihdr}. The figures illustrate both the luminance of the generated HDR maps, as well as a render of a sphere on a plane lit by the color HDR map. Since these images are not full 360\textdegree{} environment maps, we project them onto a partial hemisphere of 90\textdegree{} field of view, and use this as light source for rendering. 
The sphere rests on top of a plane positioned at the origin, facing the camera. The glossy sphere has an albedo of 1 (pure white) and a roughness coefficient of 0. We ensure that all HDR images used for lighting are exposed the same way by aligning the pixel averages with intensity between $0.2$ and $0.8$ with the ground truth. Our method's predictions closely match the ground truth, particularly with bright light sources that cast hard shadows and create specularities. Please, see the supplementary material for qualitative results on the BtP-HDR dataset. 

To further assess the accuracy of our method, we compute quantitative image comparison metrics \modif{on both renders (MSE, PSNR, and angular error) and the HDR images (HDR-VDP-3~\cite{mantiuk2023hdr} and PU21-PIQE~\cite{mantiuk2021pu}), compared to the ground truth in \cref{tab:quant_relighting}}. 
We observe that our method significantly outperforms existing methods in terms of both MSE and PSNR on the filtered Si-HDR dataset. On the BtP-HDR dataset, our method provides the best MSE score while offering a competitive PSNR value. \modif{We attribute this lower score to the fact that BtP-HDR was captured with a fixed white balance, leading to out-of-distribution color renditions. Our method predicts convincing content in saturated regions, which can lead to potentially different colors from the ground truth, as indicated by the angular error scores. Although our method is primarily designed for relighting, it also performs well in direct metrics. }

\modif{To compare with GlowGAN~\cite{wang2023glowgan}, a subset of 23 landscape images from the Si-HDR dataset is selected, estimating HDR with the specialized checkpoints. As shown in \cref{tab:quant_glowgan}, our method is competitive even against the specialized GlowGAN. }

\subsection{3D lighting representation}

\begin{table}
\centering
\footnotesize
\setlength{\tabcolsep}{2pt}
\begin{tabular}{lccccccc}
& \multicolumn{3}{c}{Chess room} & \; & \multicolumn{3}{c}{Small class} \\
\midrule
& \multicolumn{2}{c}{Relight} & \multicolumn{1}{c}{HDR} & & \multicolumn{2}{c}{Relight} & \multicolumn{1}{c}{HDR} \\
\midrule
 Scene      & MSE$_\downarrow$ & PSNR$_\uparrow$ & P-PSNR$_\uparrow$ & & MSE$_\downarrow$ & PSNR$_\uparrow$ & P-PSNR$_\uparrow$ \\
 \midrule
 \textsmaller[1.5]{PanoHDR-NeRF}     & \num{0.0134} & \num{18.97} & \num{27.61} & & \num{0.0077} & \num{21.68} & \num{28.57} \\
 Ours  & \cellcolor{red!25}\num{0.0026} & \cellcolor{red!25}\num{30.50} & \cellcolor{red!25}\num{30.29} & & \cellcolor{red!25}\num{0.0049} & \cellcolor{red!25}\num{23.20} & \cellcolor{red!25}\num{30.77}  \\
\end{tabular}
\caption{Quantitative evaluation of renderings lit by novel views of PanoHDR-NeRF~\cite{gera2022panohdrnerf} and Ours, as compared to the captured HDR ground truth panoramas provided in \cite{gera2022panohdrnerf}. Our method achieves the best scores on both scenes for all metrics. ``P-PSNR'' refers to PU21-PSNR.}
\label{tab:quant_3D}
\end{table}

\begin{figure*}
   \centering
   \footnotesize
   \setlength{\tabcolsep}{0.5pt}
   \newlength{\tmplength}
   \setlength{\tmplength}{0.11\linewidth}
   \newlength{\cbarheight}
   \setlength{\cbarheight}{1.3cm}
    \begin{tabular}{cccccccccc}
    Input & LDR & HDRCNN & MaskHDR & SingleHDR & ExpandNet & Exposure Diff & Ours & GT & \\
    \includegraphics[width=\tmplength]{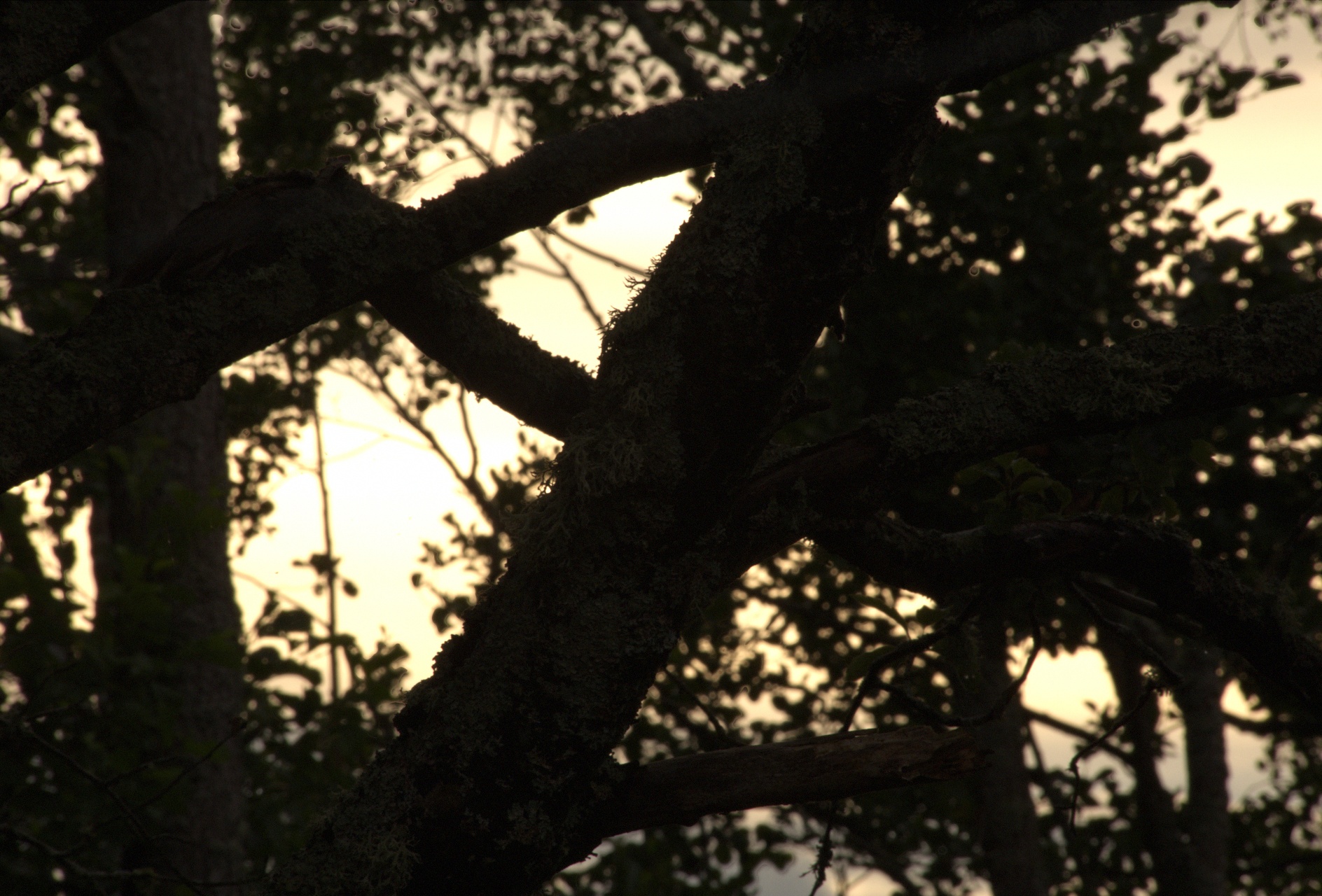}&
    \includegraphics[width=\tmplength]{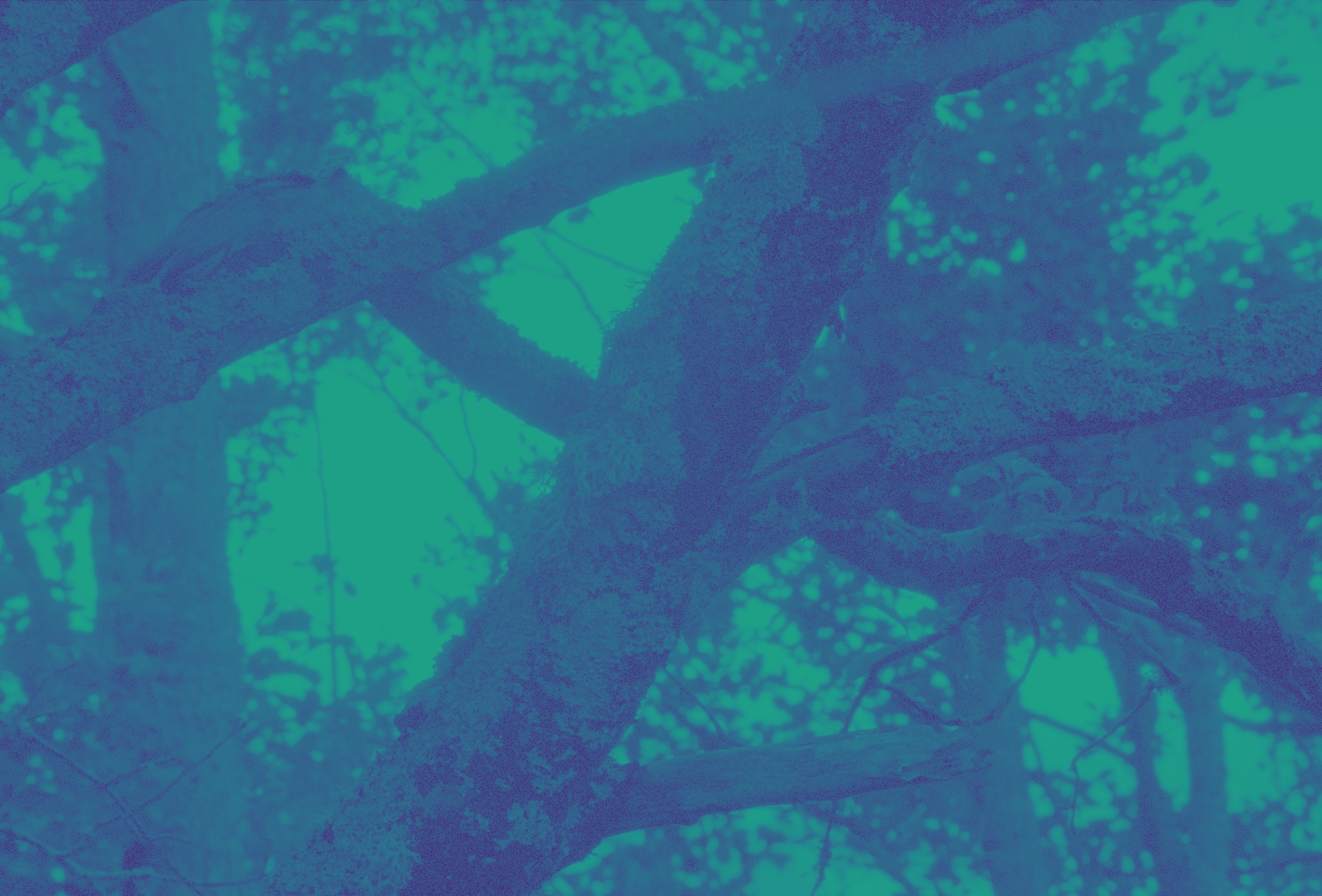}&
    \includegraphics[width=\tmplength]{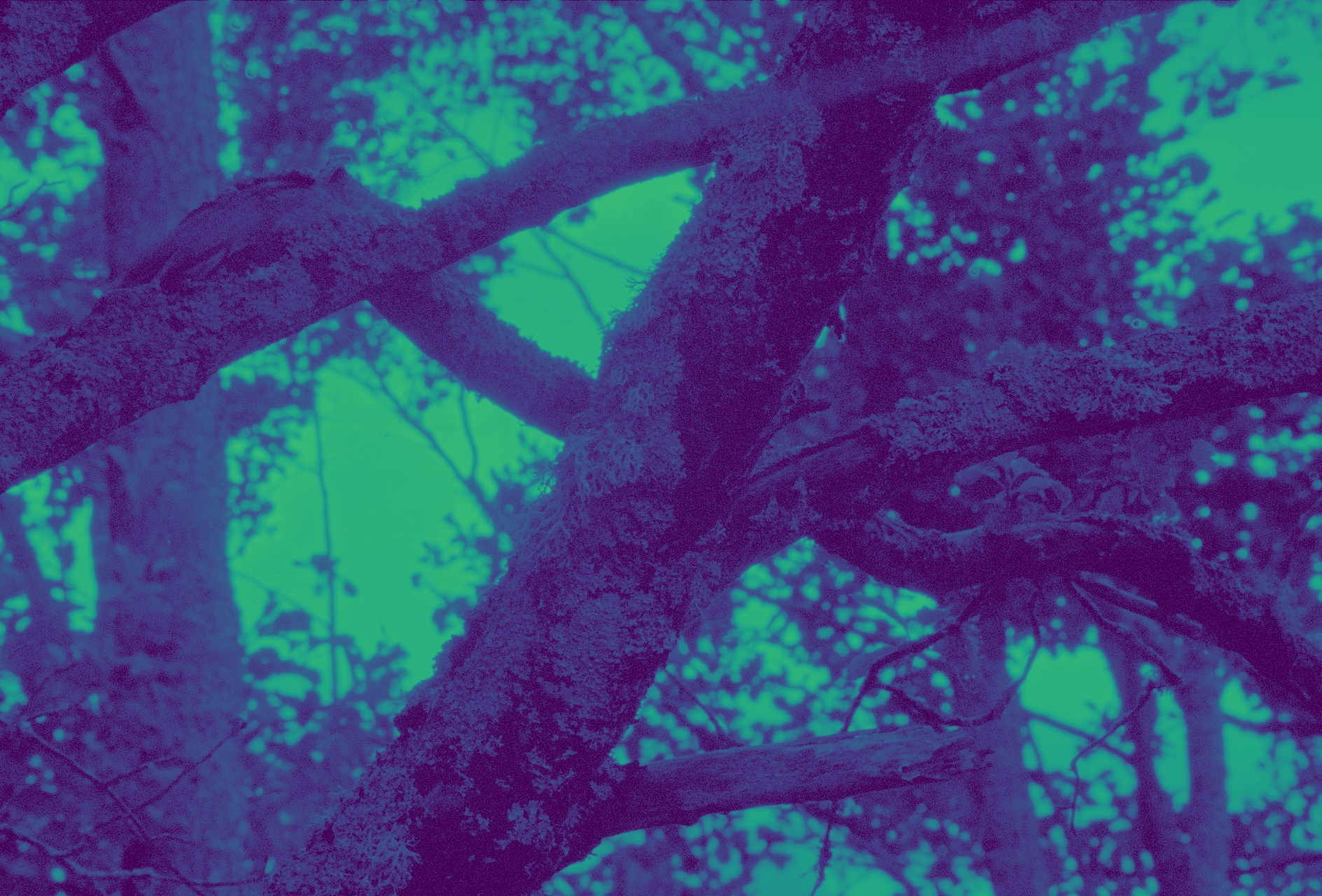}&
    \includegraphics[width=\tmplength]{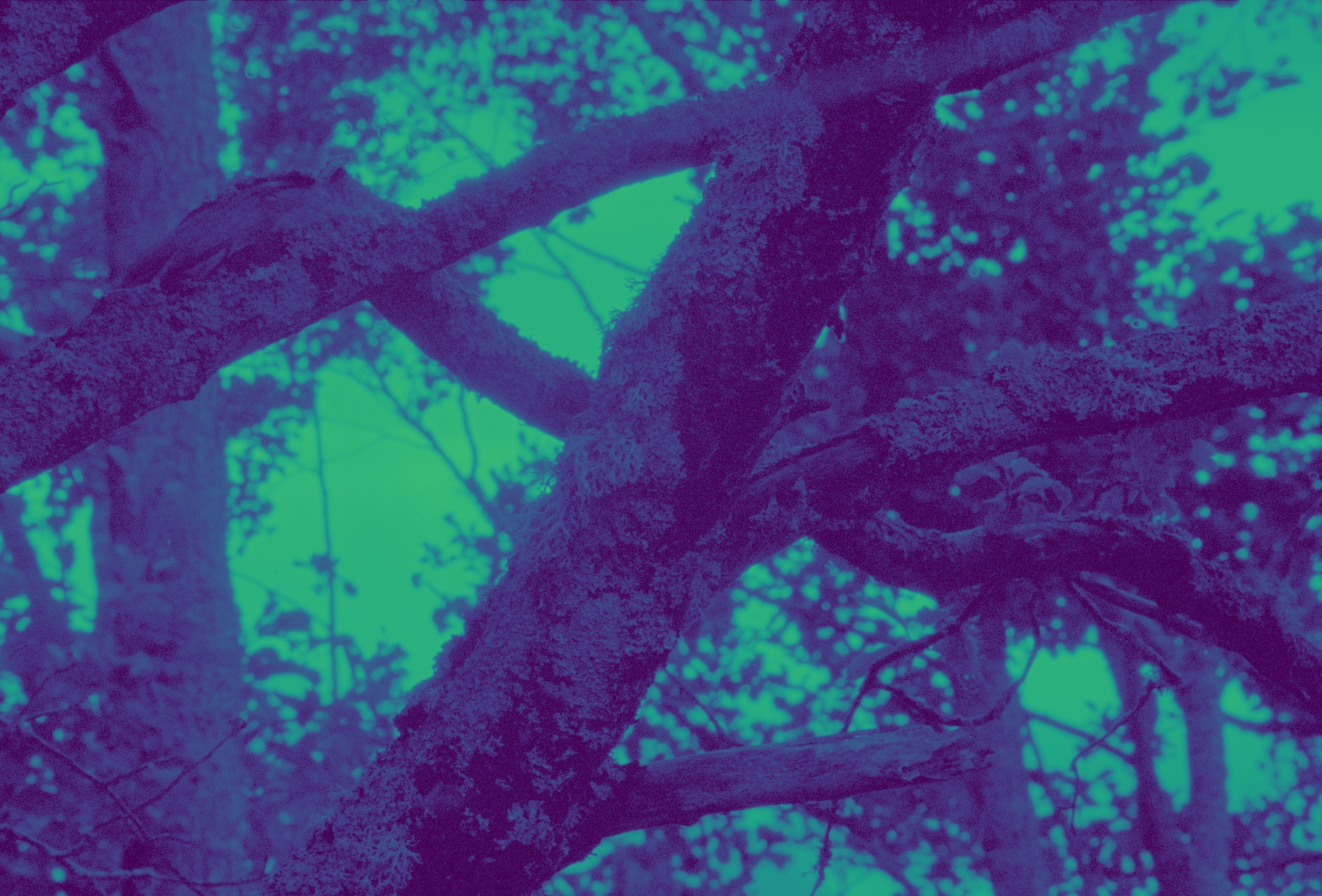}&
    \includegraphics[width=\tmplength]{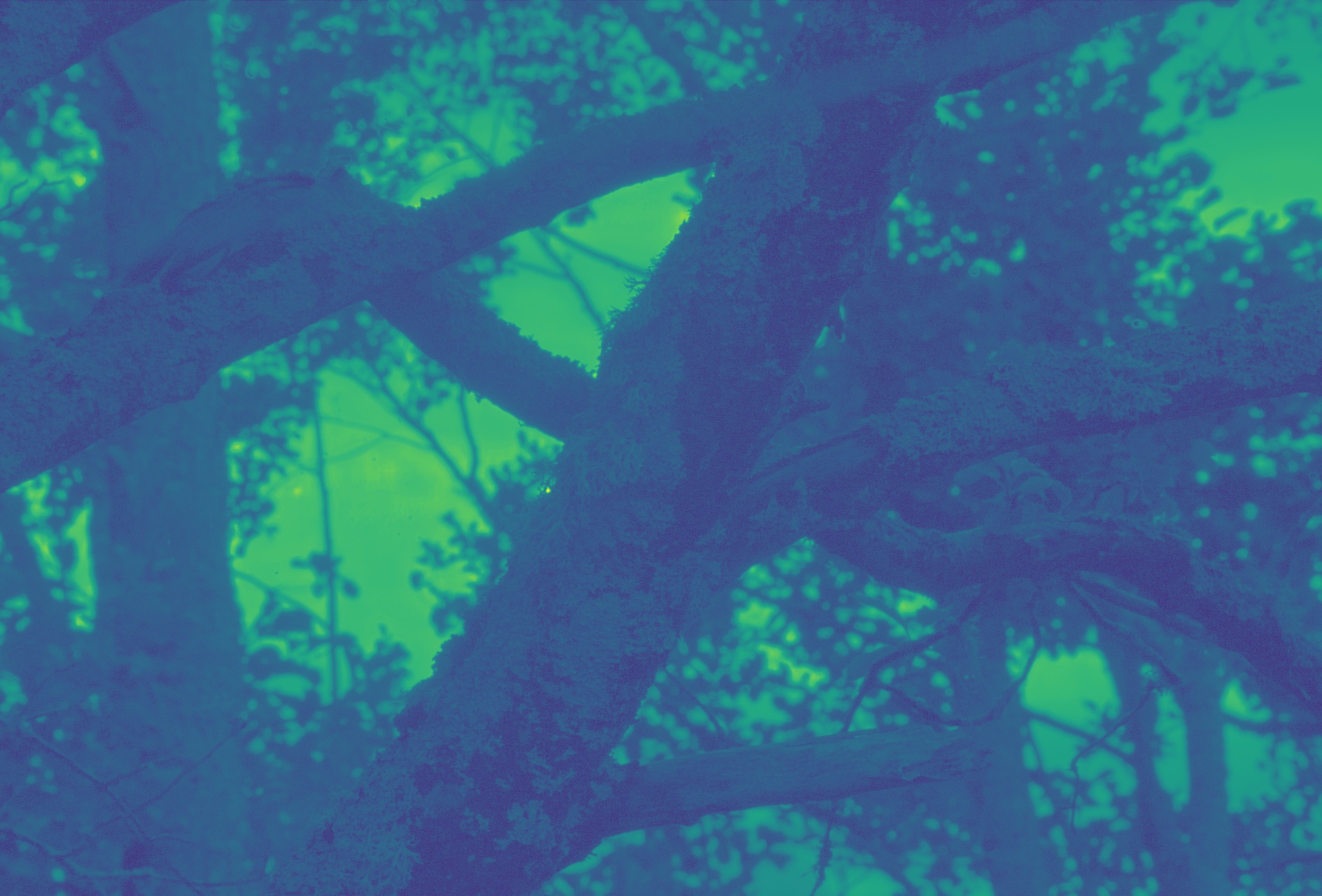}&
    \includegraphics[width=\tmplength]{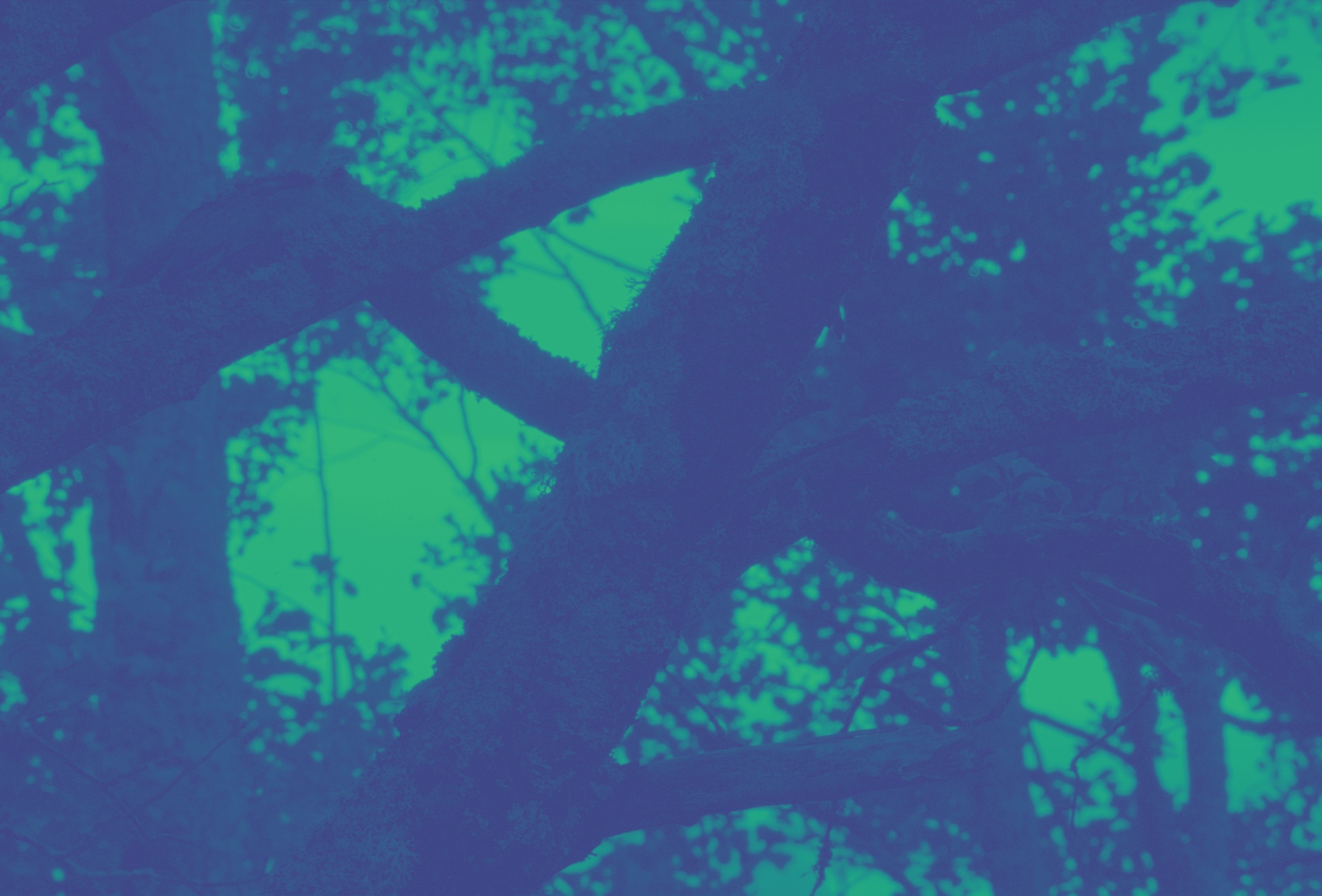}&
    \includegraphics[width=\tmplength]{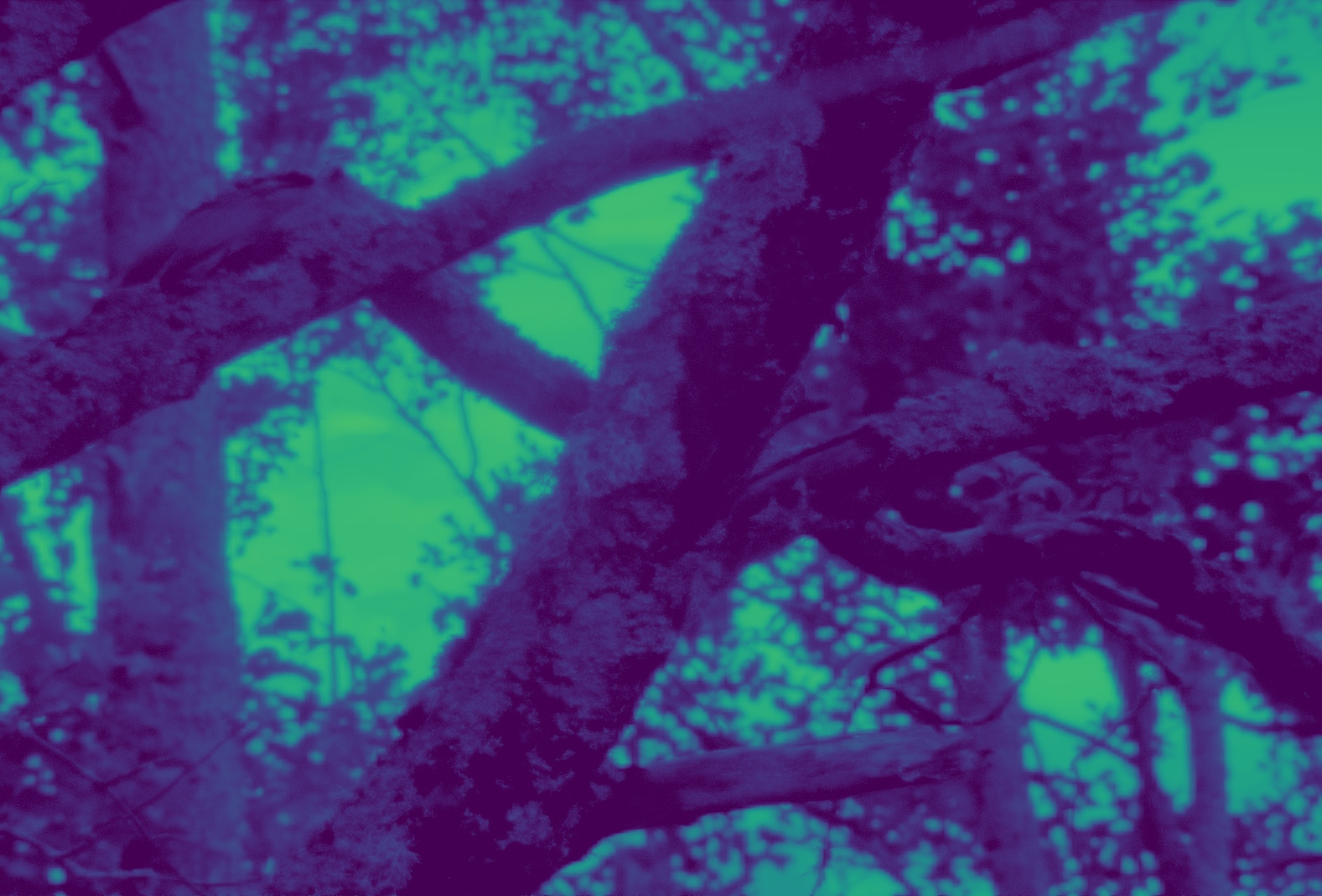}&
    \includegraphics[width=\tmplength]{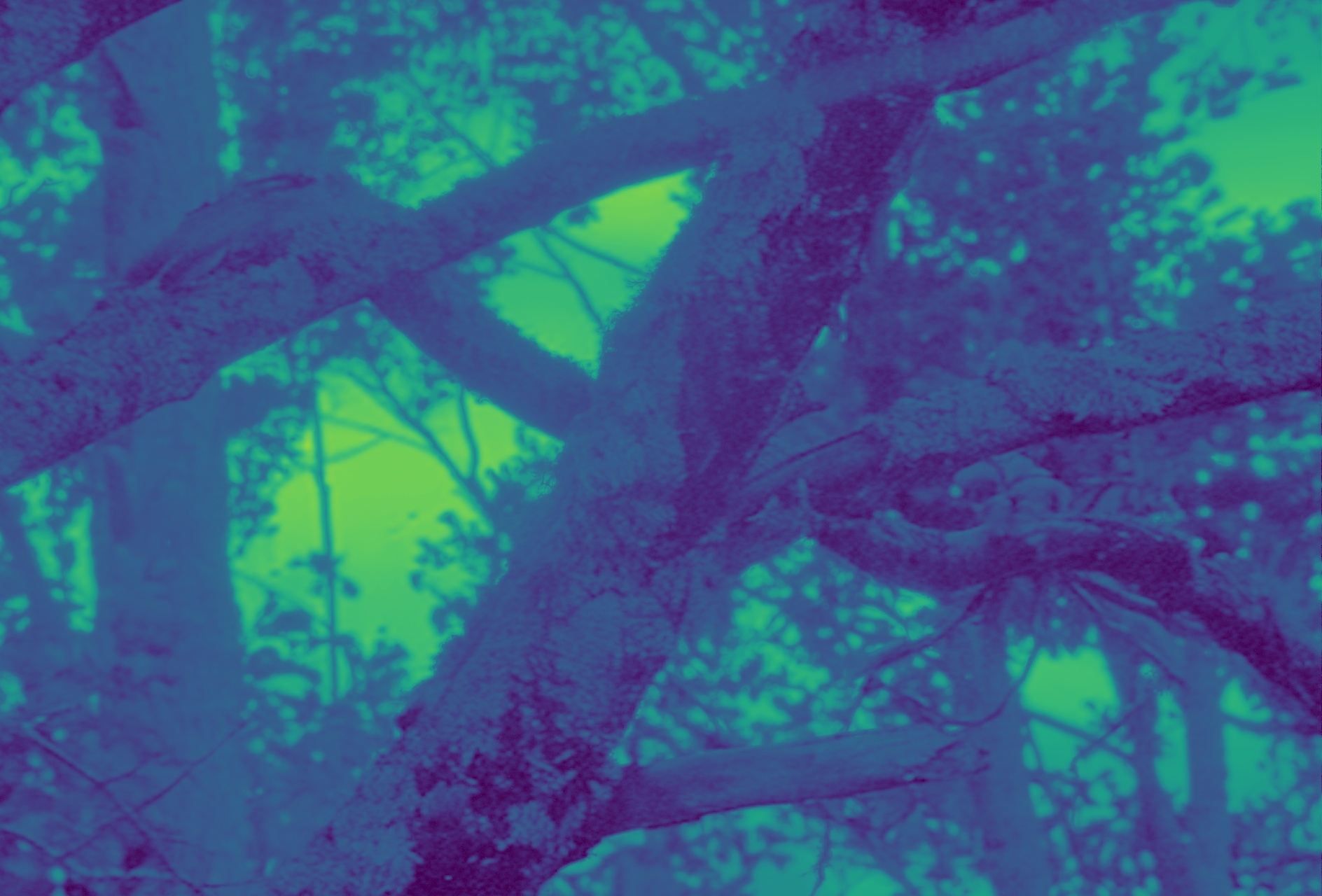}&
    \includegraphics[width=\tmplength]{figures/relighting_glossy_si_hdr/energy/042/diffusionhdr/042.jpg} &
    \includegraphics[height=\cbarheight]{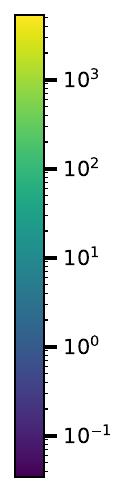}
     \\
    &
    \includegraphics[width=\tmplength]{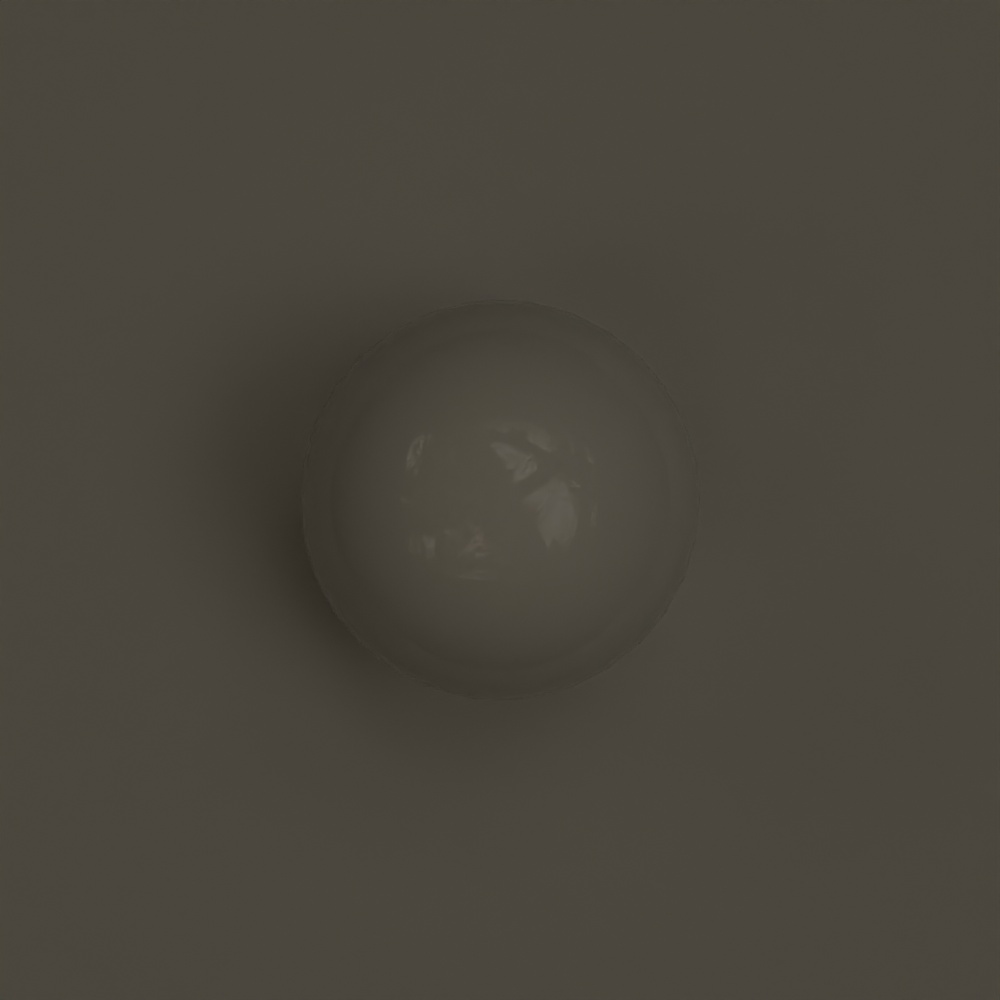}&
    \includegraphics[width=\tmplength]{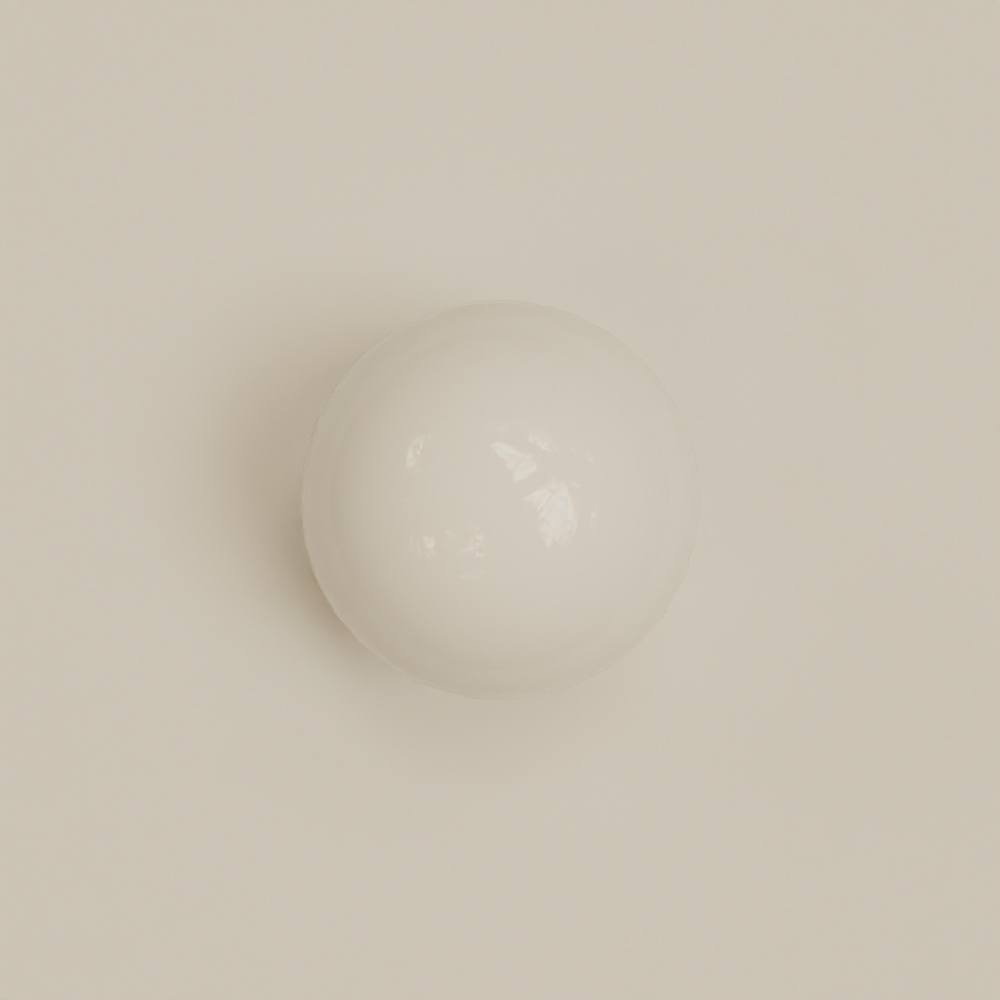}&
    \includegraphics[width=\tmplength]{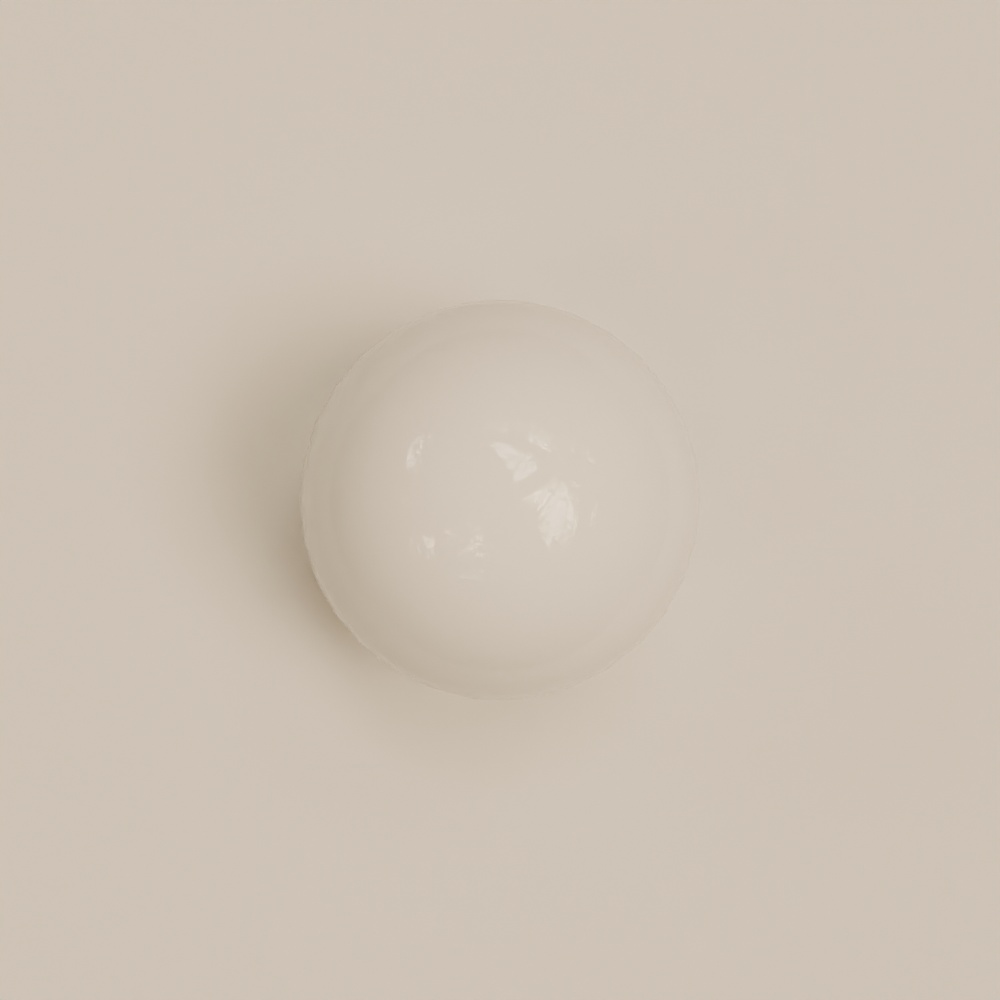}&
    \includegraphics[width=\tmplength]{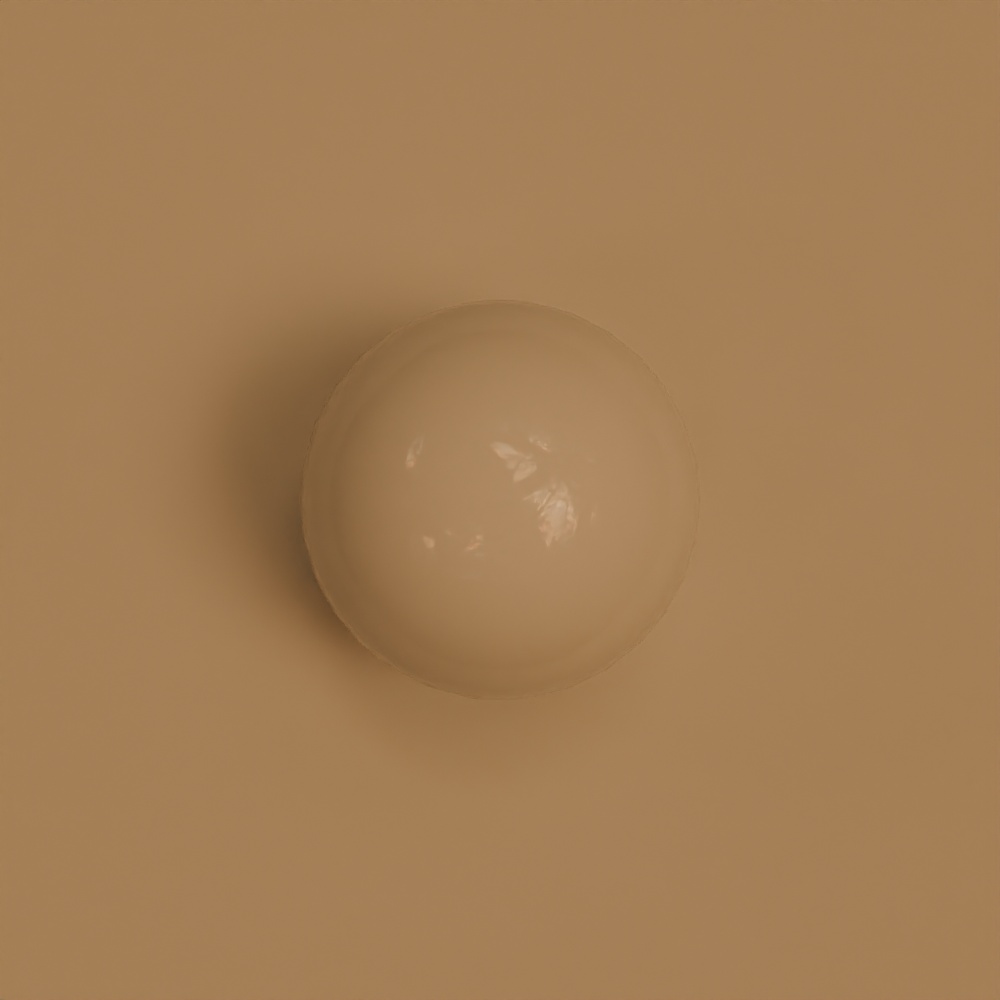}&
    \includegraphics[width=\tmplength]{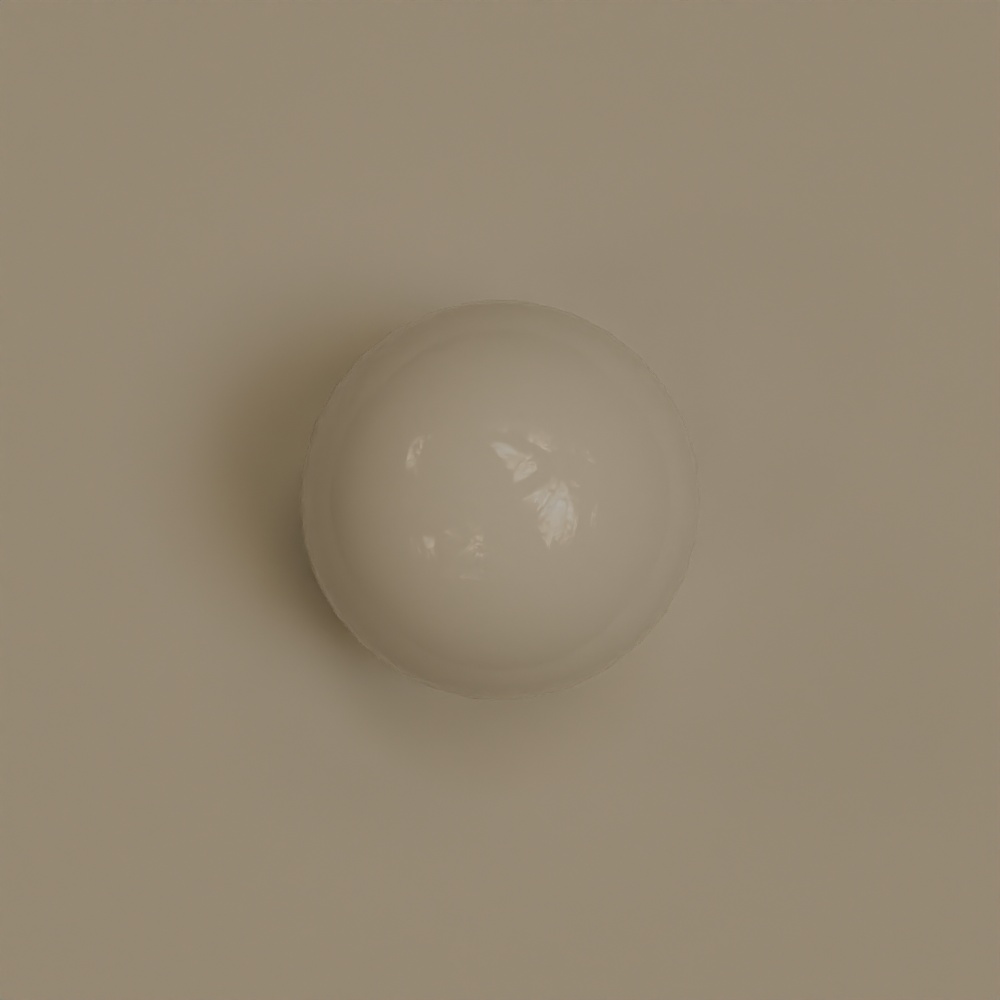}&
    \includegraphics[width=\tmplength]{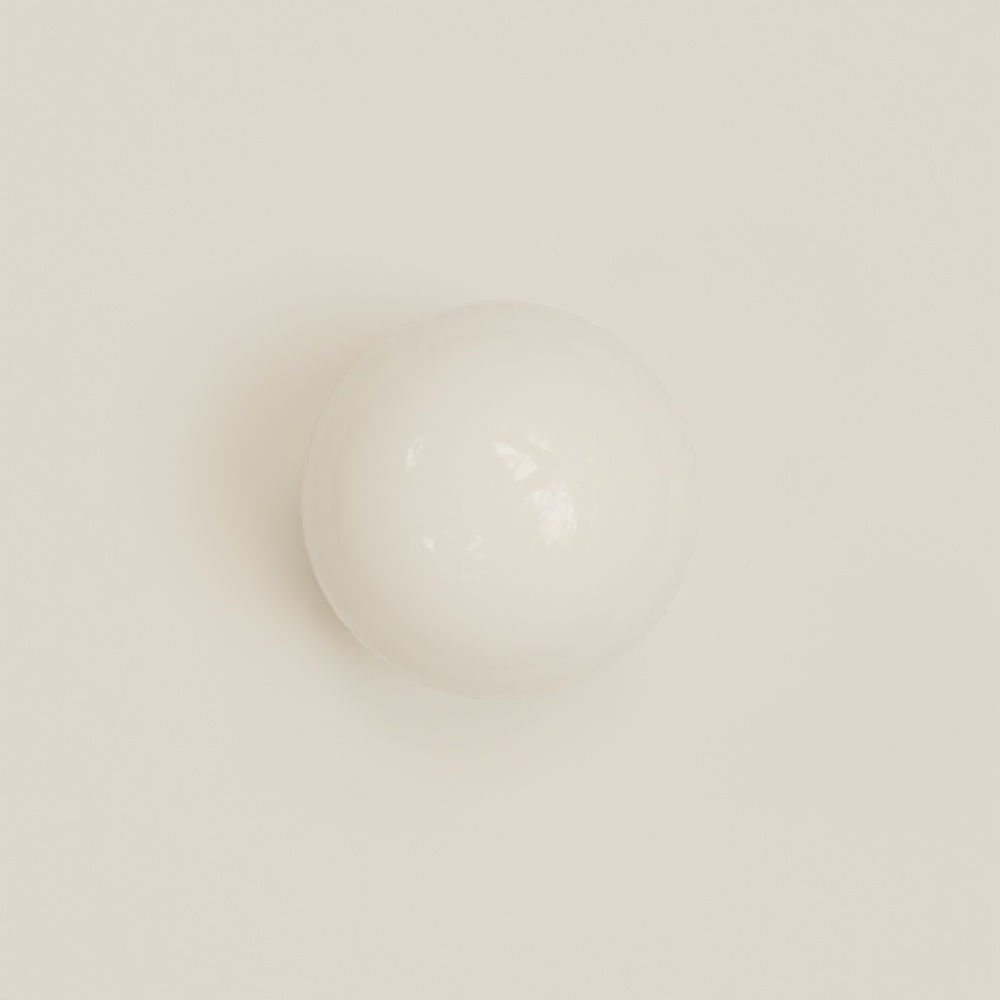}&
    \includegraphics[width=\tmplength]{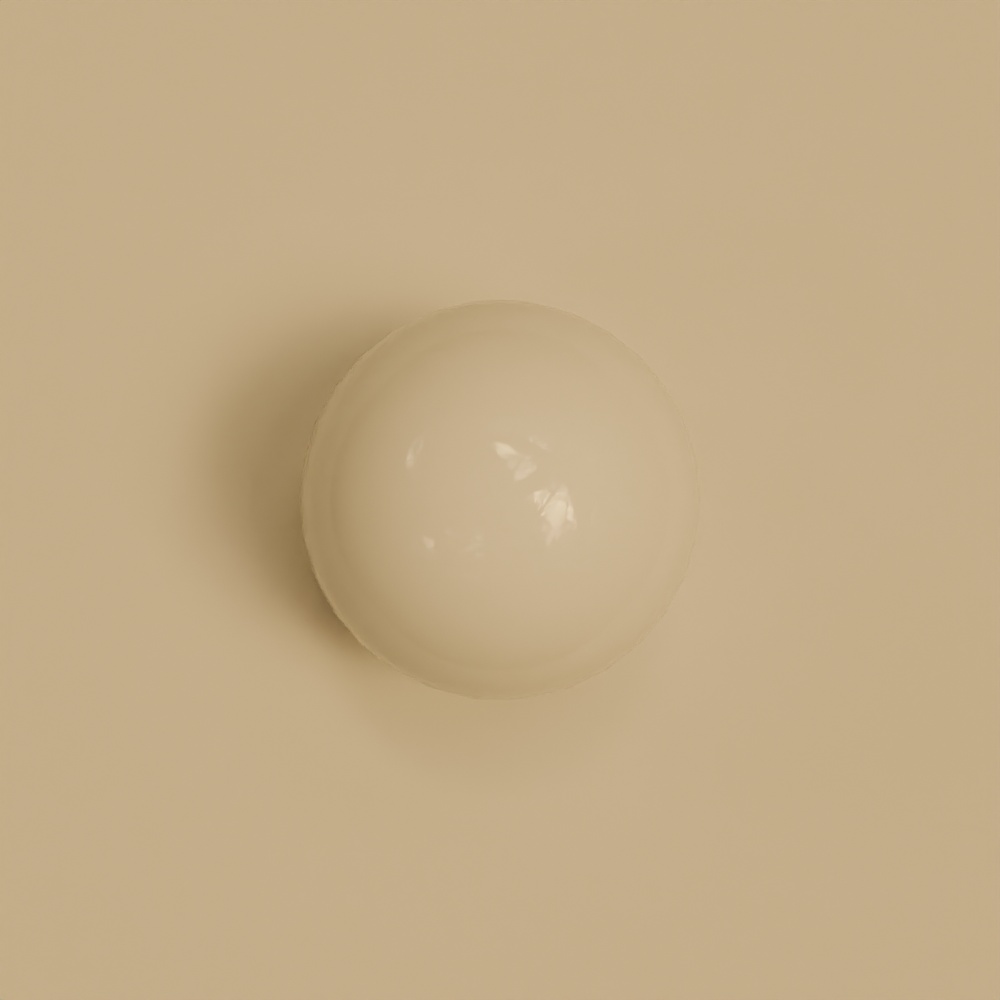}&
    \includegraphics[width=\tmplength]{figures/relighting_glossy_si_hdr/042/diffusionhdr/042.jpg} & \\
    \includegraphics[width=\tmplength]{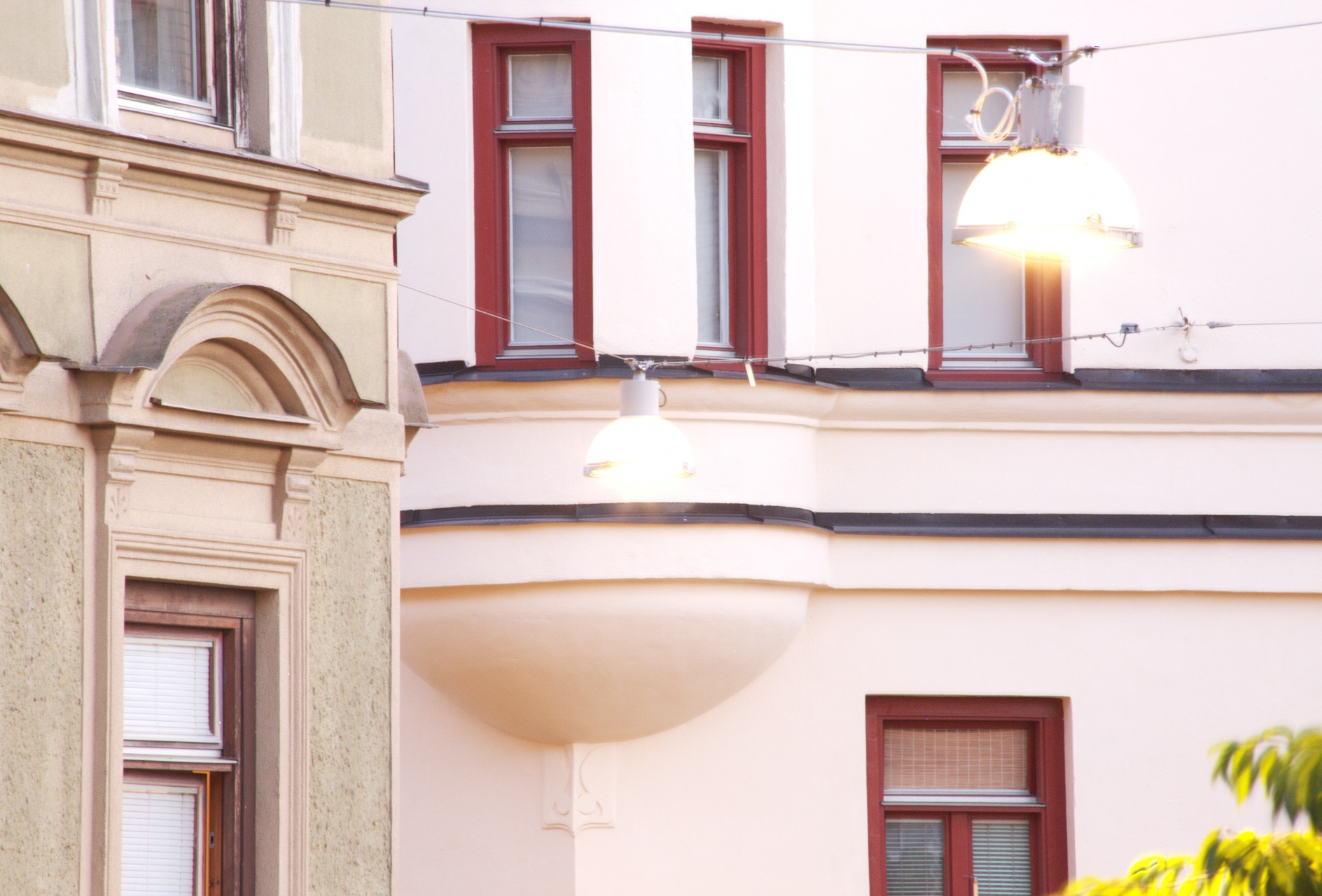}&
    \includegraphics[width=\tmplength]{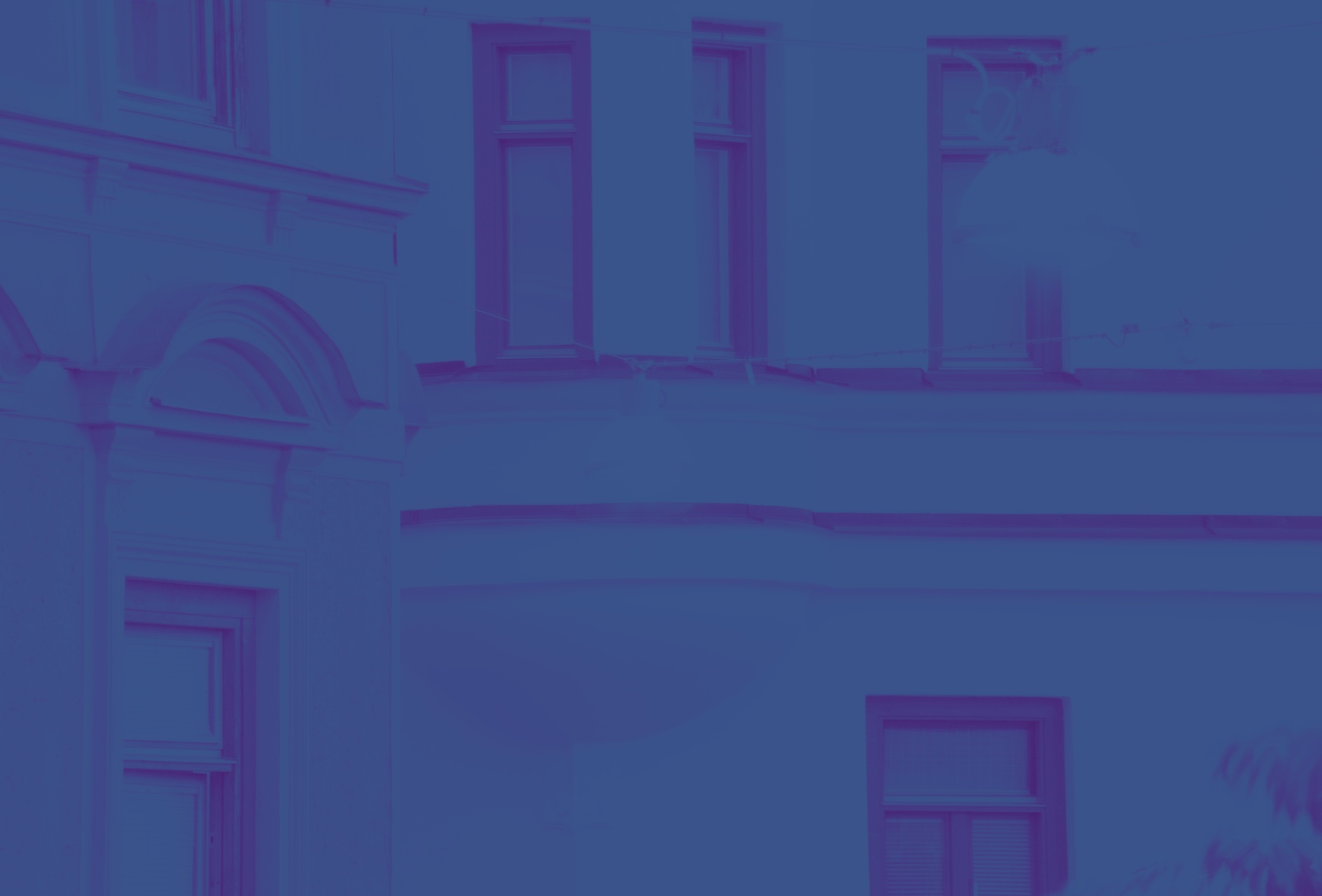}&
    \includegraphics[width=\tmplength]{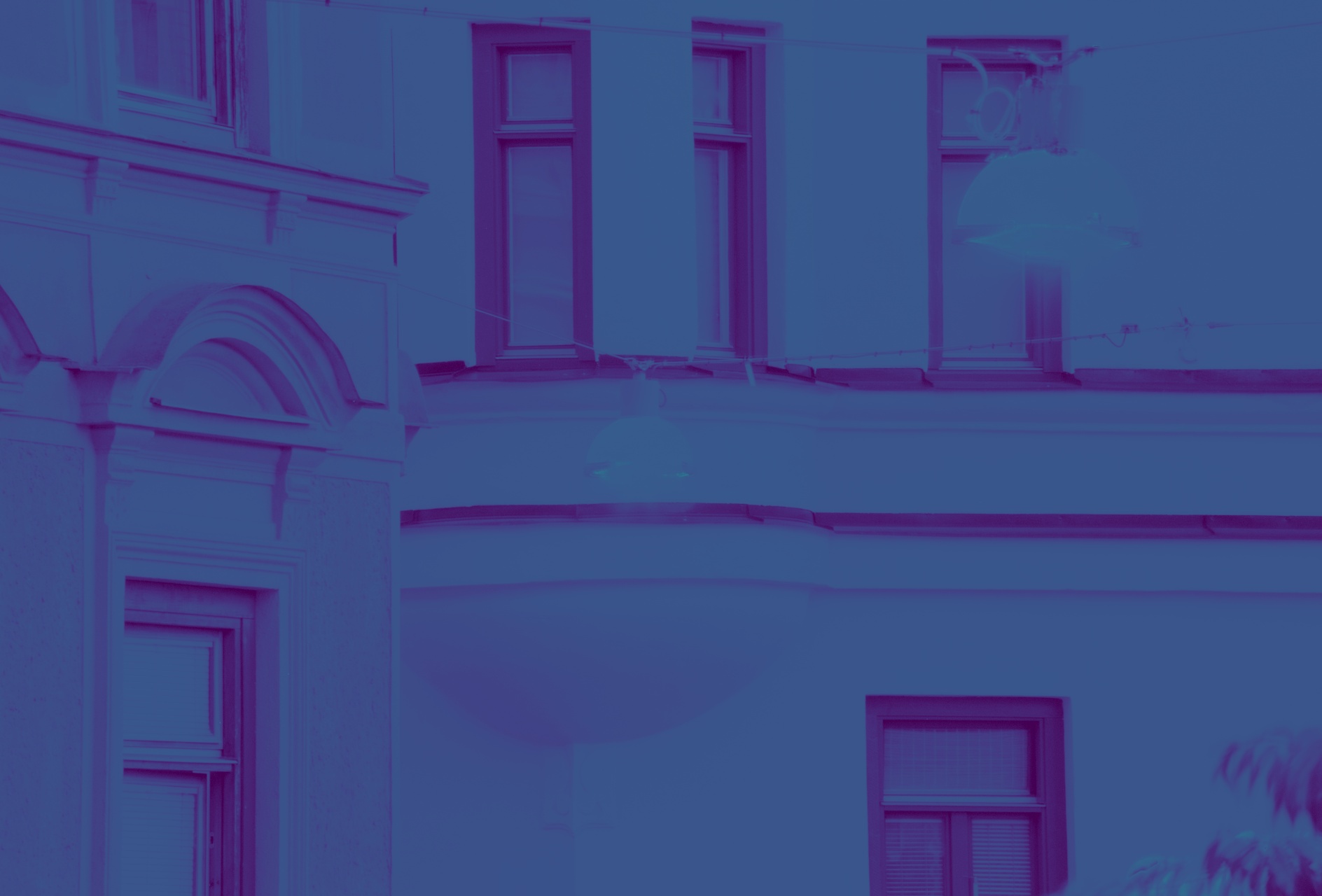}&
    \includegraphics[width=\tmplength]{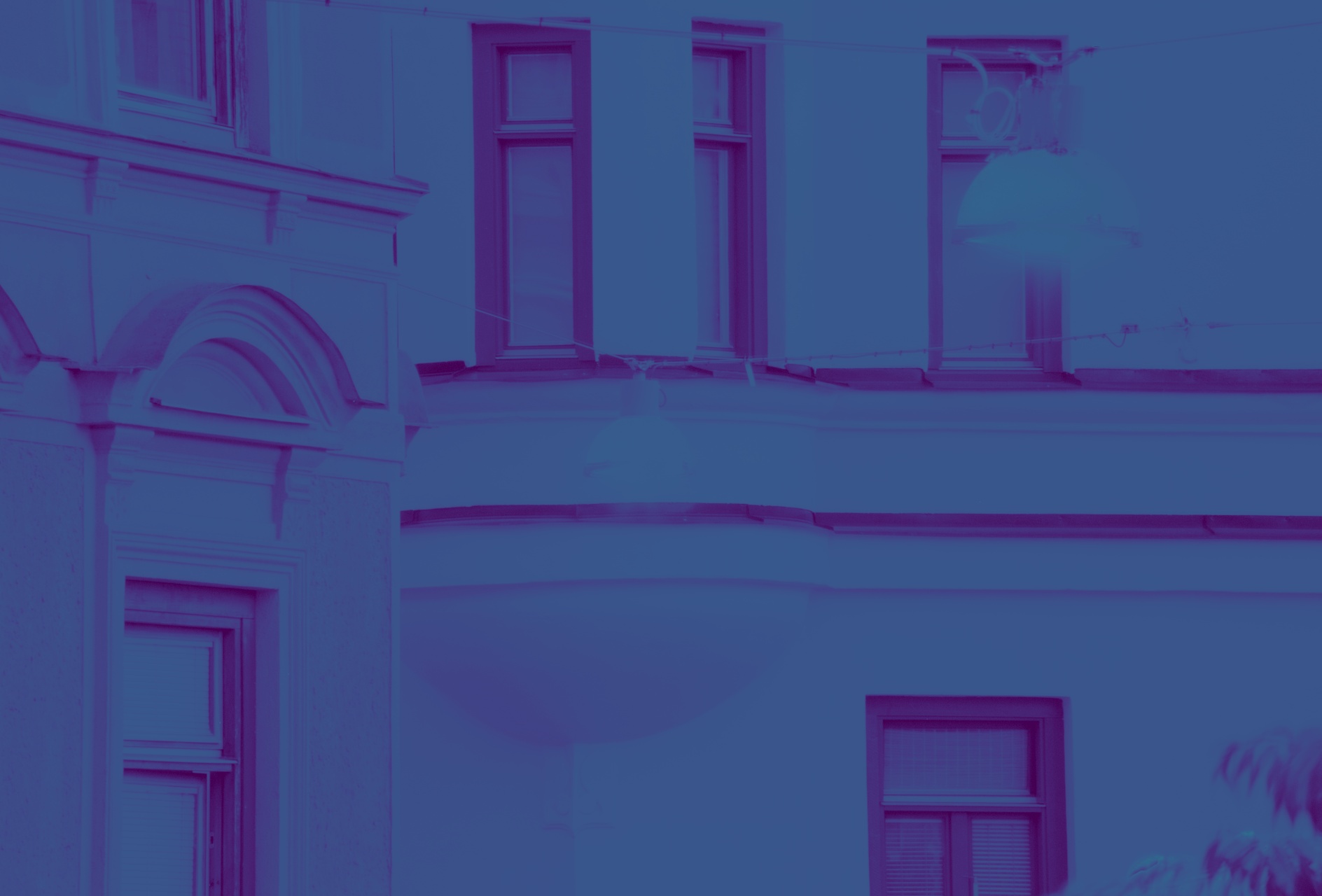}&
    \includegraphics[width=\tmplength]{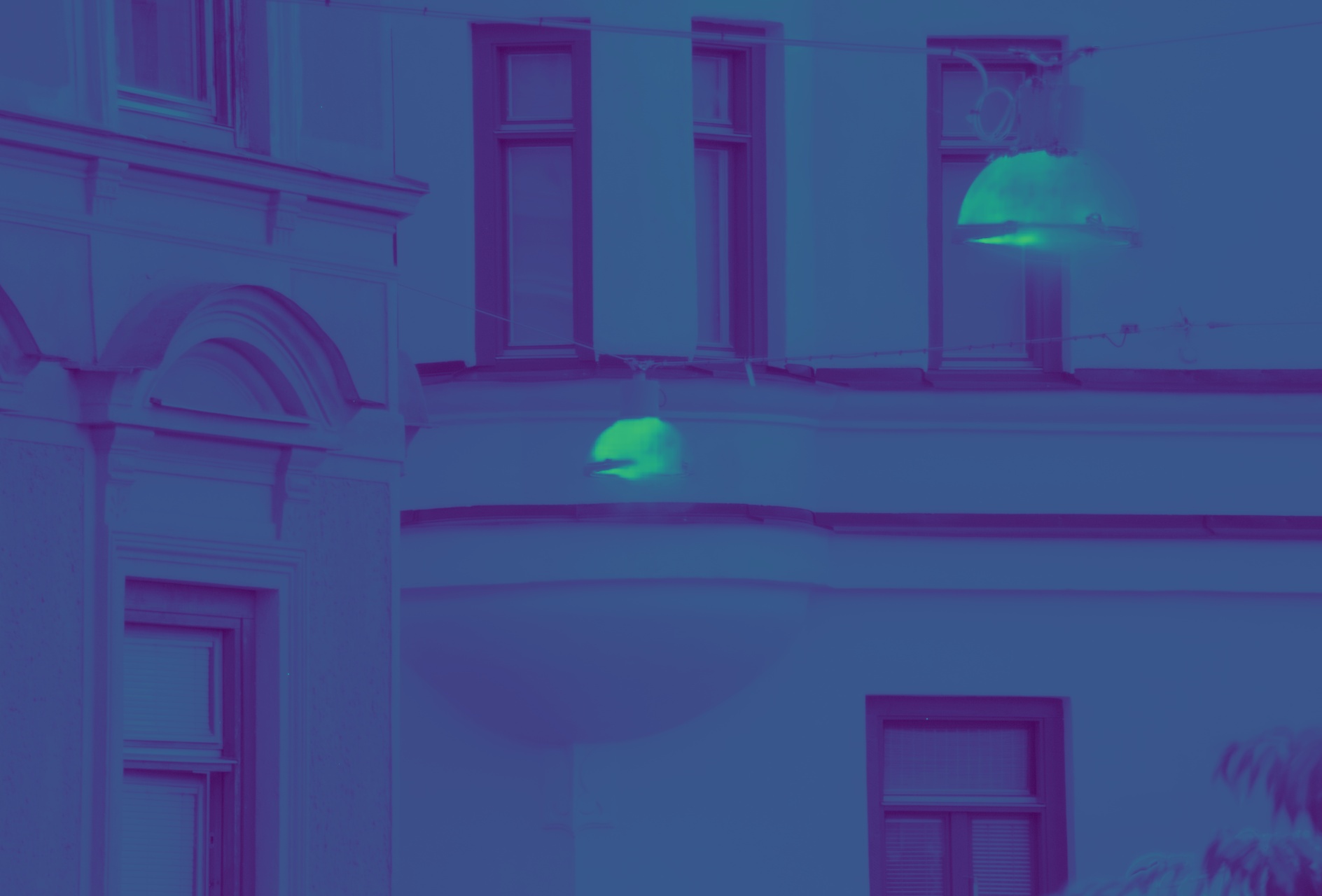}&
    \includegraphics[width=\tmplength]{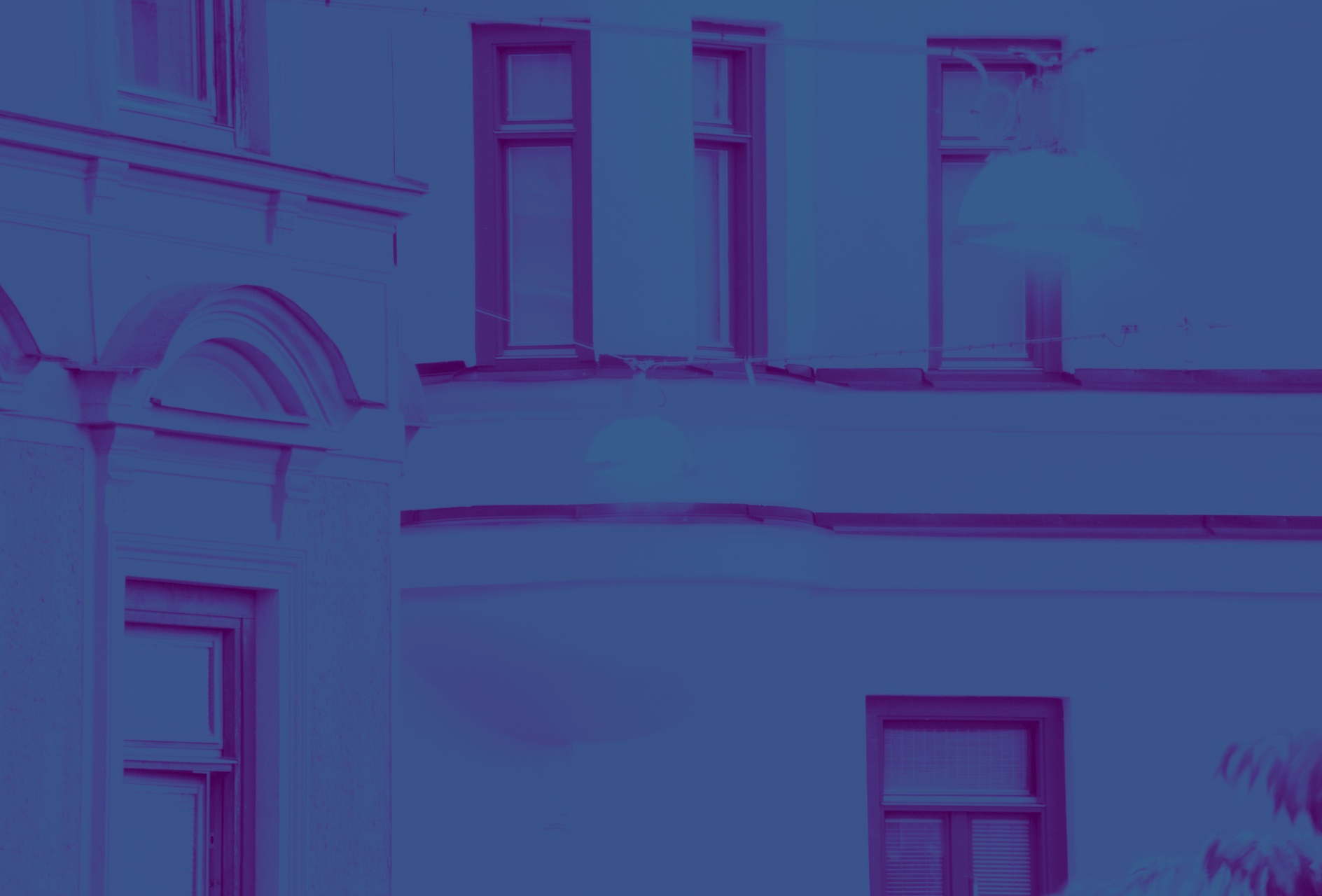}&
    \includegraphics[width=\tmplength]{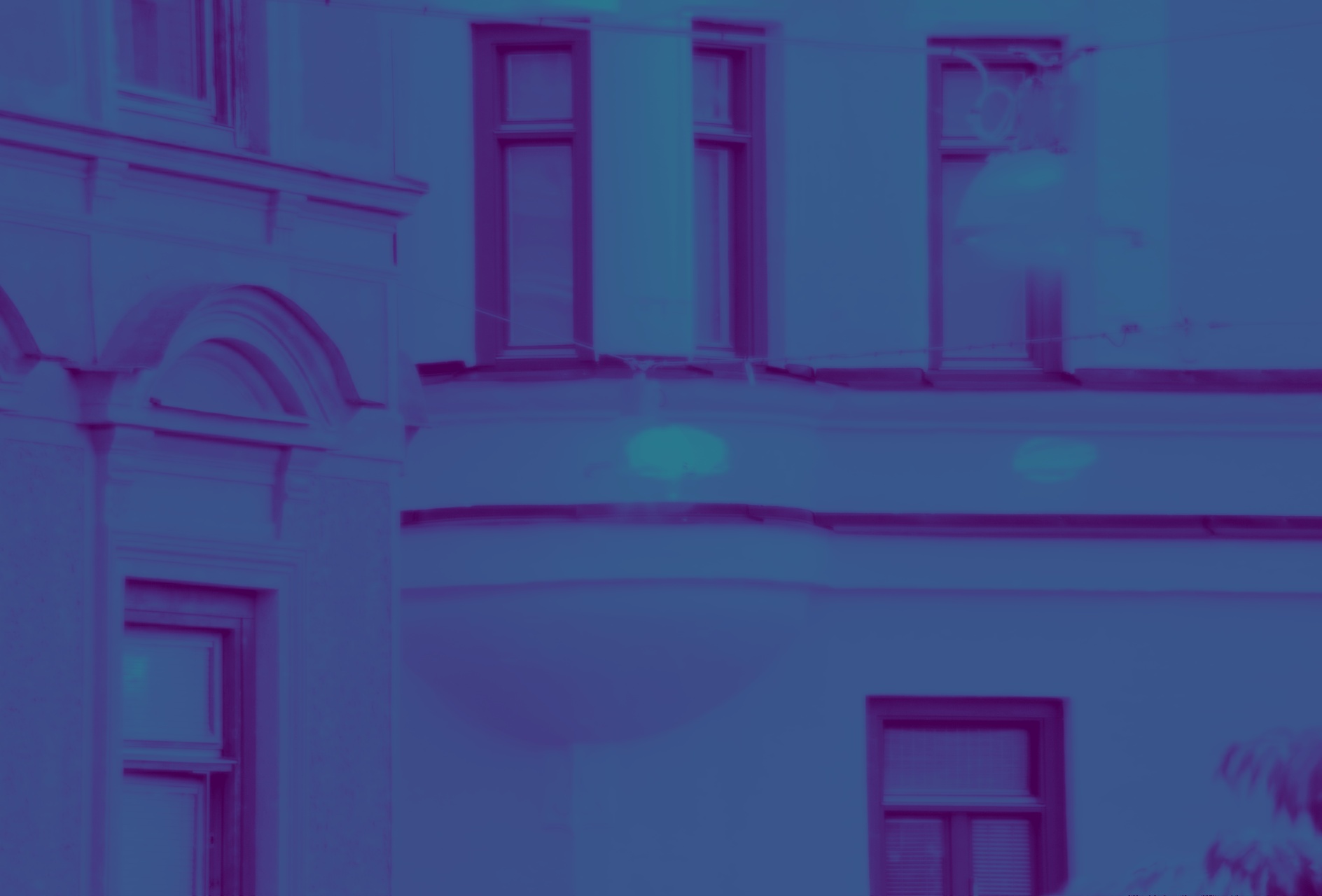}&
    \includegraphics[width=\tmplength]{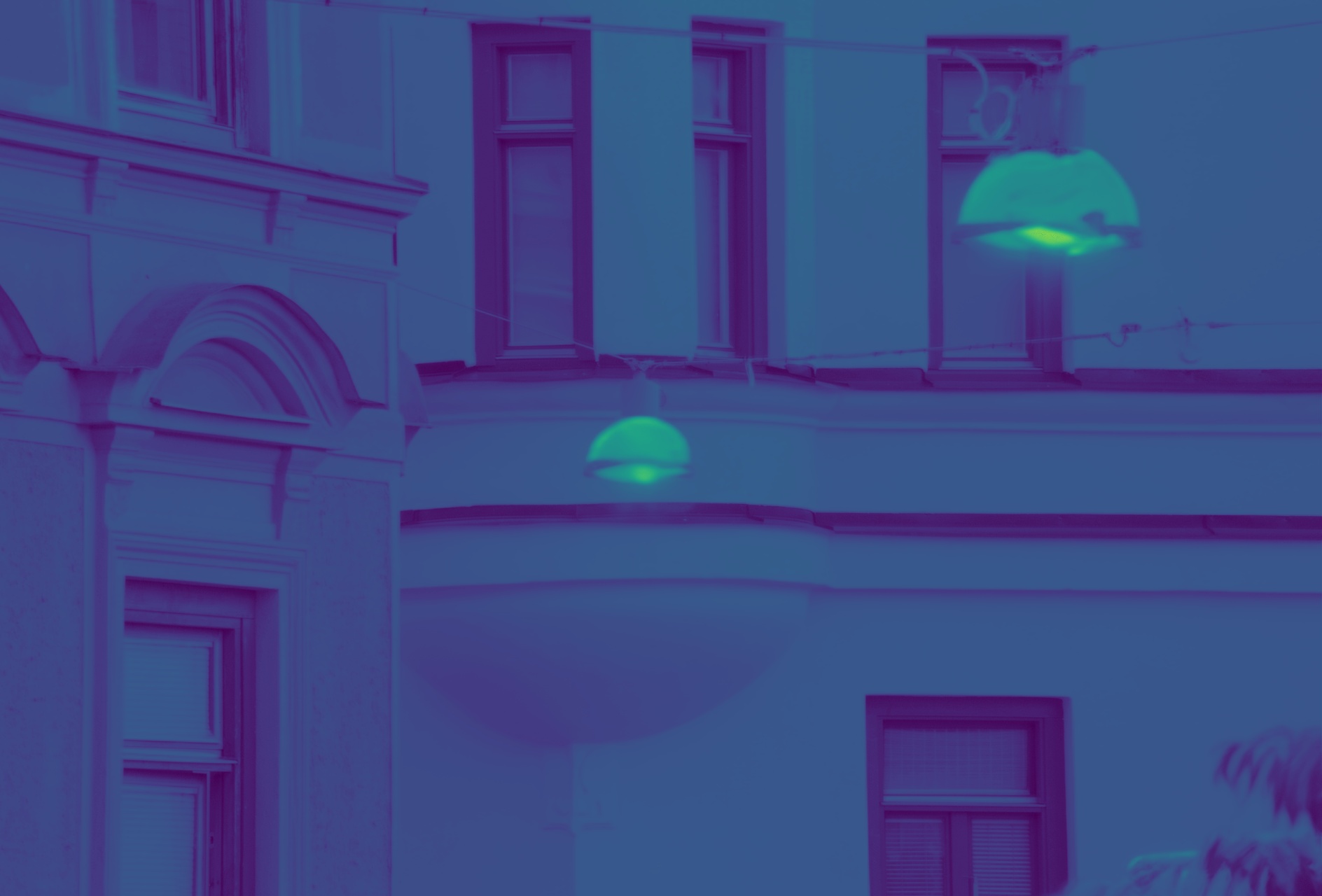}&
    \includegraphics[width=\tmplength]{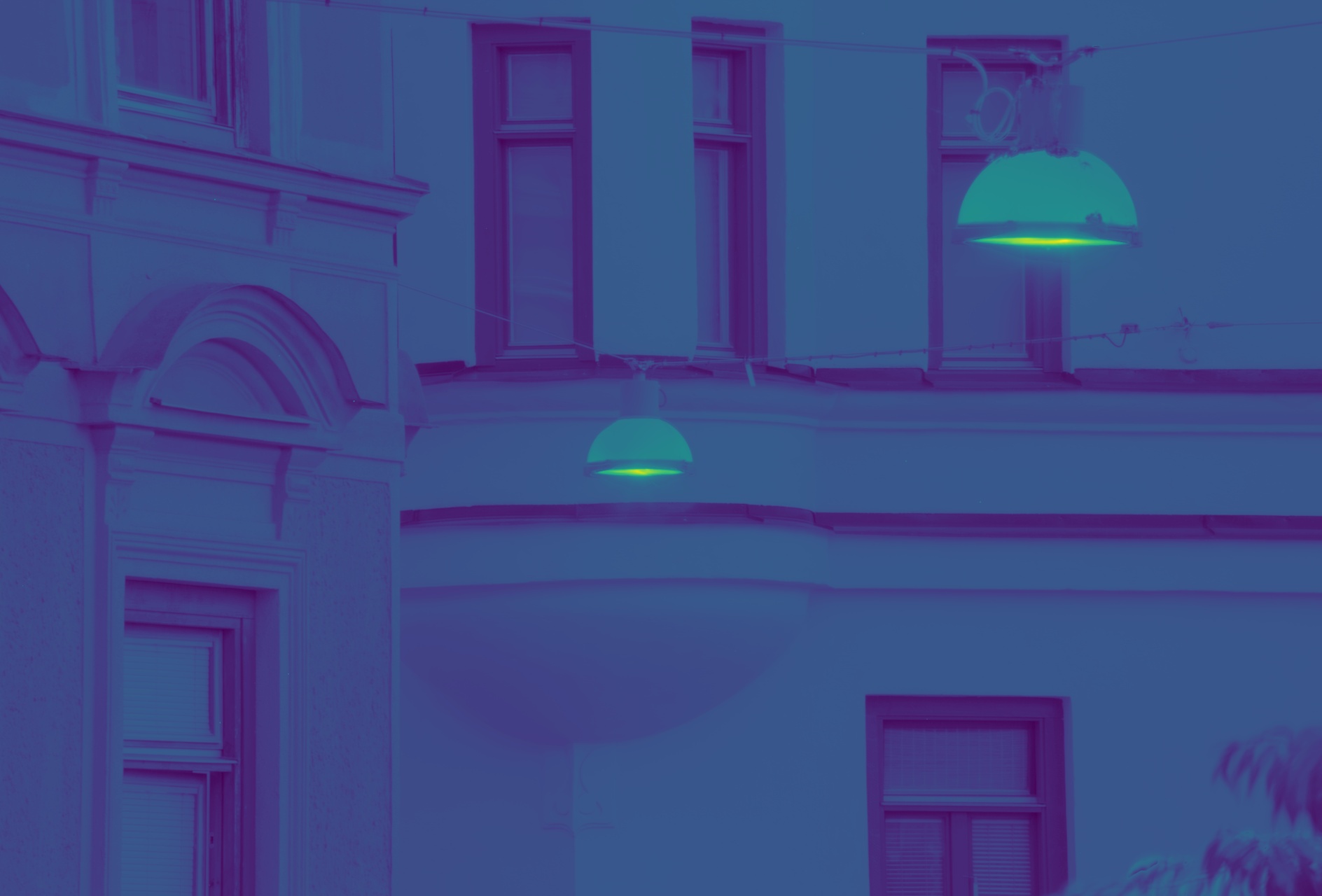} &
    \includegraphics[height=\cbarheight]{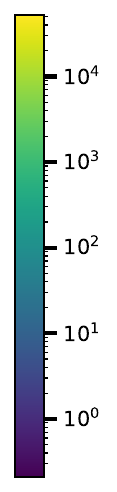}\\
    &
    \includegraphics[width=\tmplength]{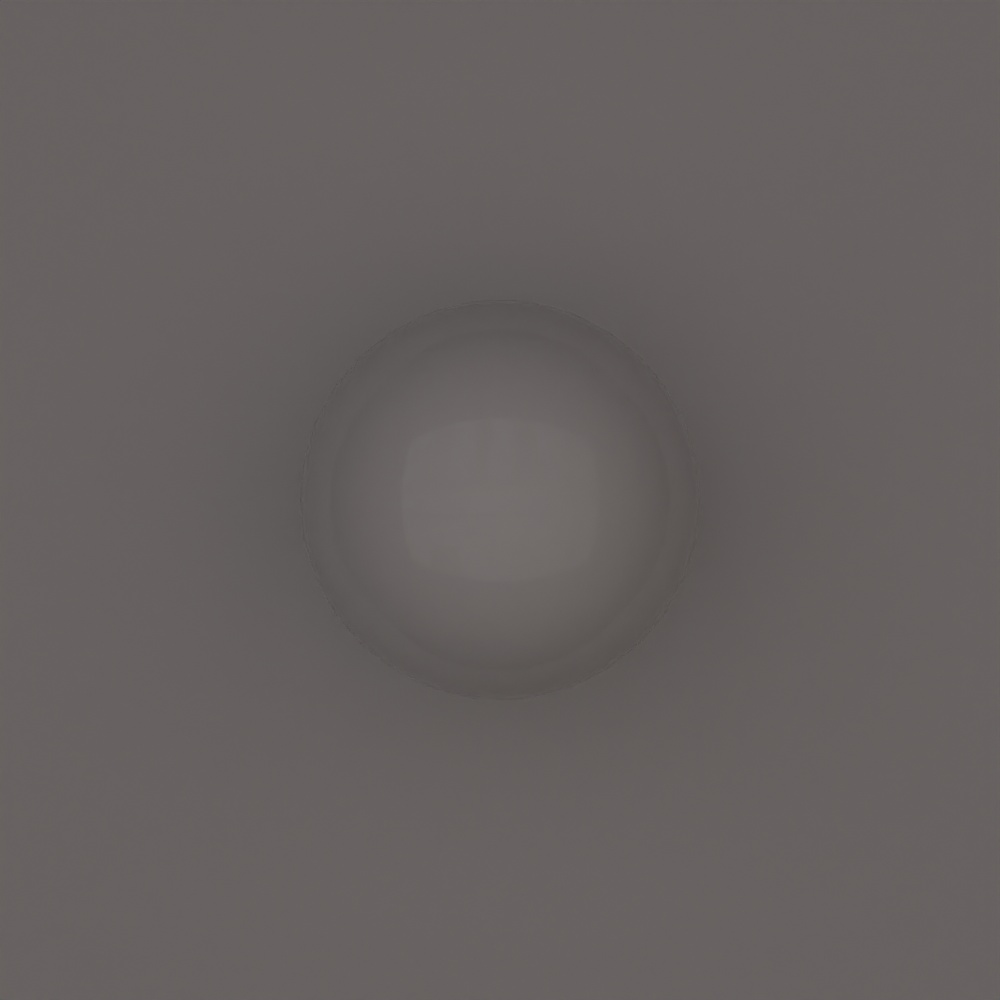}&
    \includegraphics[width=\tmplength]{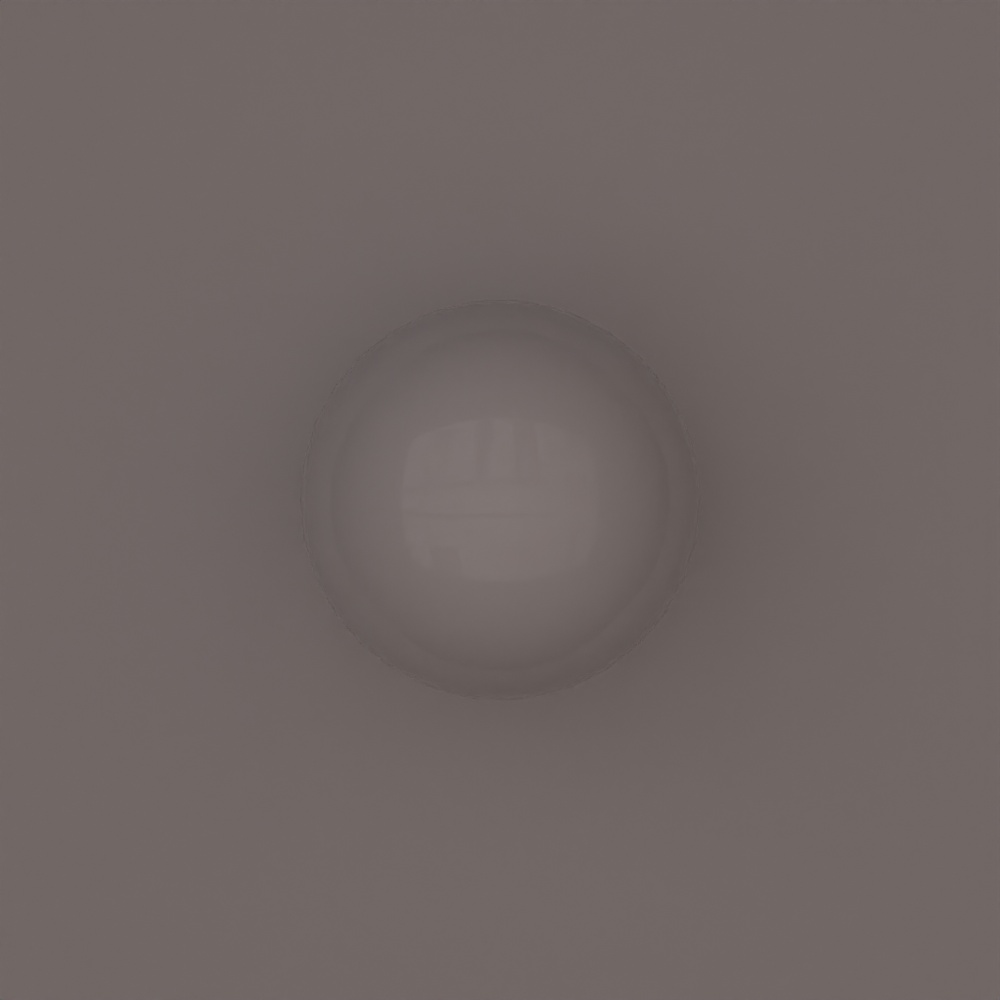}&
    \includegraphics[width=\tmplength]{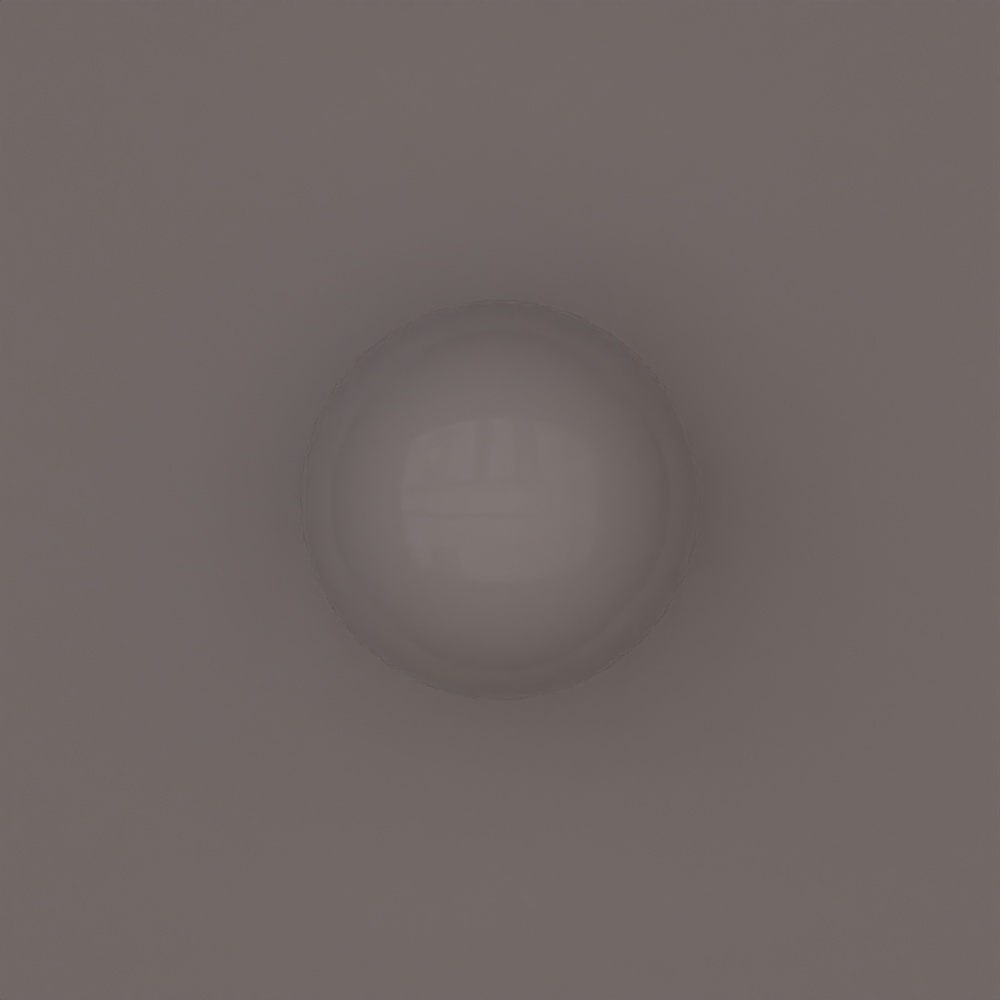}&
    \includegraphics[width=\tmplength]{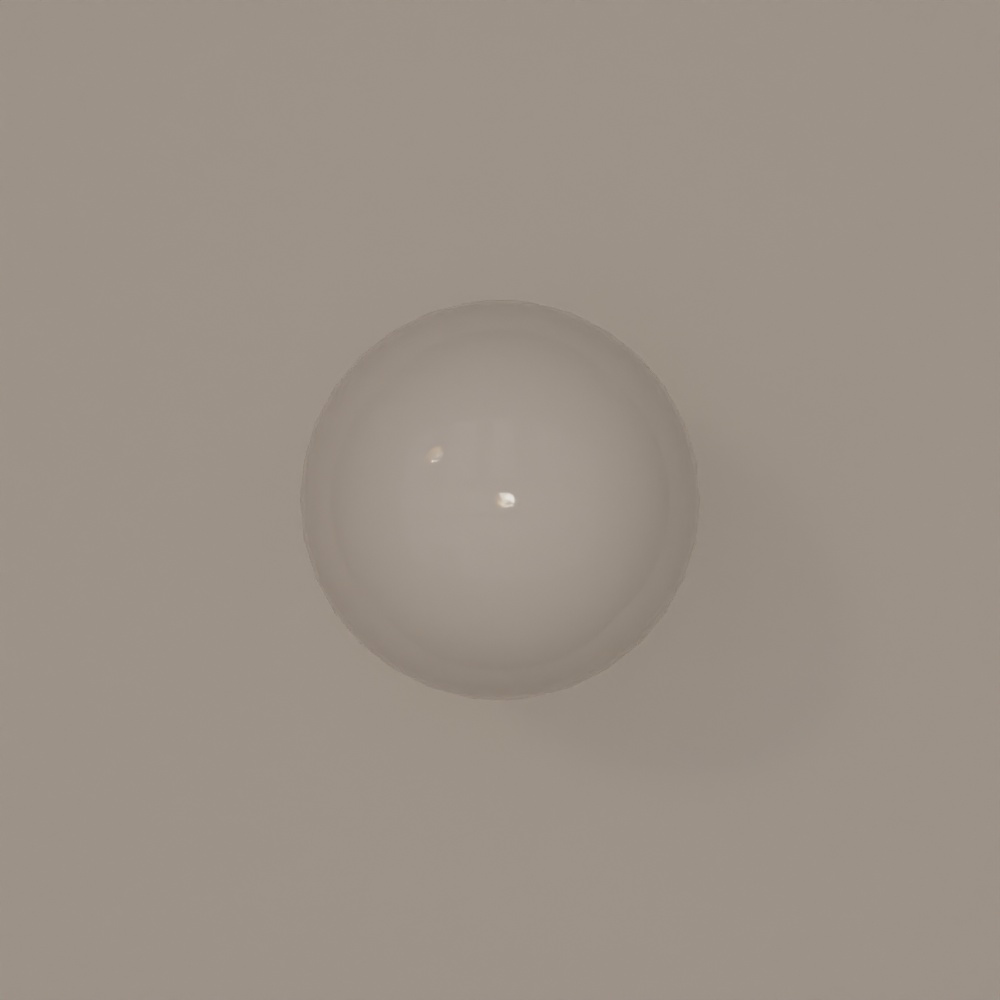}&
    \includegraphics[width=\tmplength]{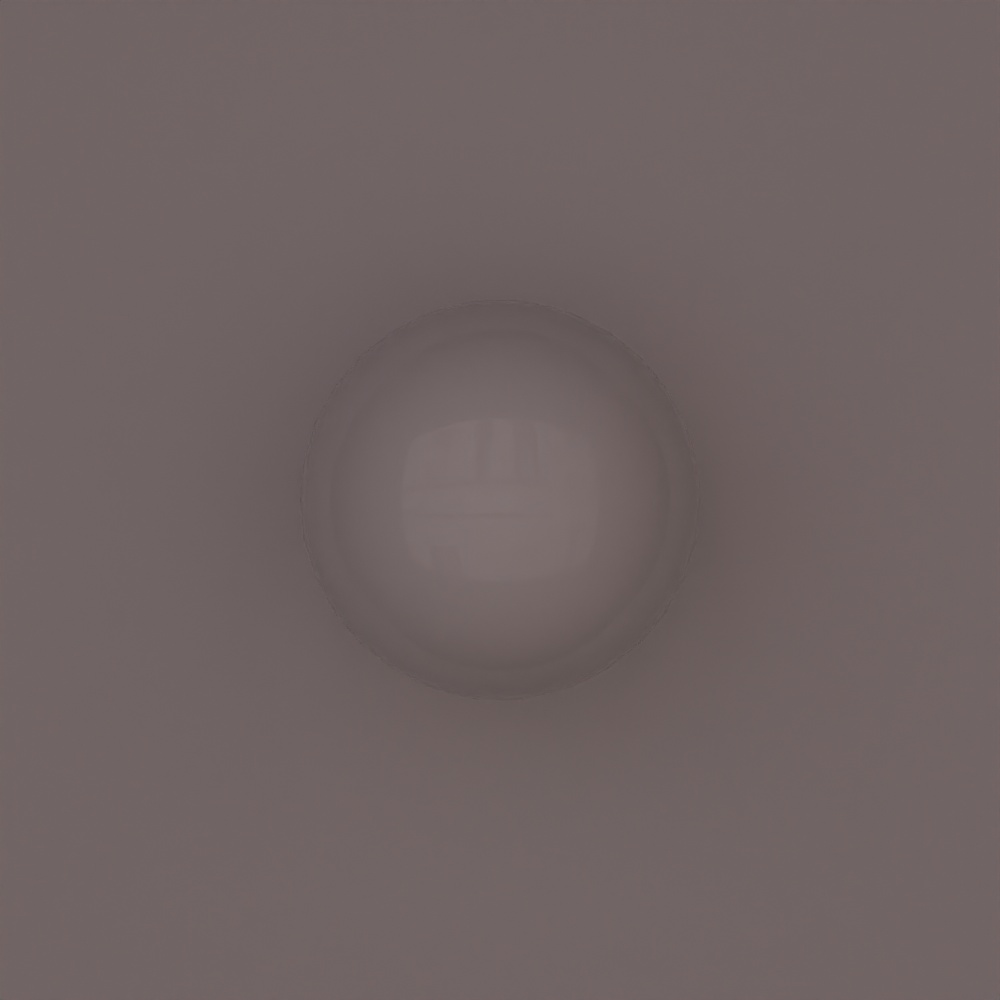}&
    \includegraphics[width=\tmplength]{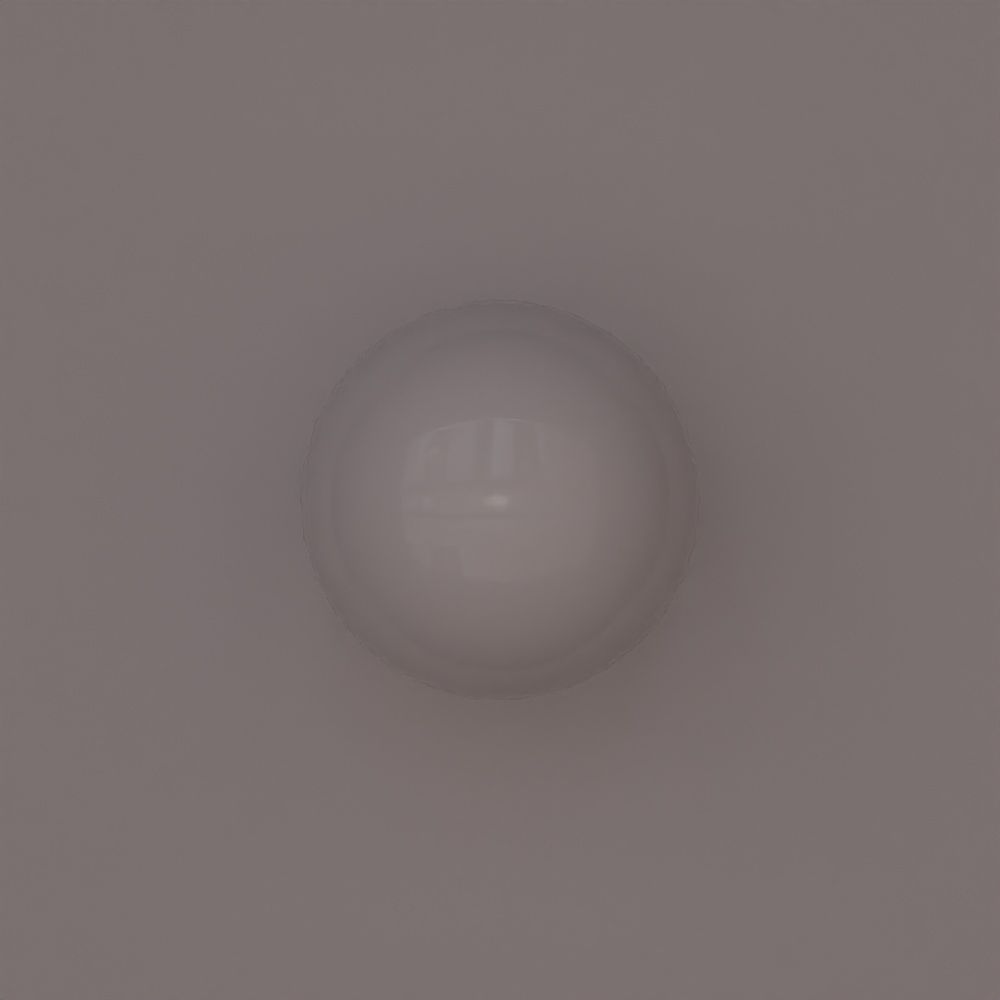}&
    \includegraphics[width=\tmplength]{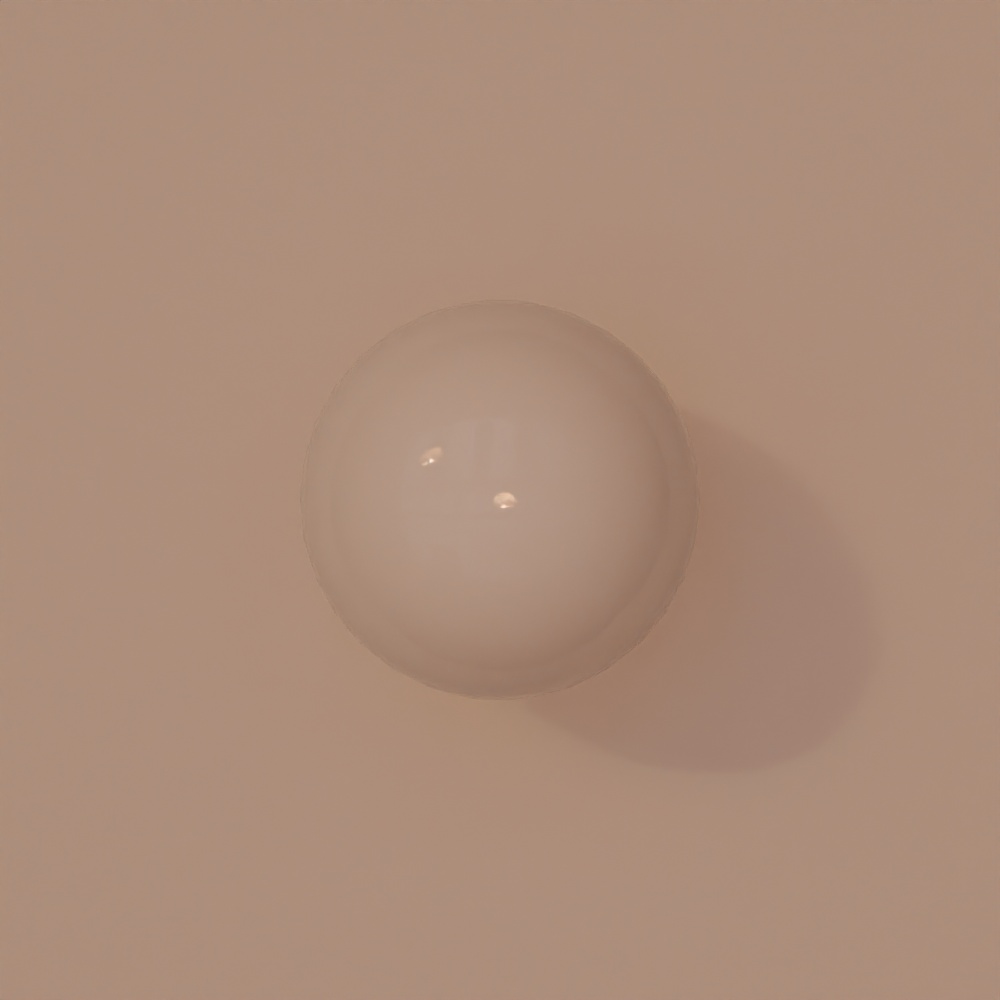}&
    \includegraphics[width=\tmplength]{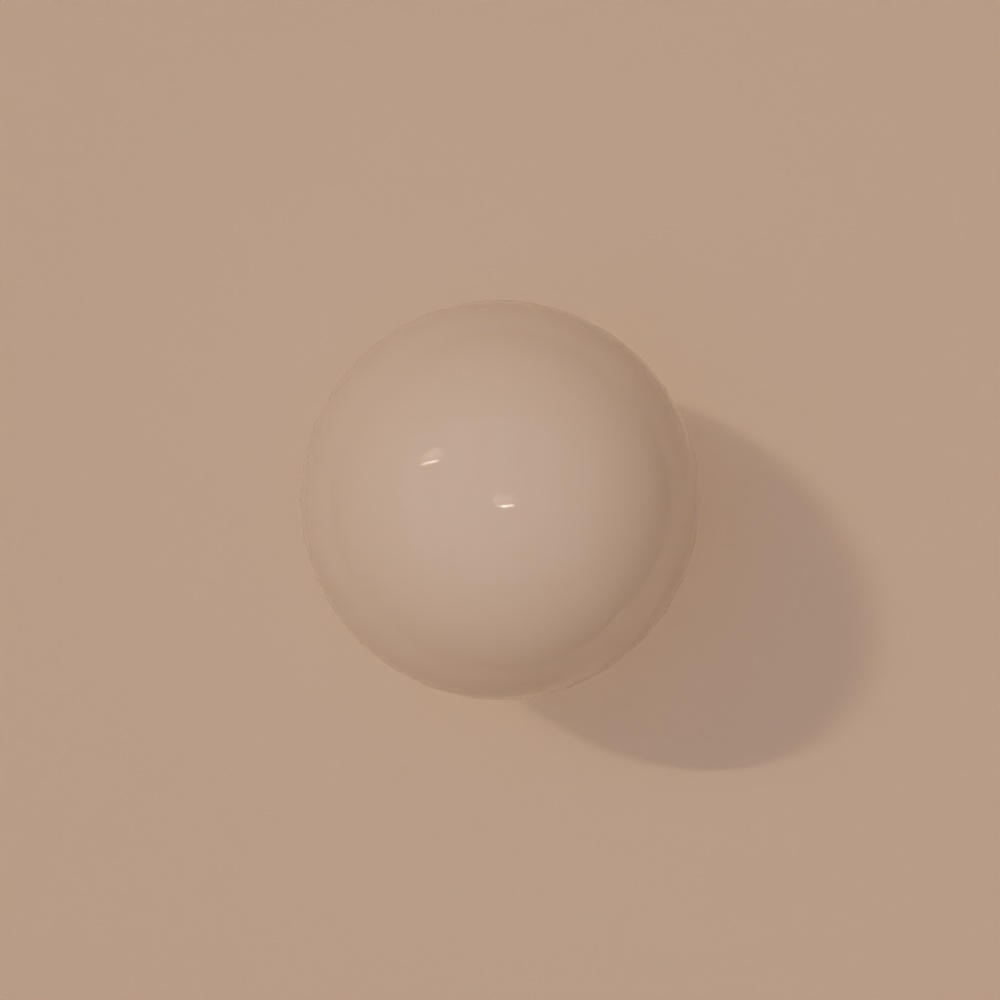} & \\
    \includegraphics[width=\tmplength]{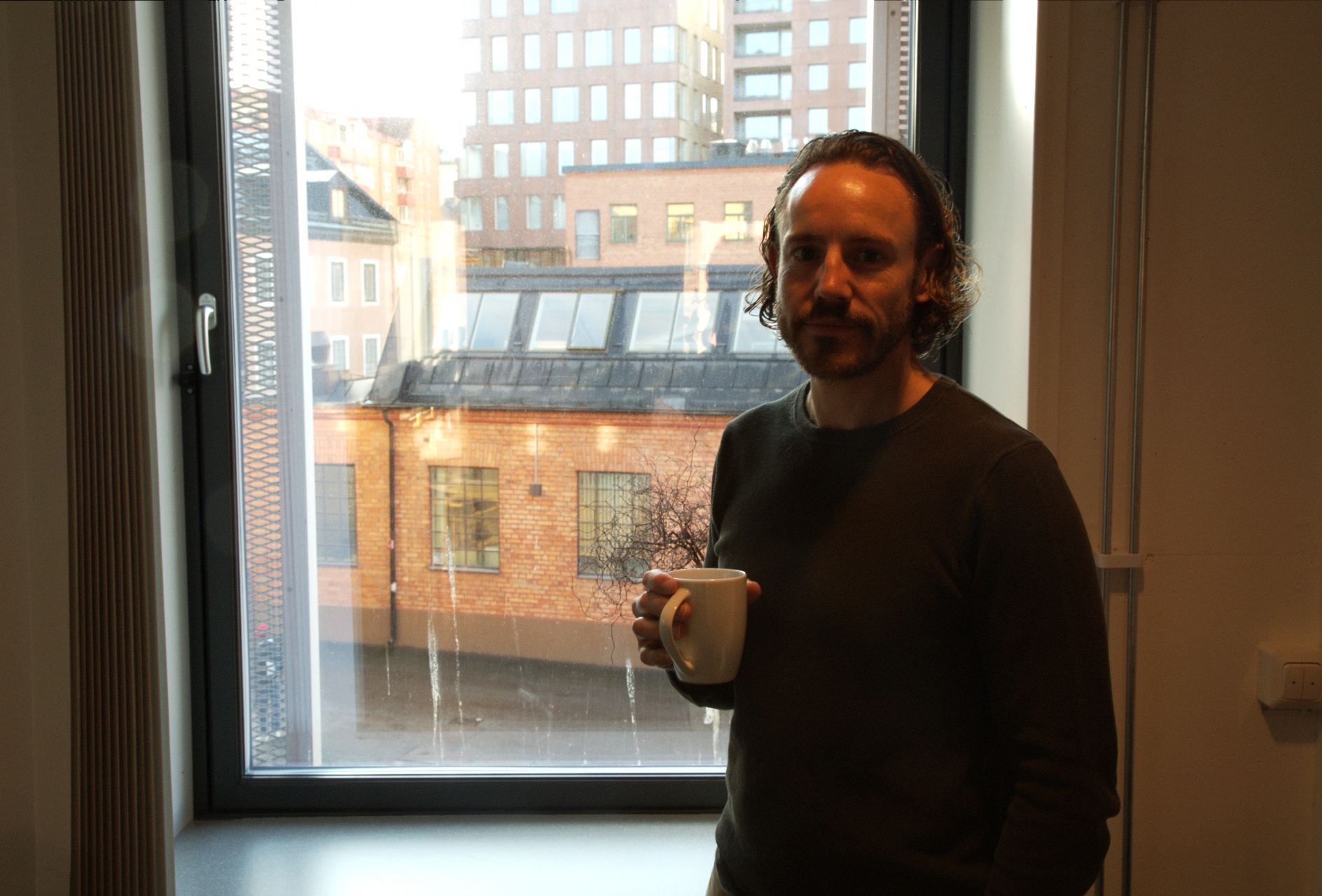}&
    \includegraphics[width=\tmplength]{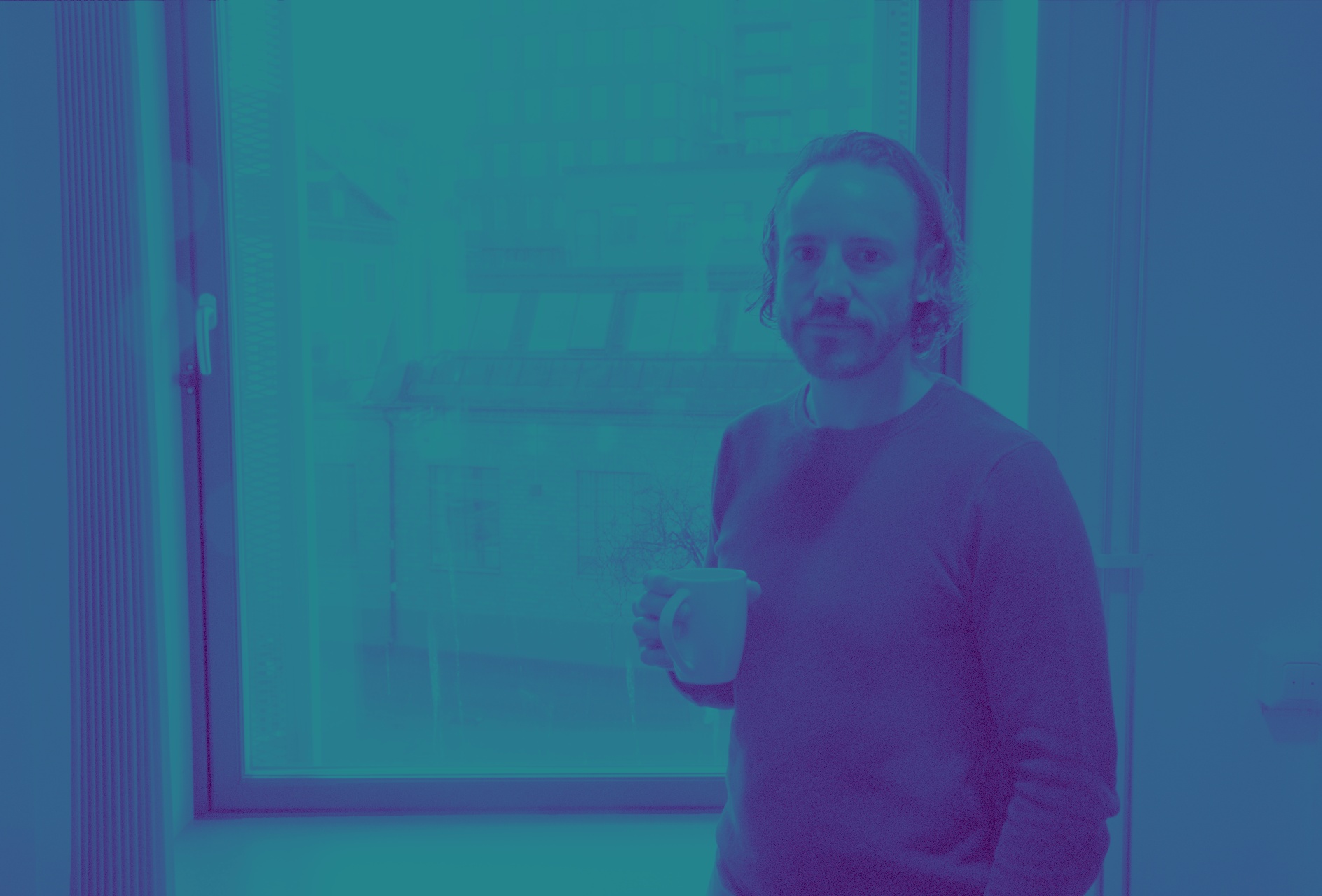}&
    \includegraphics[width=\tmplength]{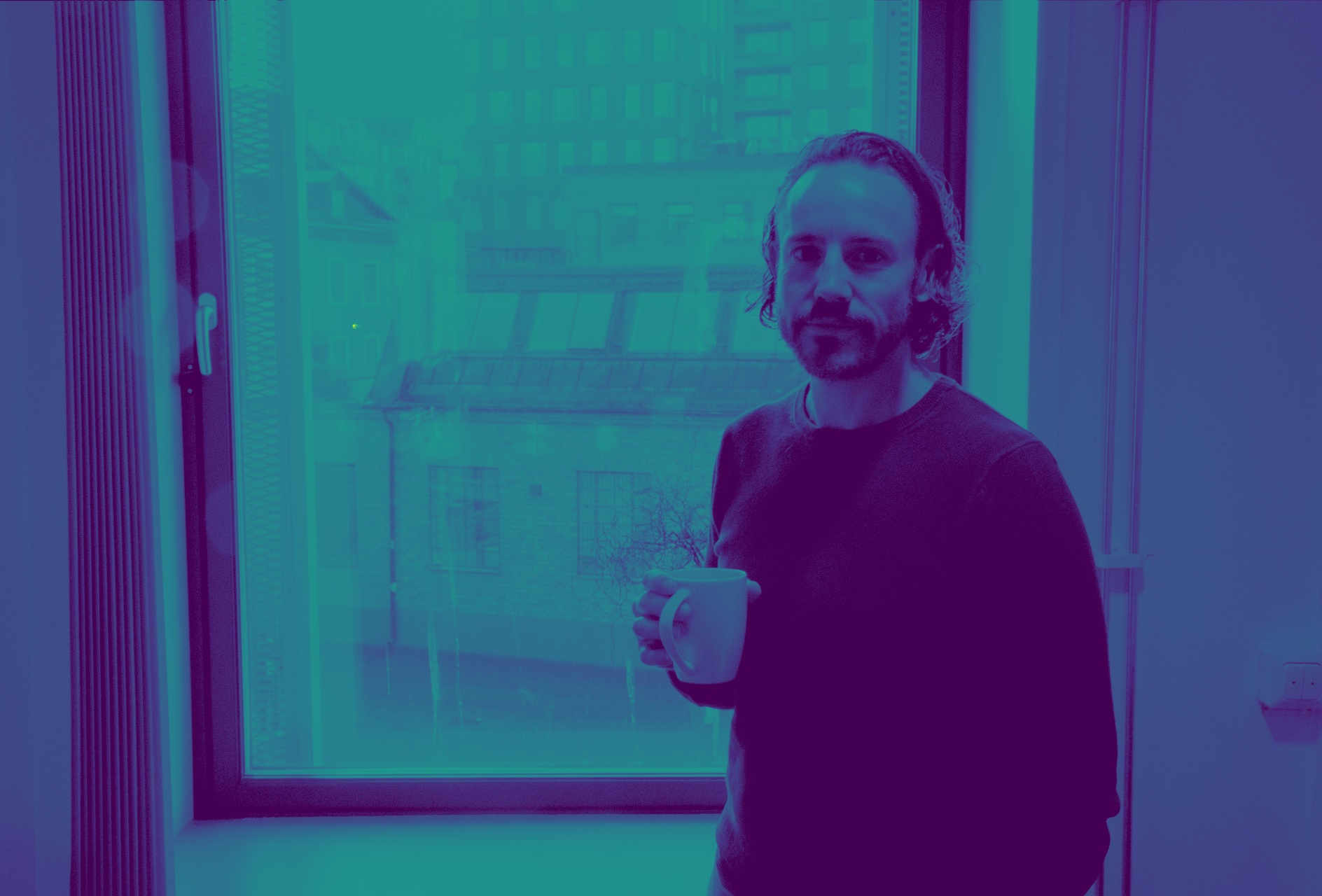}&
    \includegraphics[width=\tmplength]{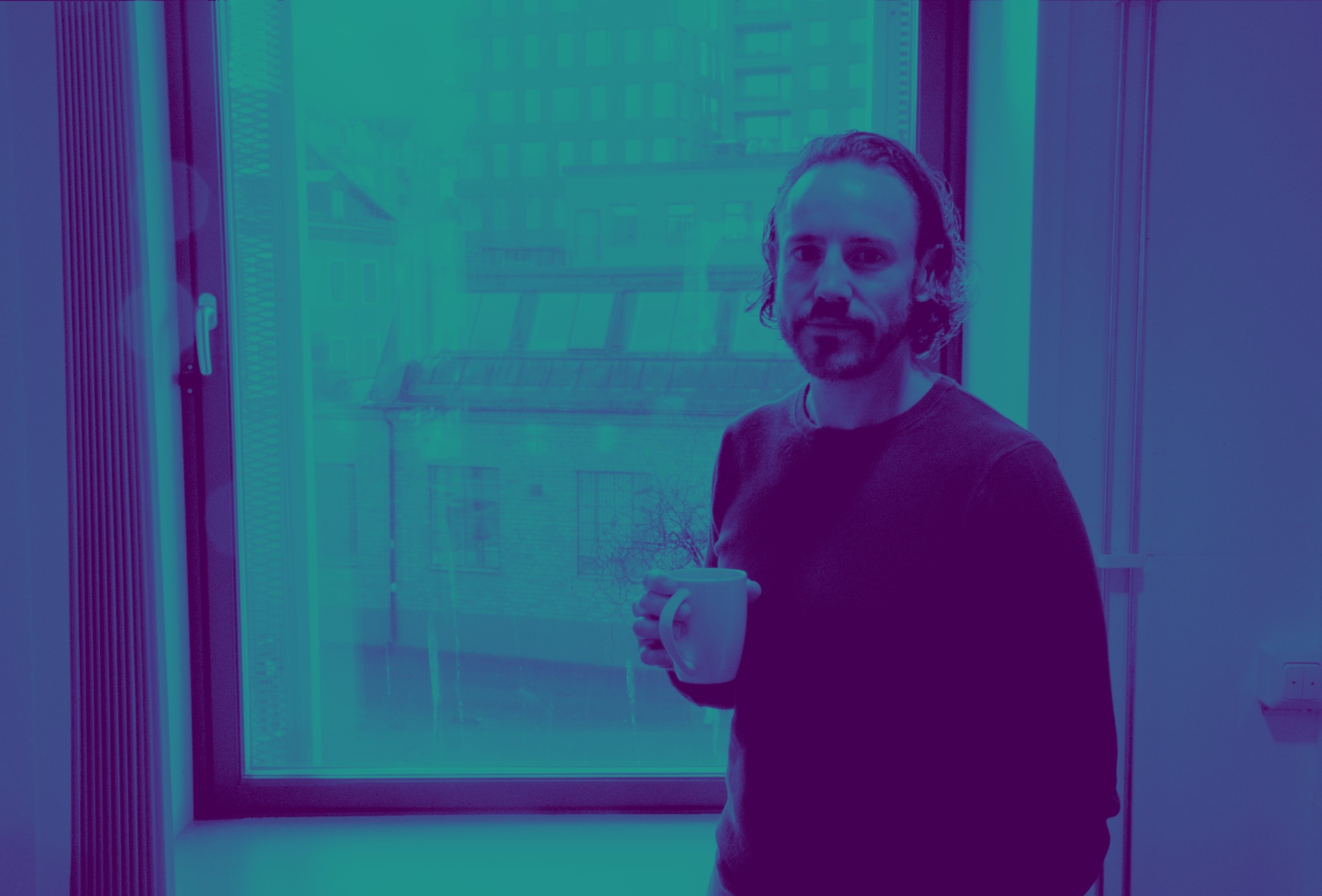}&
    \includegraphics[width=\tmplength]{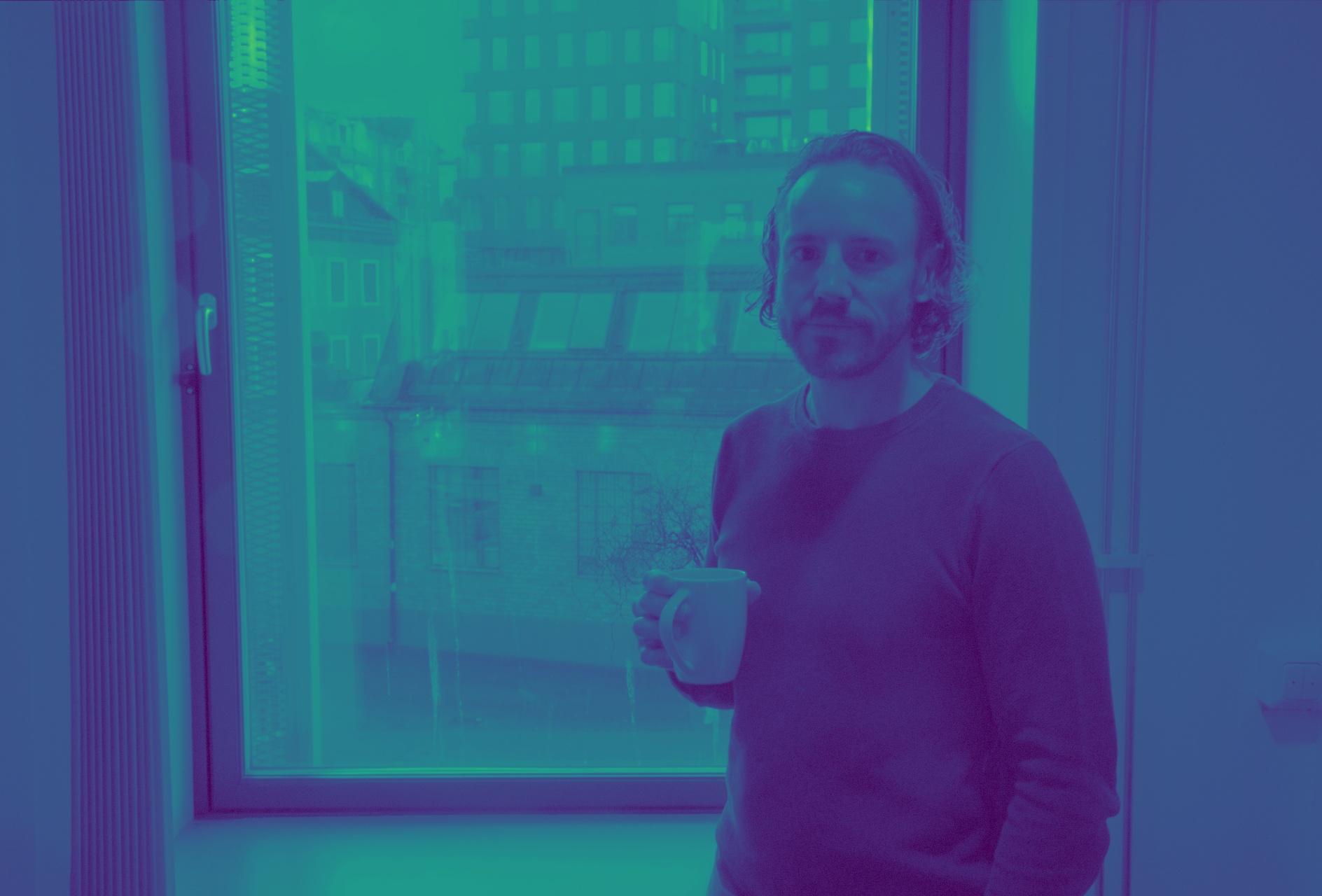}&
    \includegraphics[width=\tmplength]{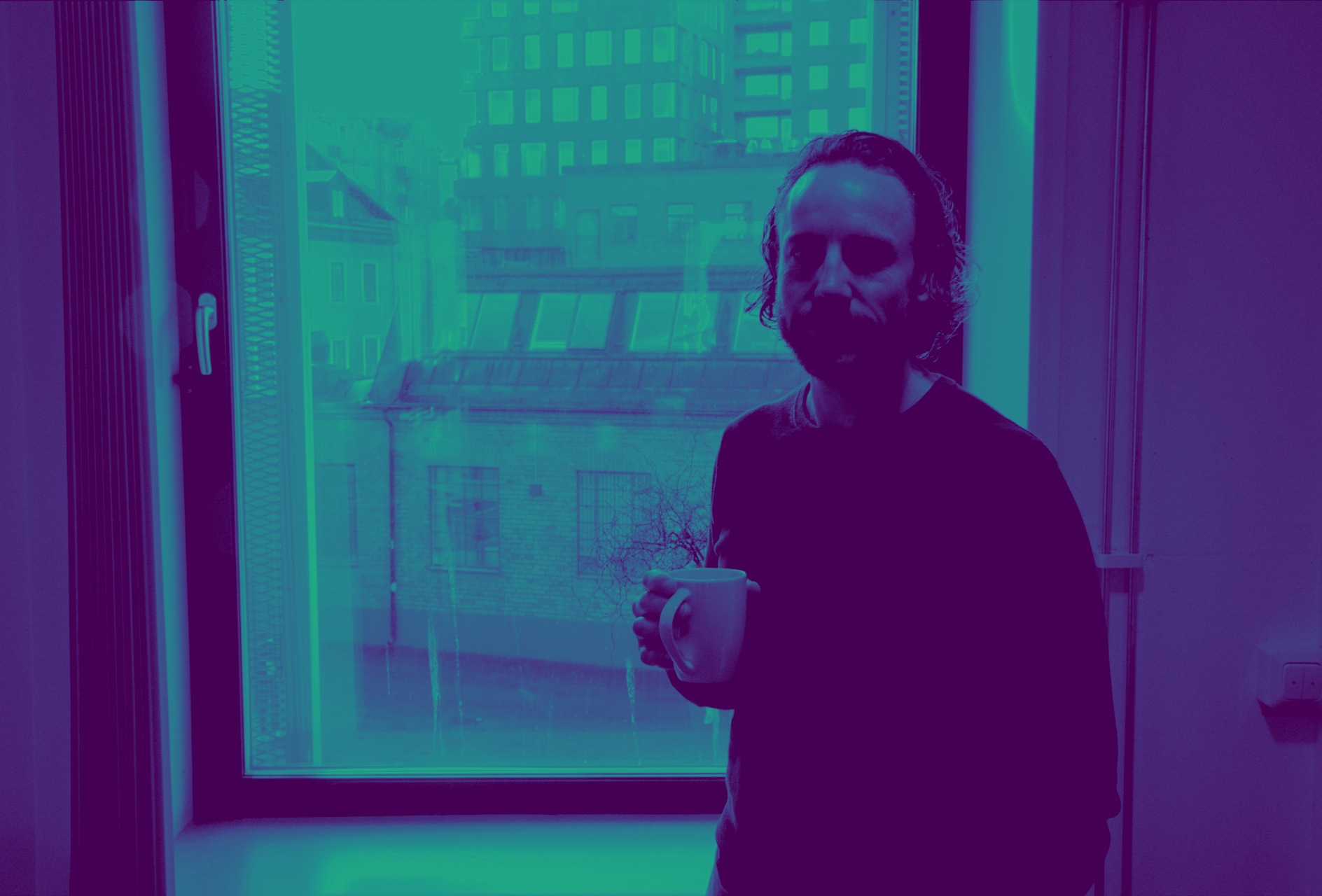}&
    \includegraphics[width=\tmplength]{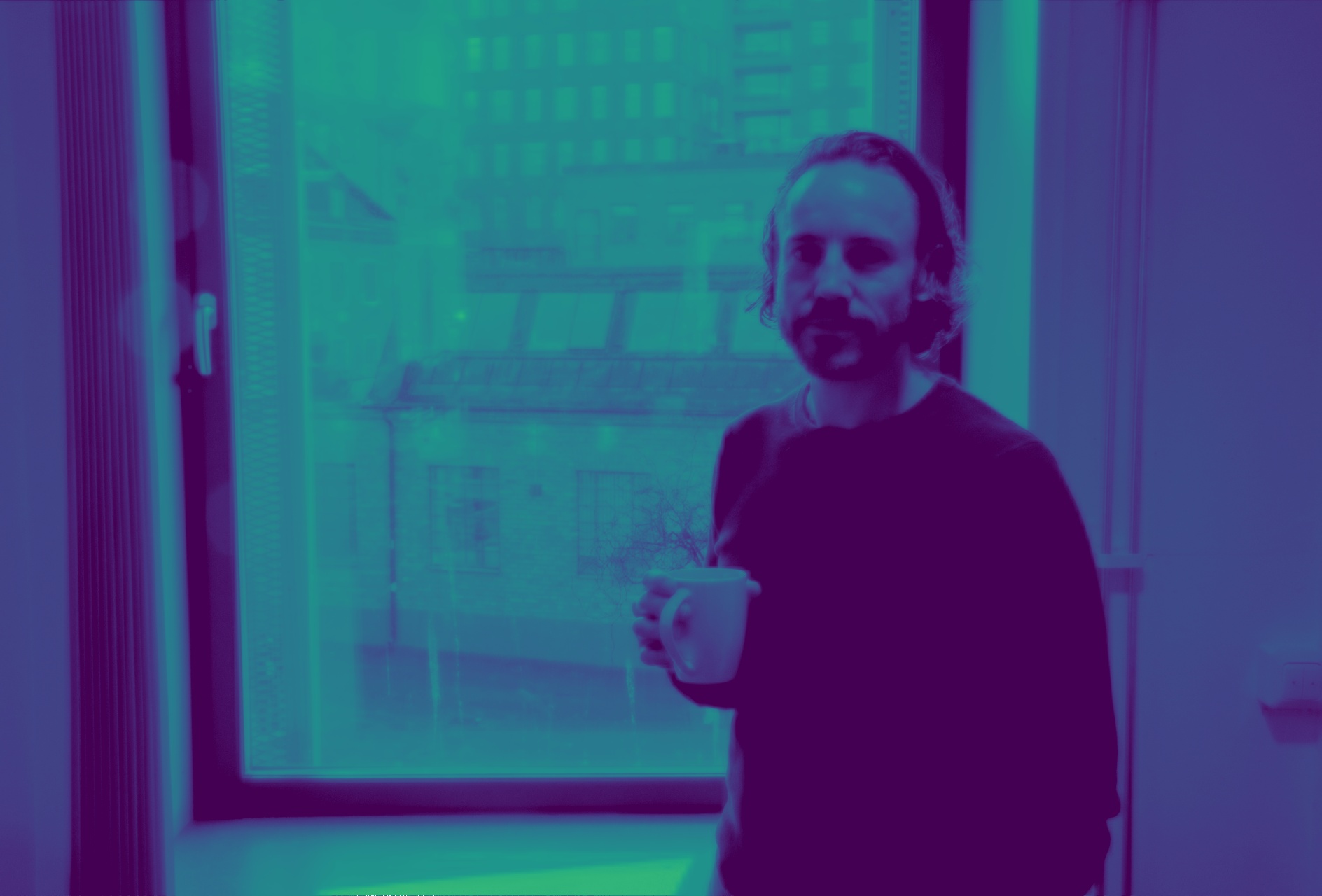}&
    \includegraphics[width=\tmplength]{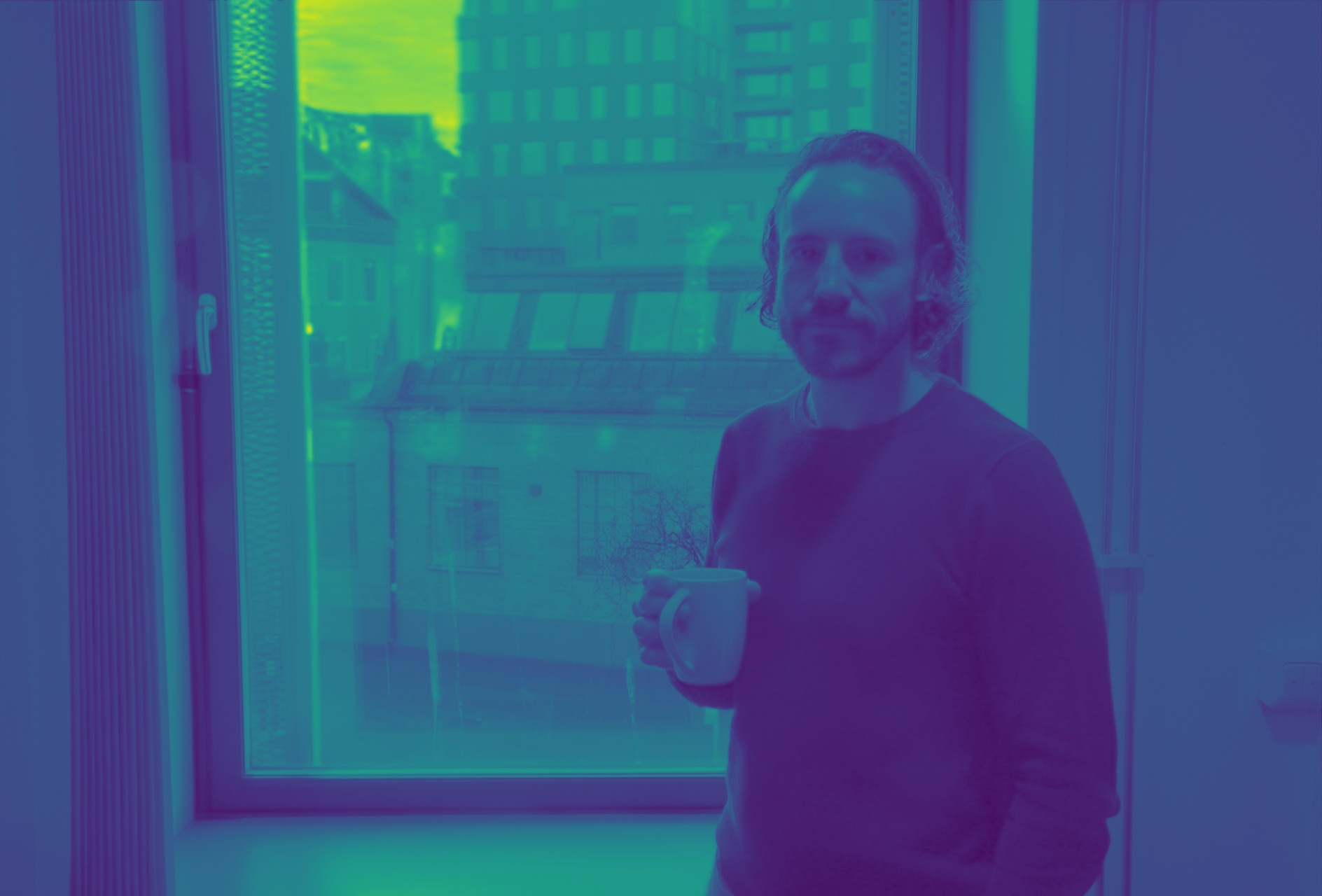}&
    \includegraphics[width=\tmplength]{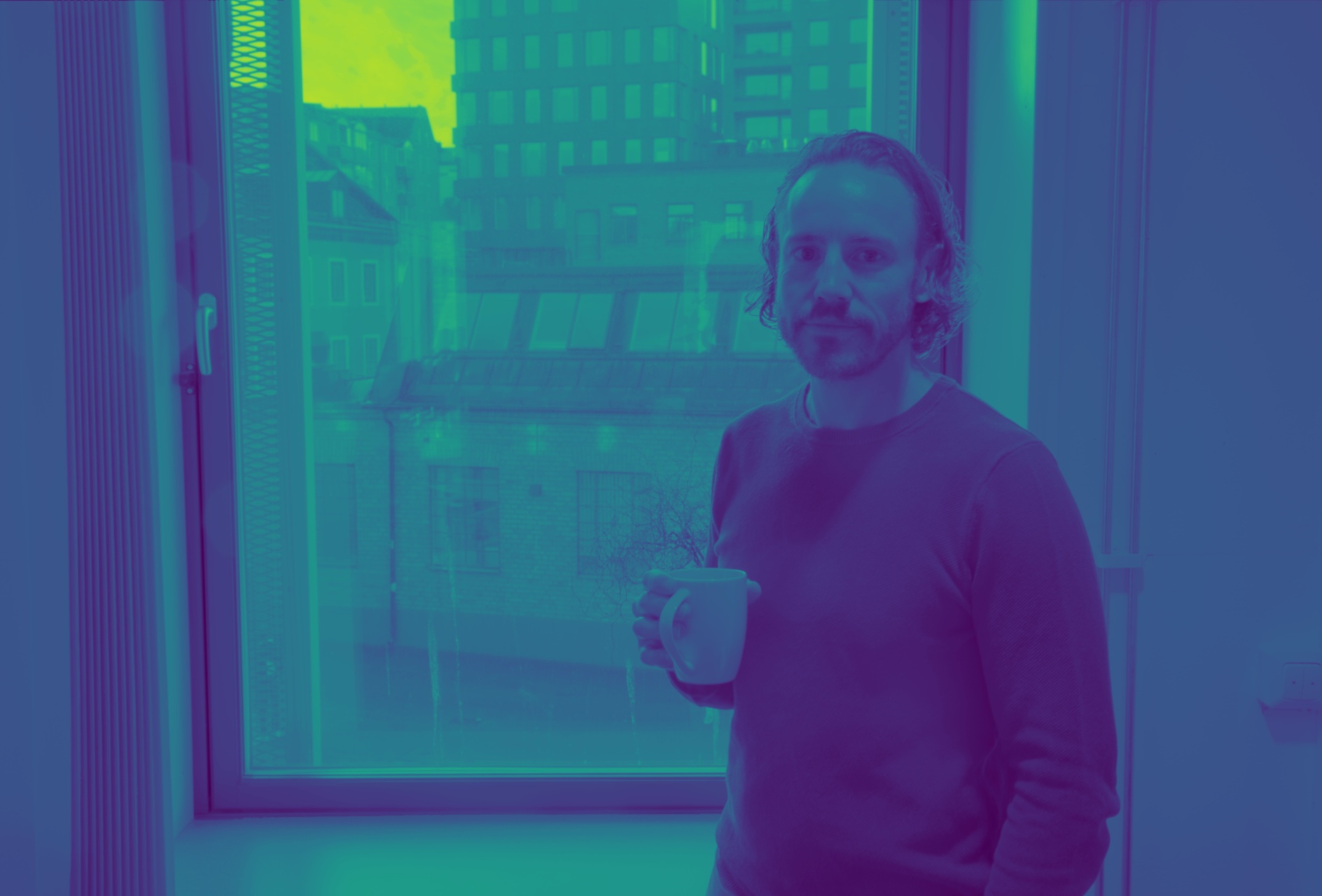} &
    \includegraphics[height=\cbarheight]{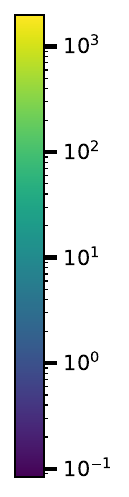}\\
    &
    \includegraphics[width=\tmplength]{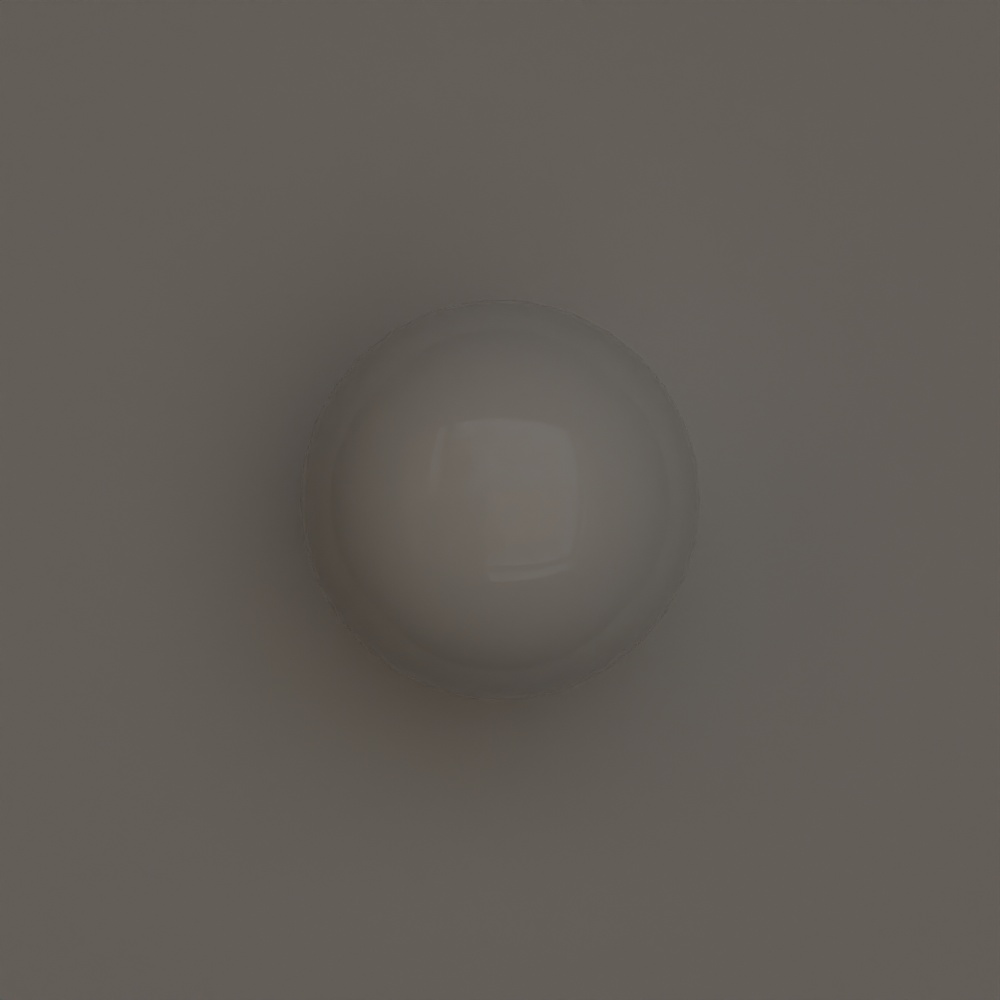}&
    \includegraphics[width=\tmplength]{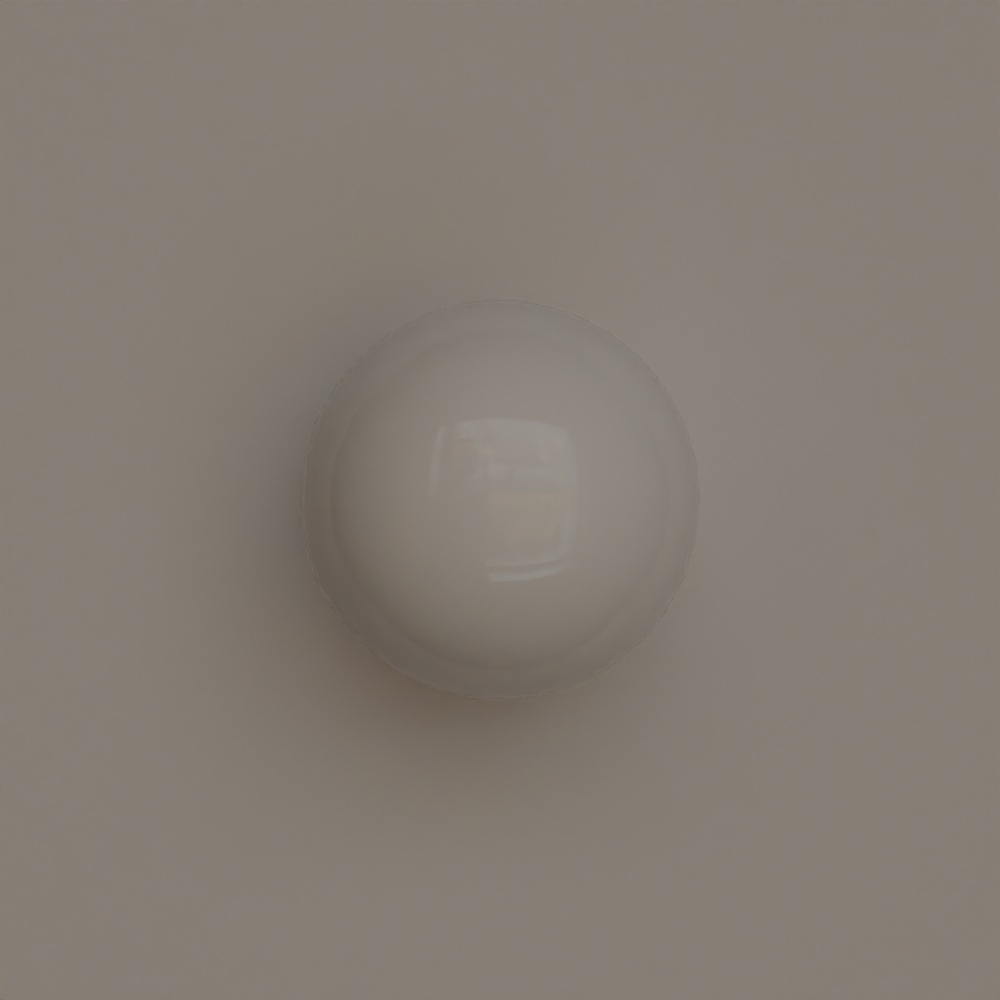}&
    \includegraphics[width=\tmplength]{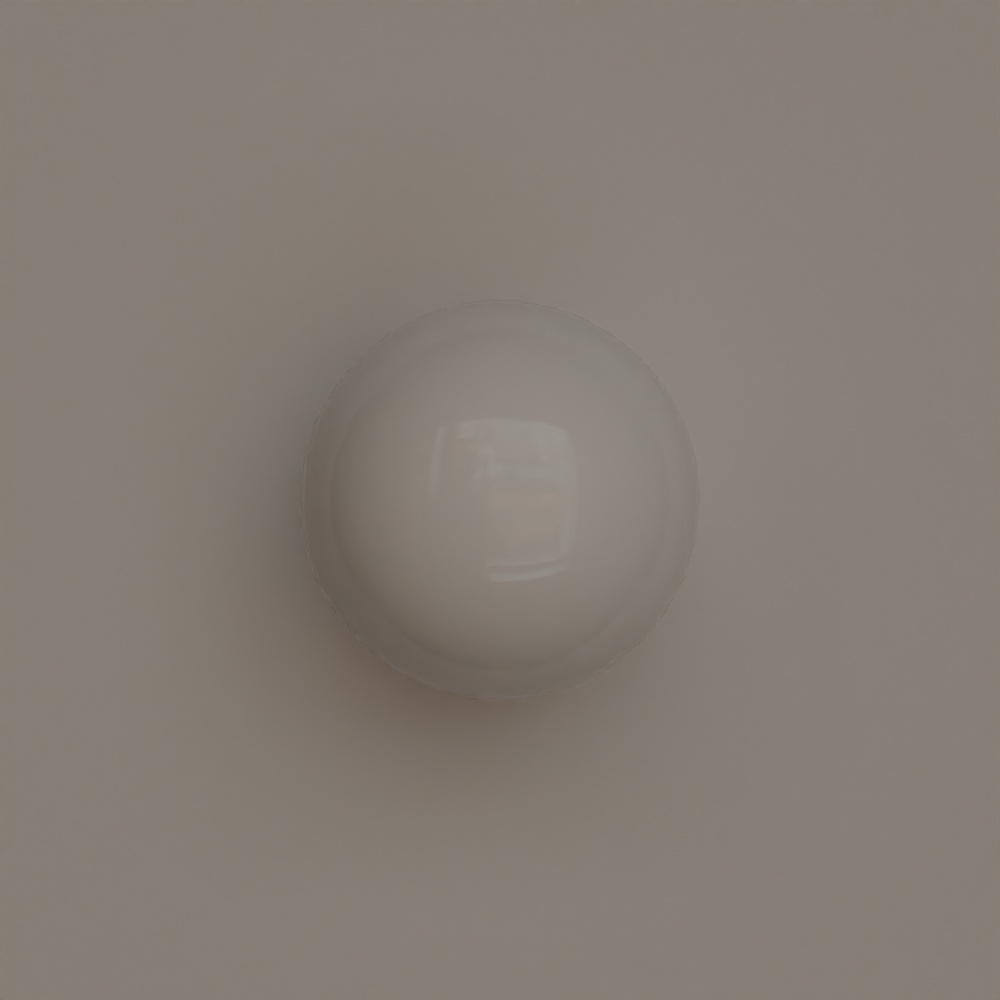}&
    \includegraphics[width=\tmplength]{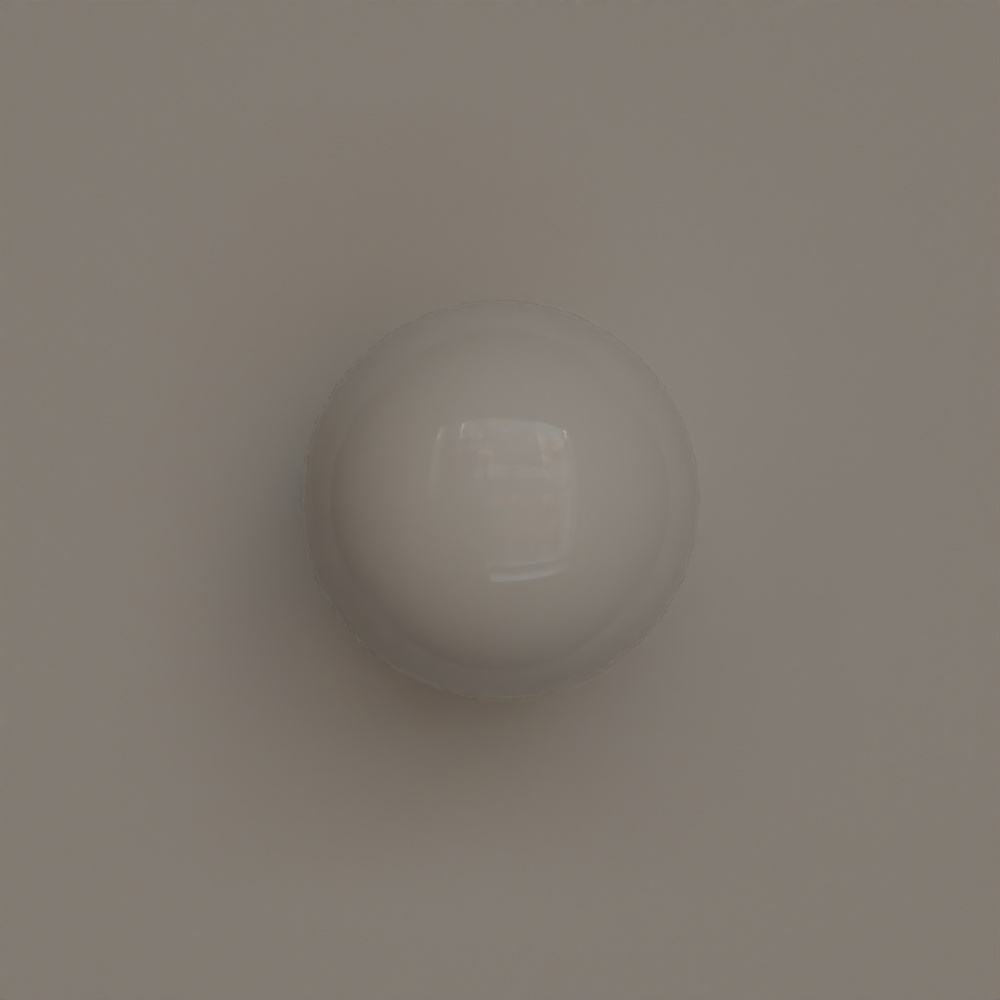}&
    \includegraphics[width=\tmplength]{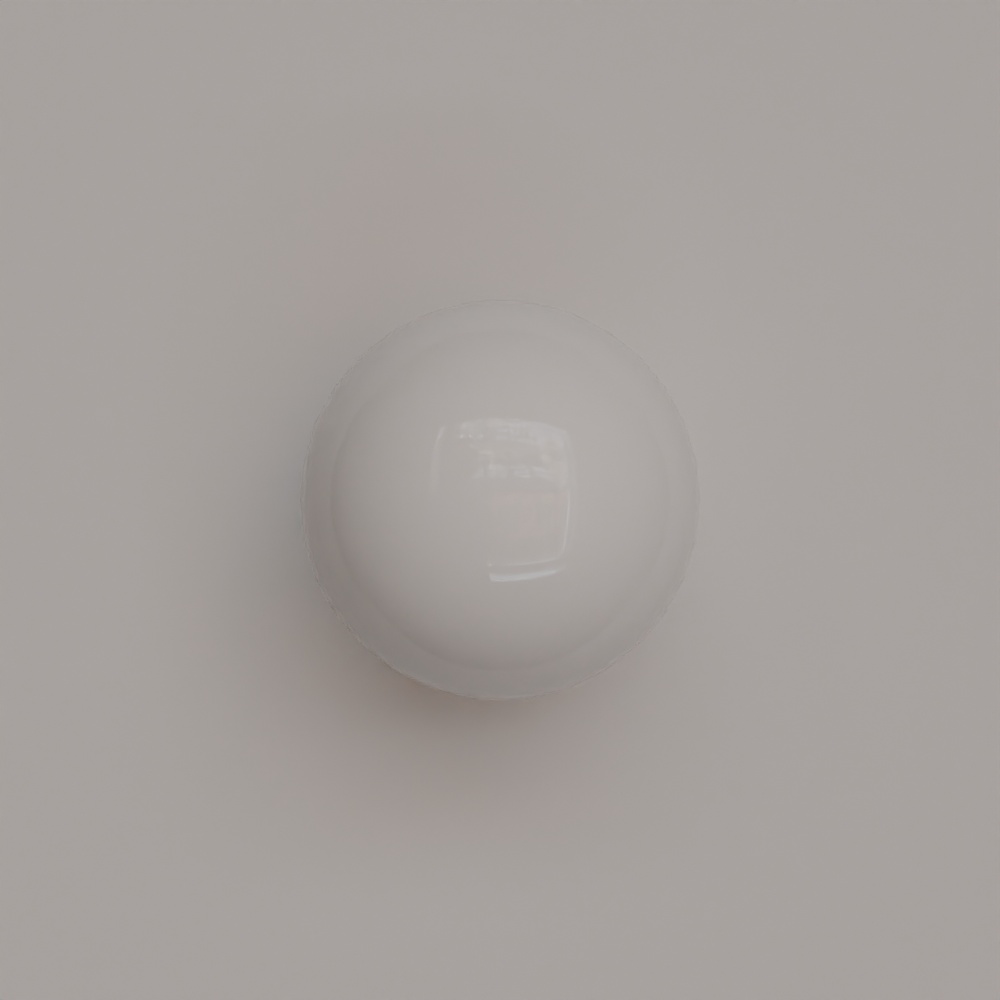}&
    \includegraphics[width=\tmplength]{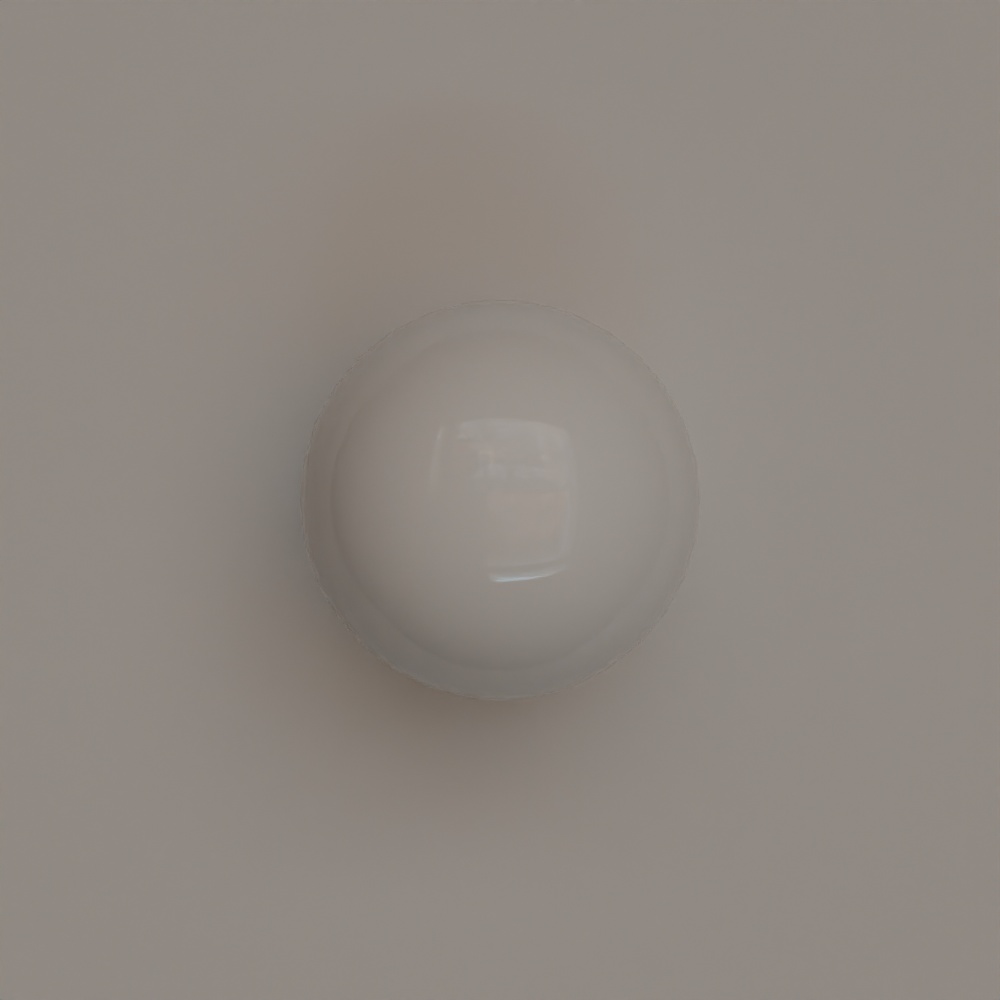}&
    \includegraphics[width=\tmplength]{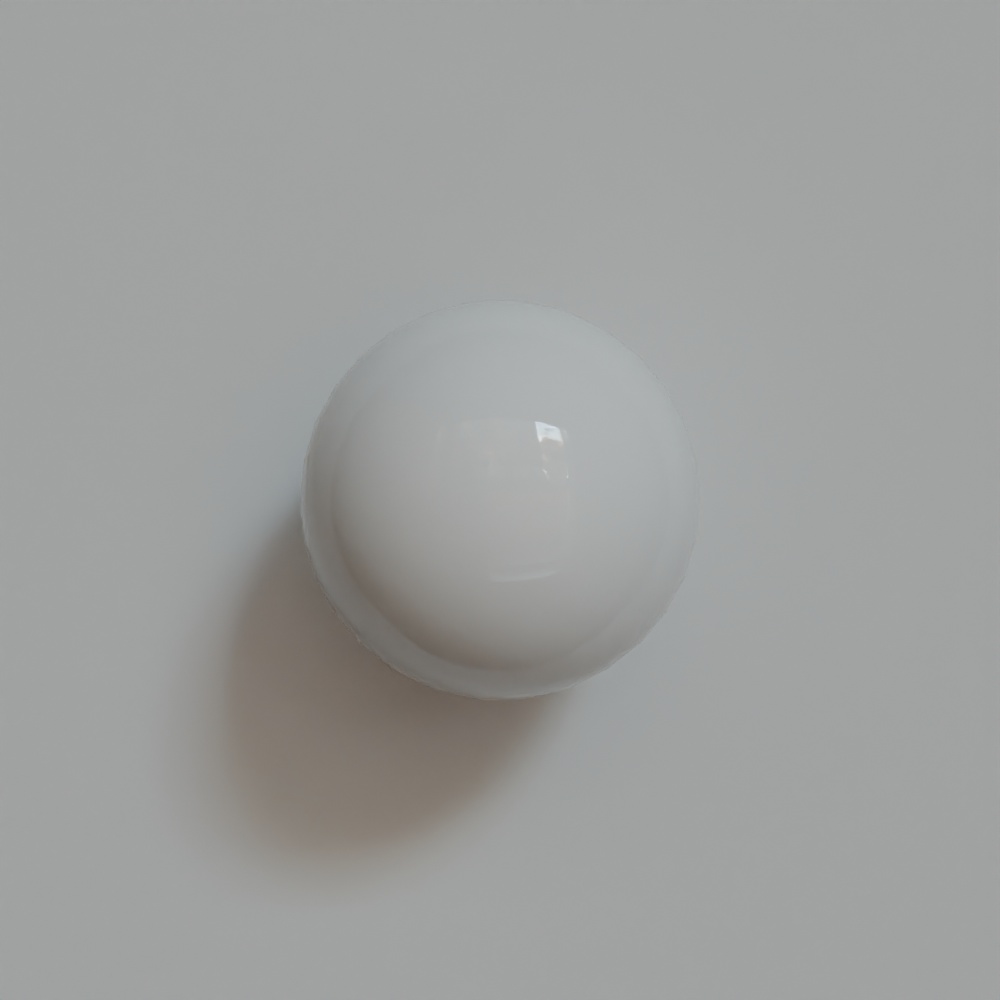}&
    \includegraphics[width=\tmplength]{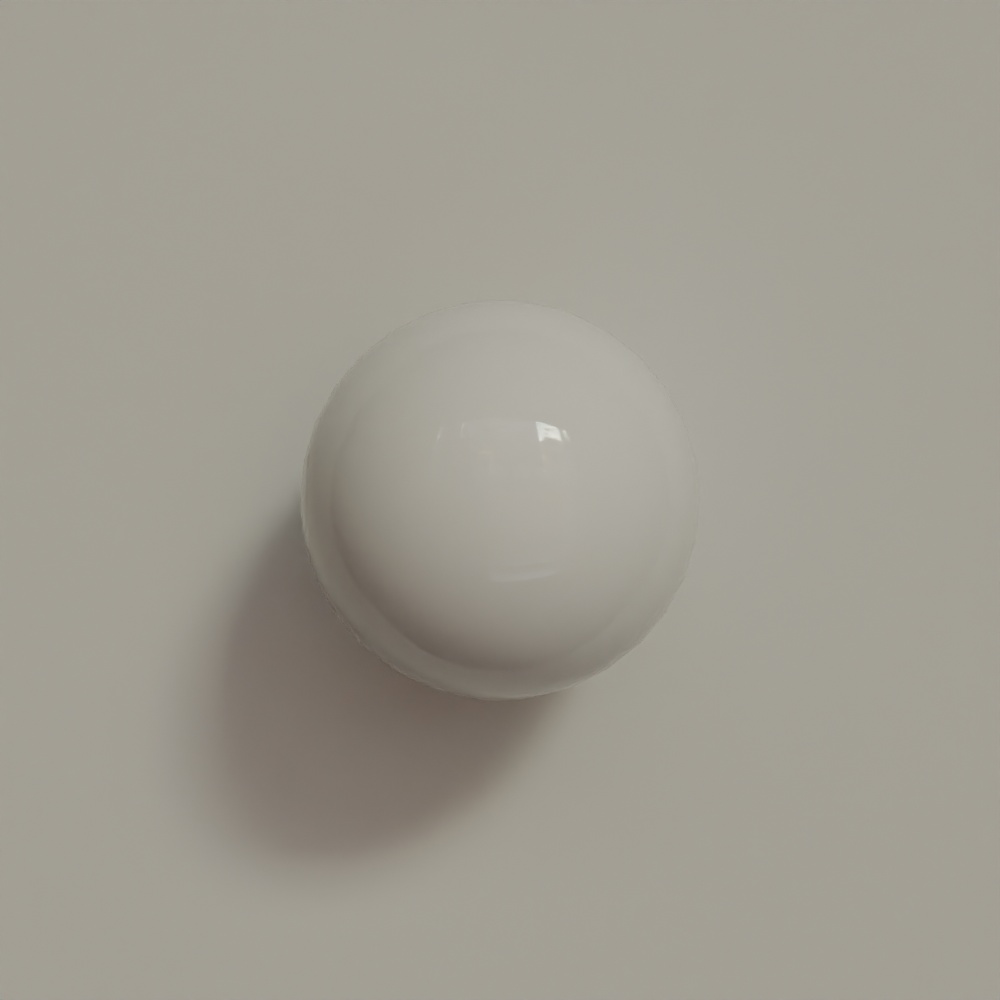} & \\
    \includegraphics[width=\tmplength]{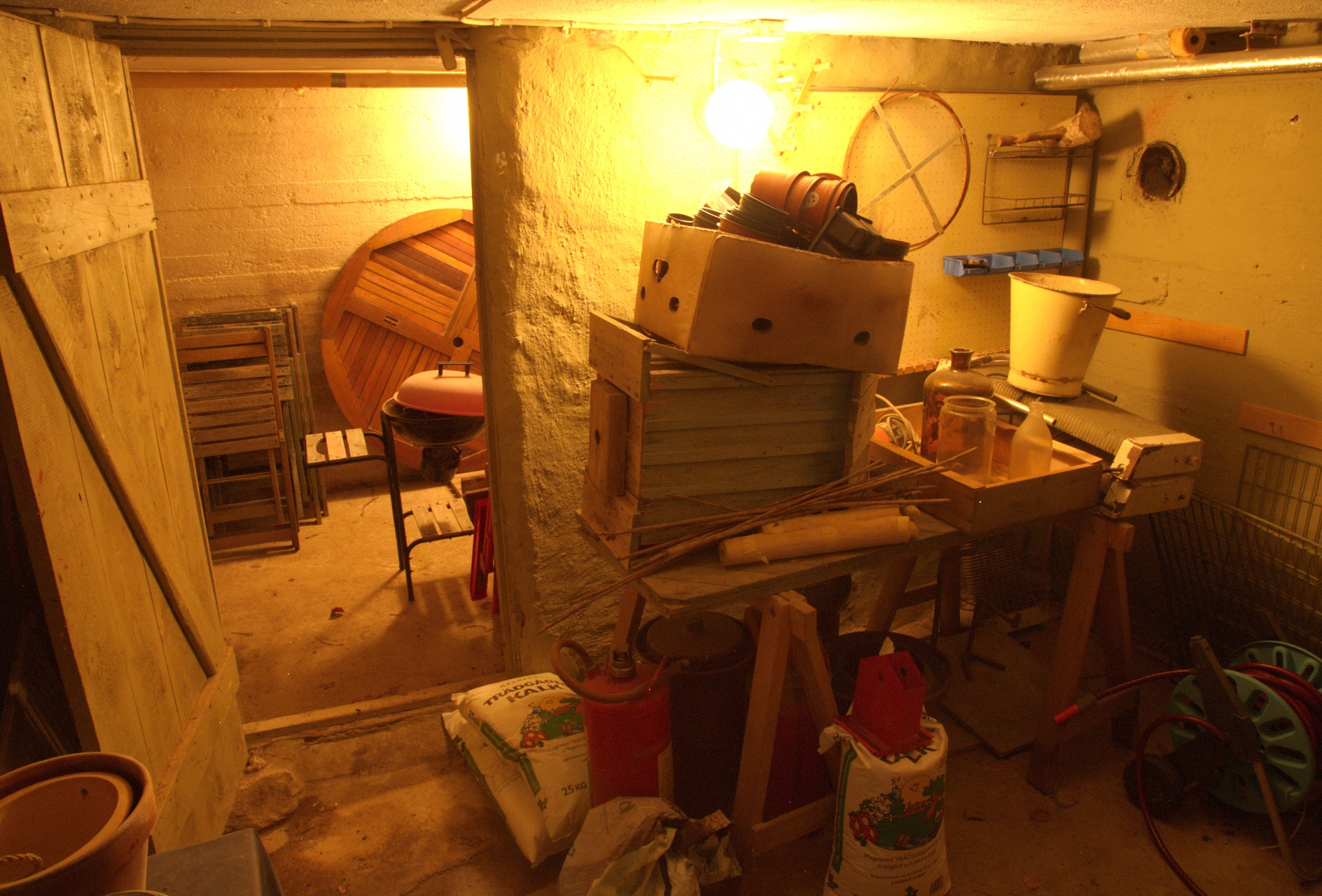}&
    \includegraphics[width=\tmplength]{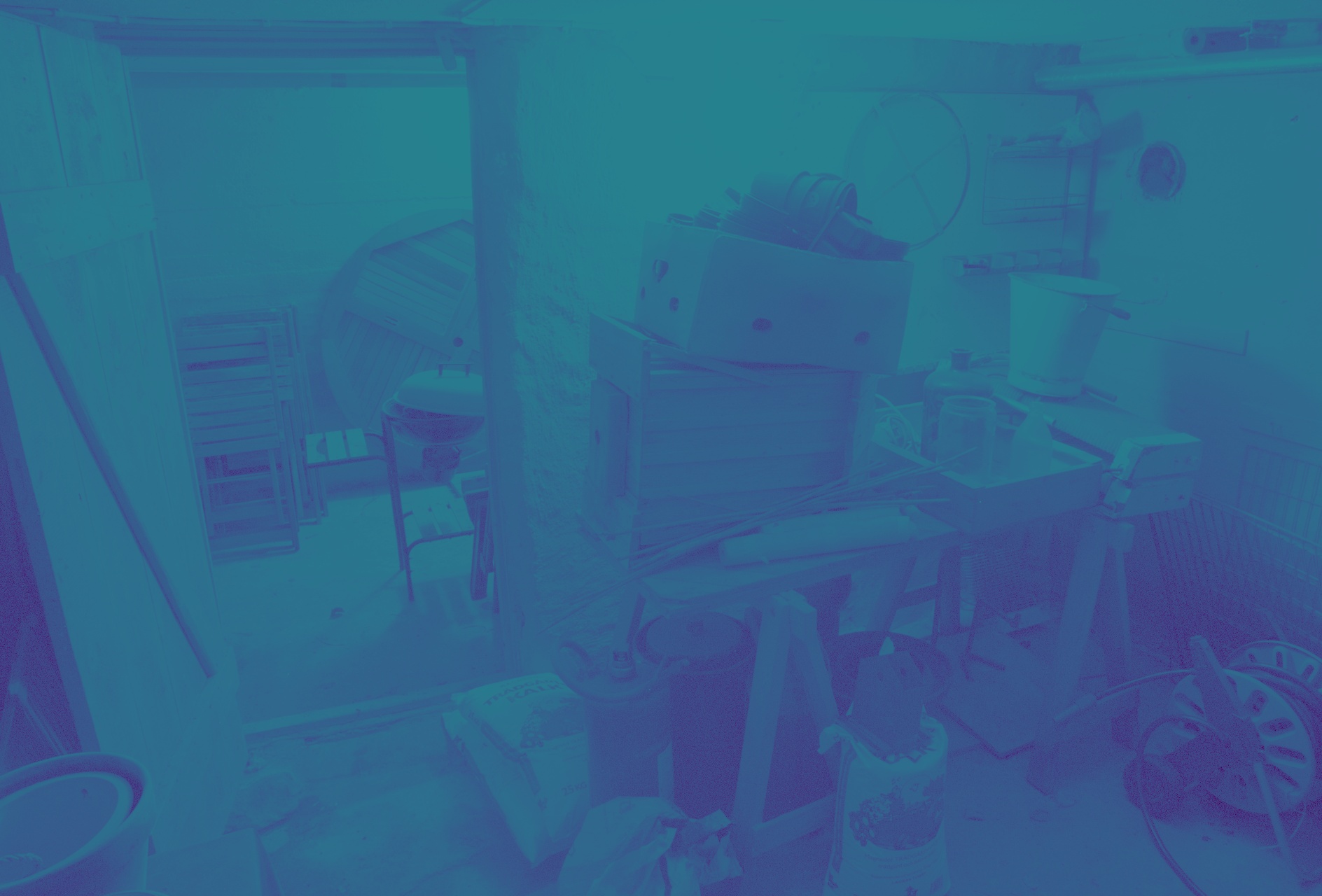}&
    \includegraphics[width=\tmplength]{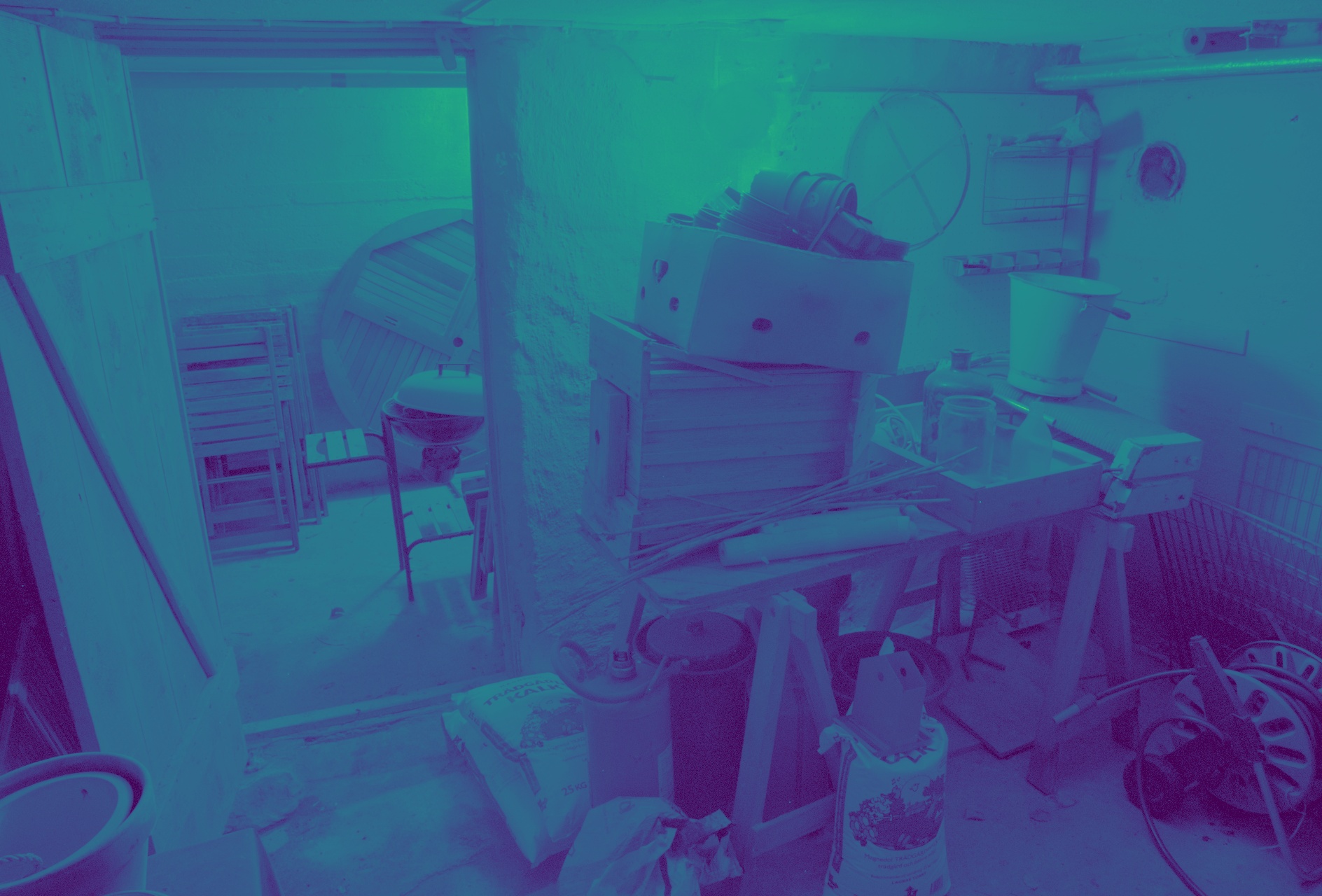}&
    \includegraphics[width=\tmplength]{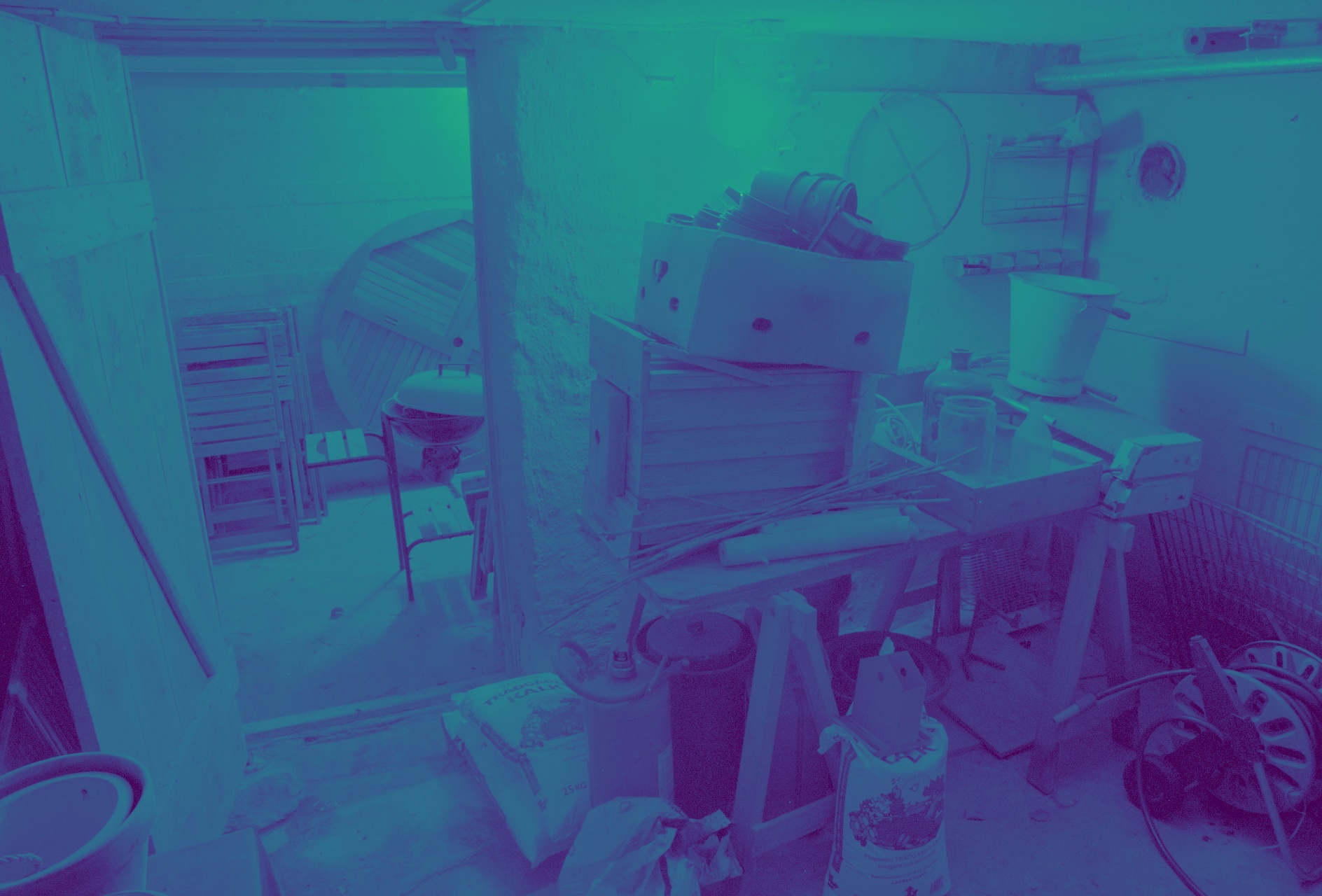}&
    \includegraphics[width=\tmplength]{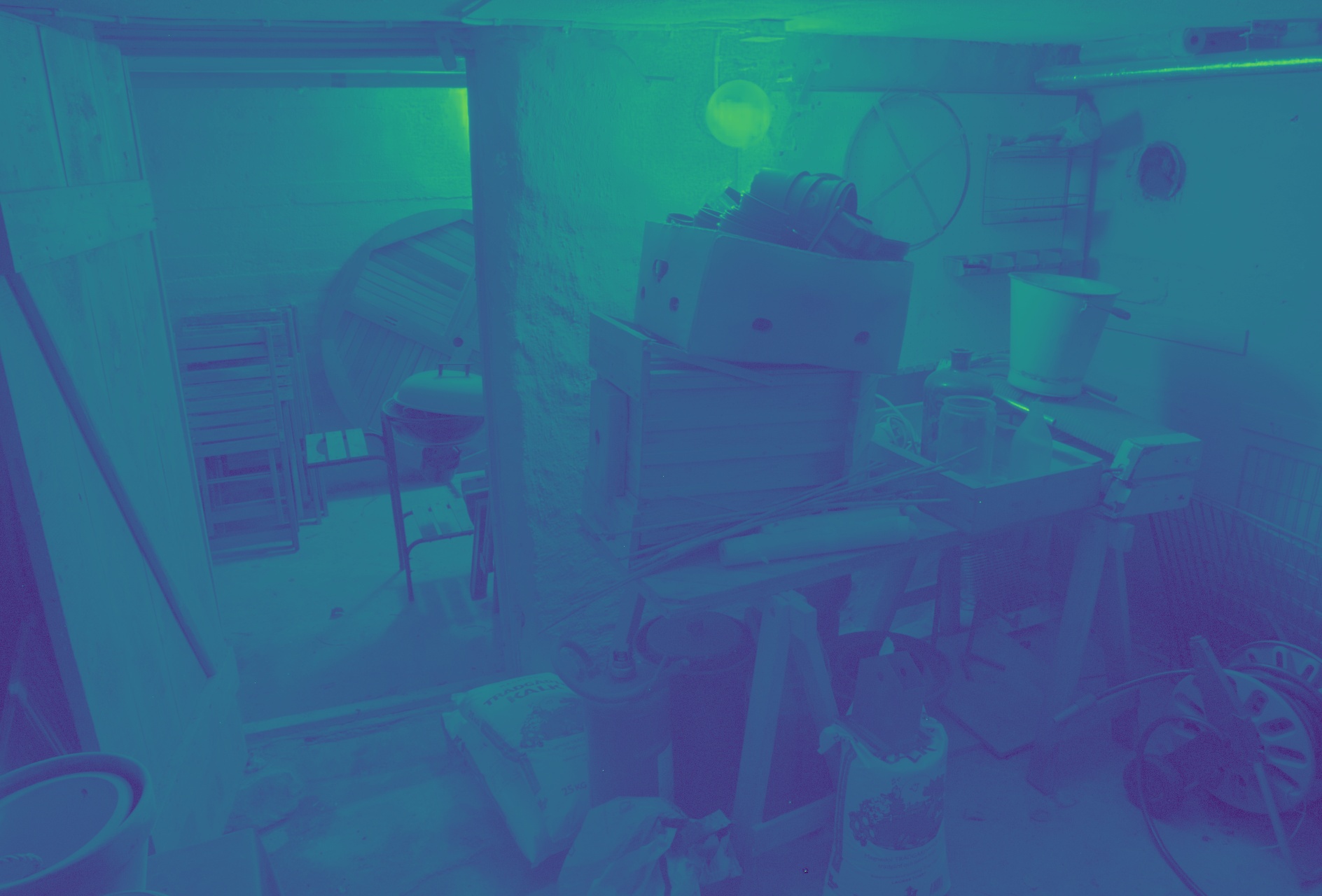}&
    \includegraphics[width=\tmplength]{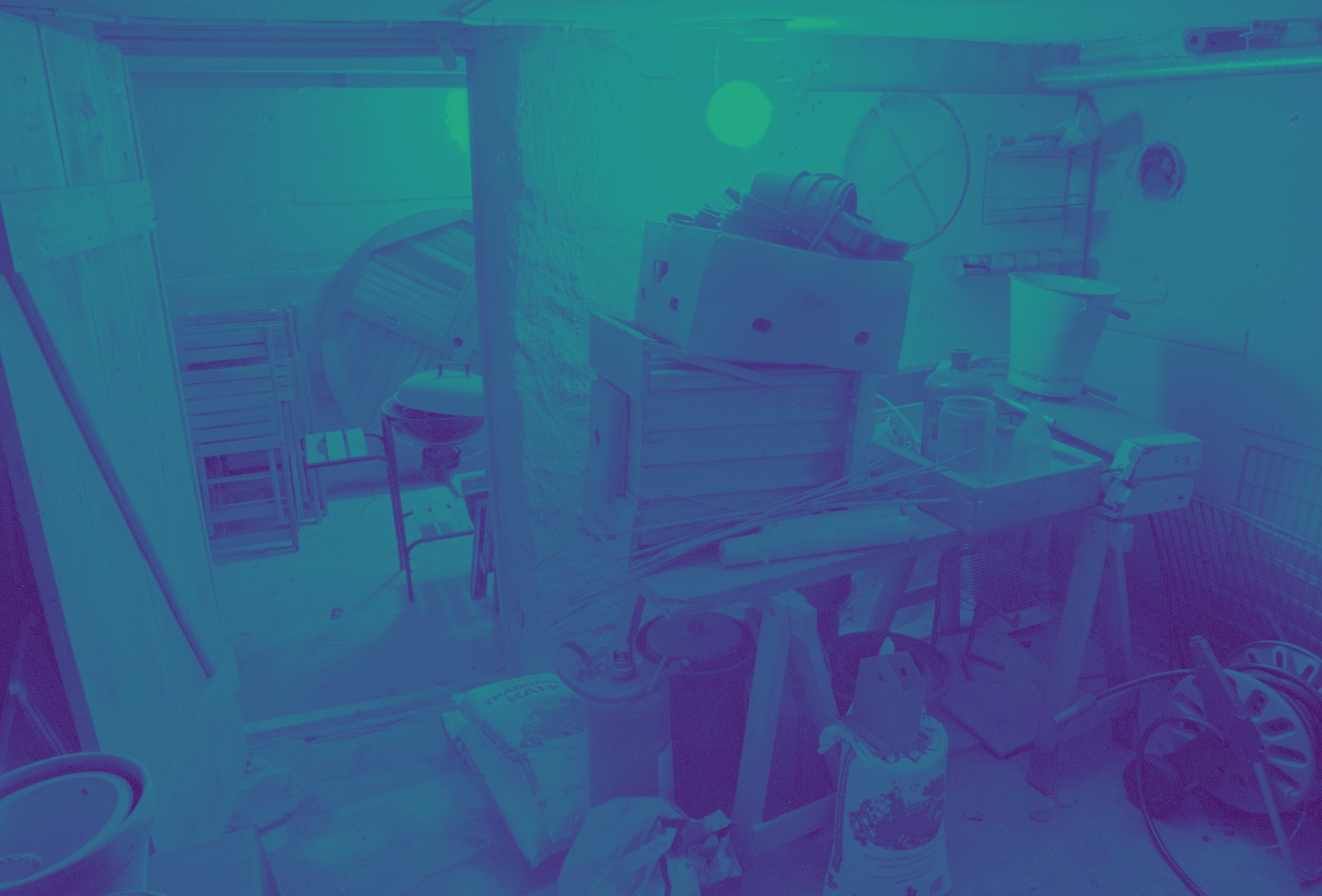}&
    \includegraphics[width=\tmplength]{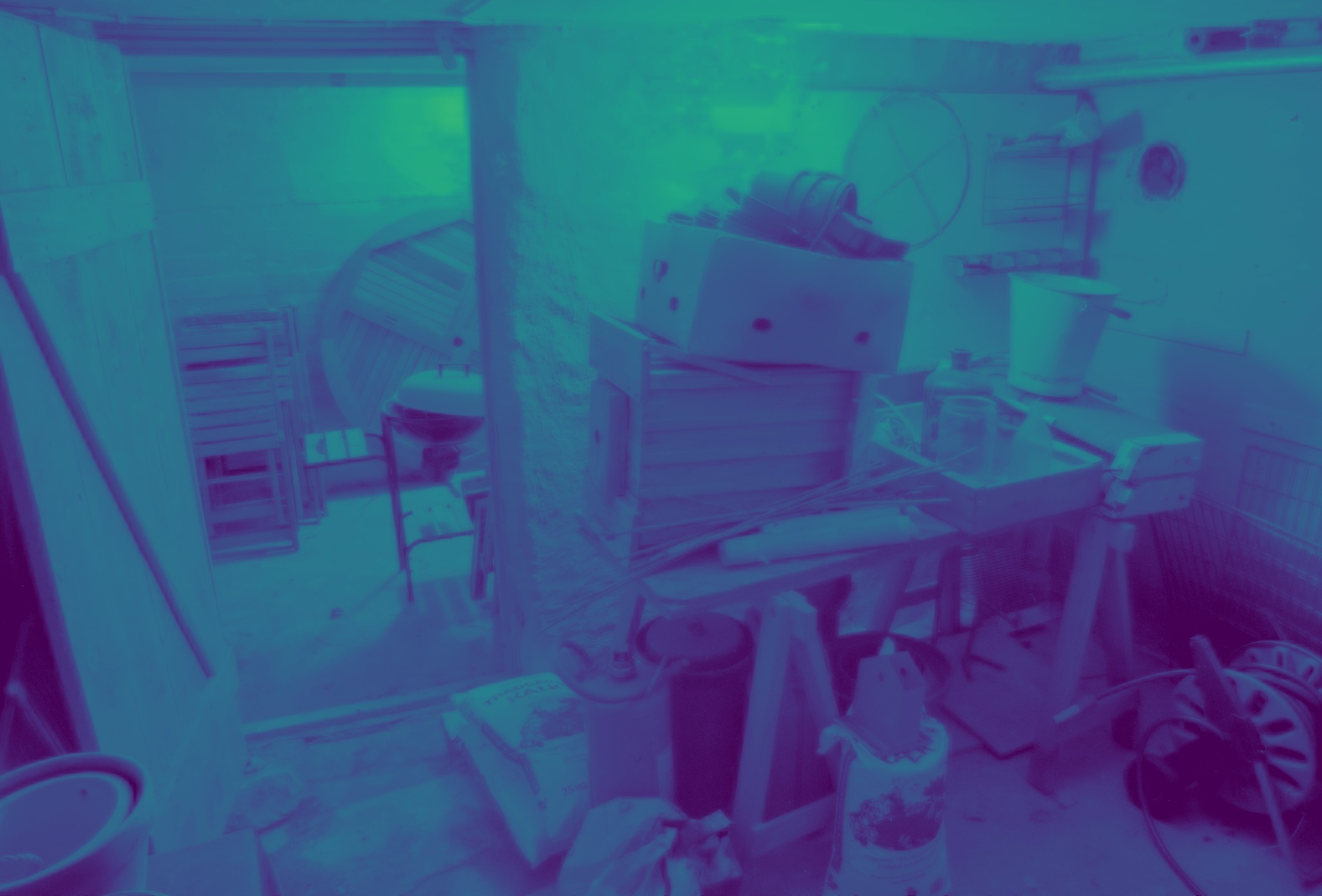}&
    \includegraphics[width=\tmplength]{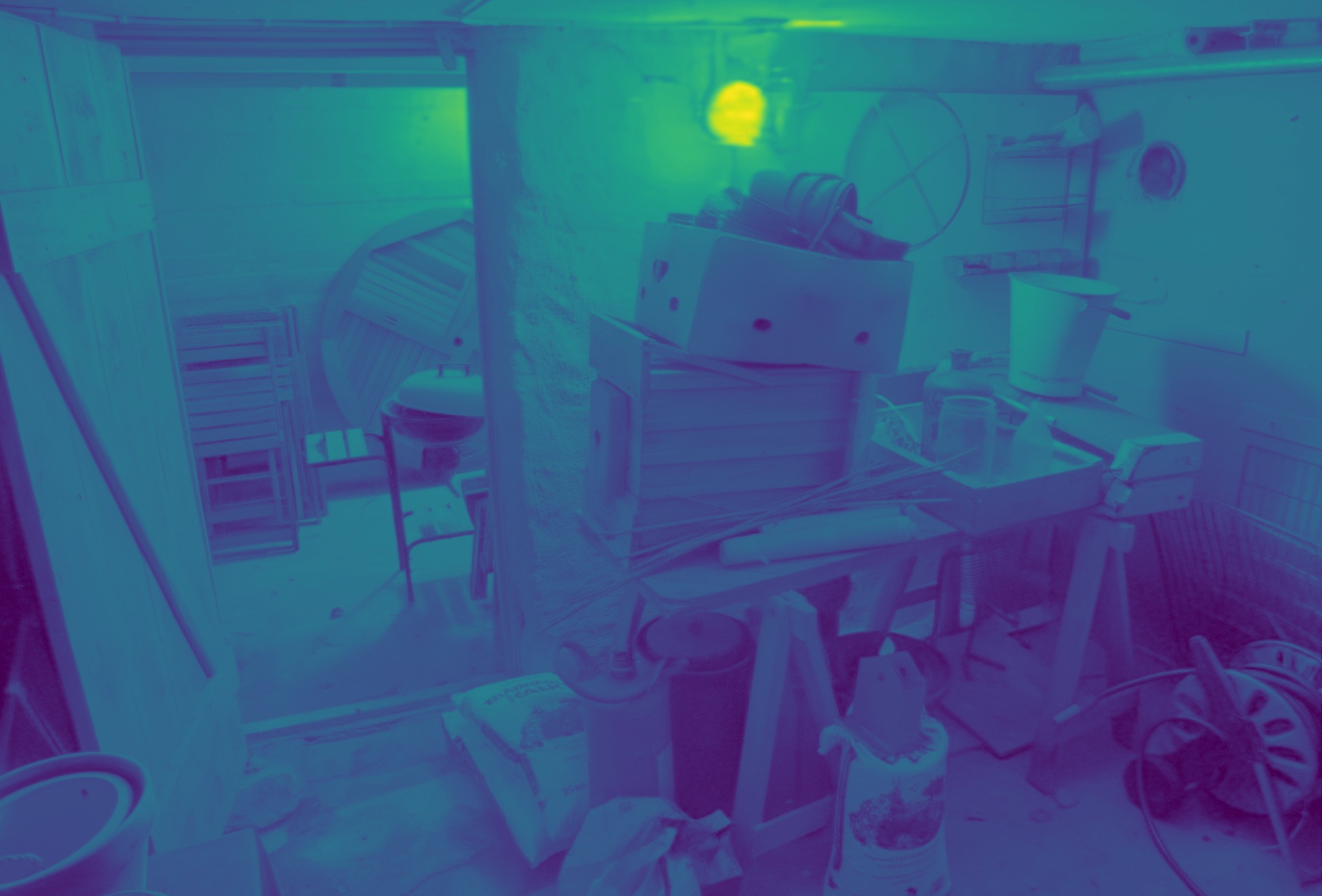}&
    \includegraphics[width=\tmplength]{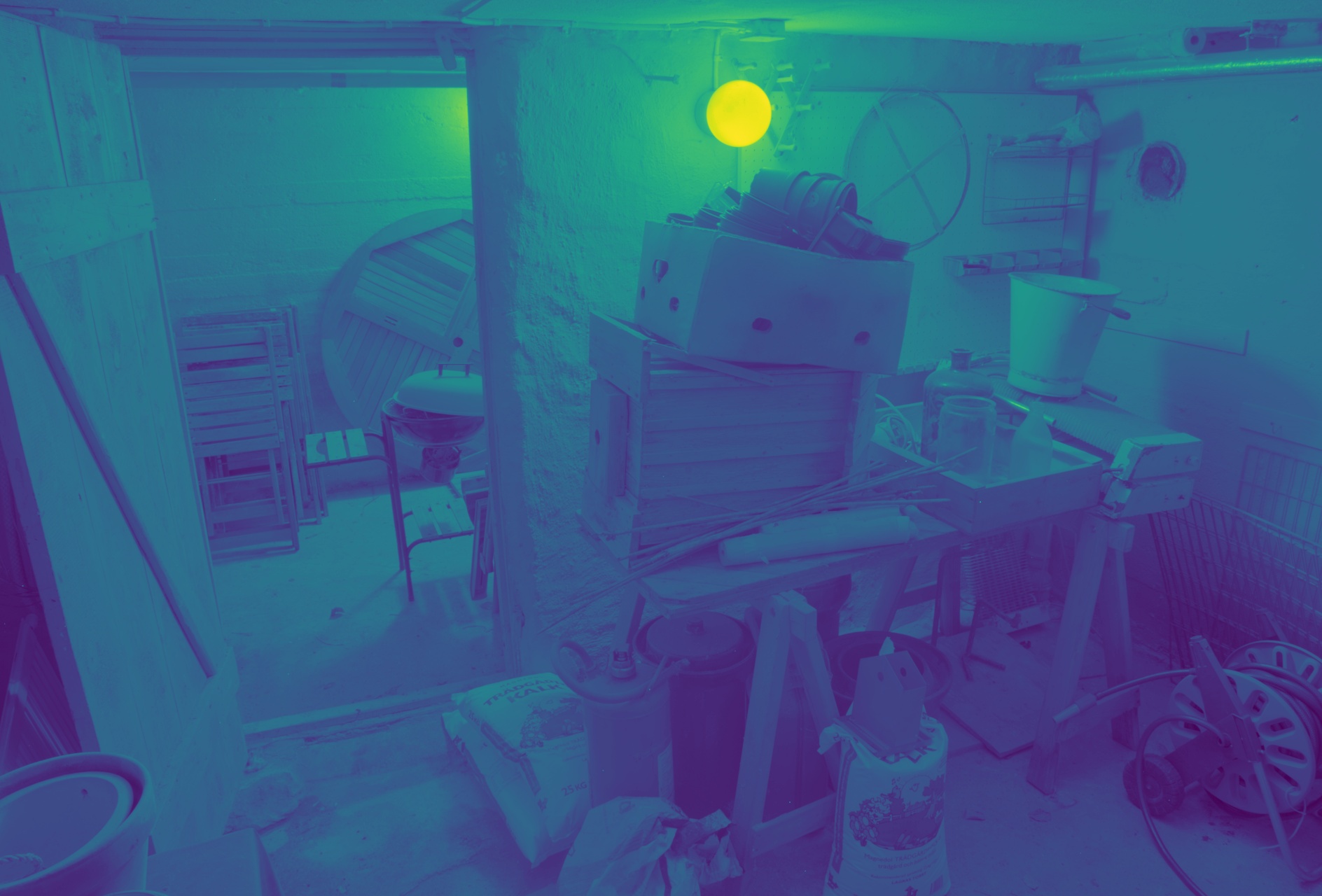} &
    \includegraphics[height=\cbarheight]{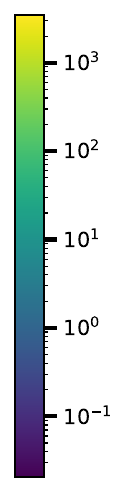}\\
    &
    \includegraphics[width=\tmplength]{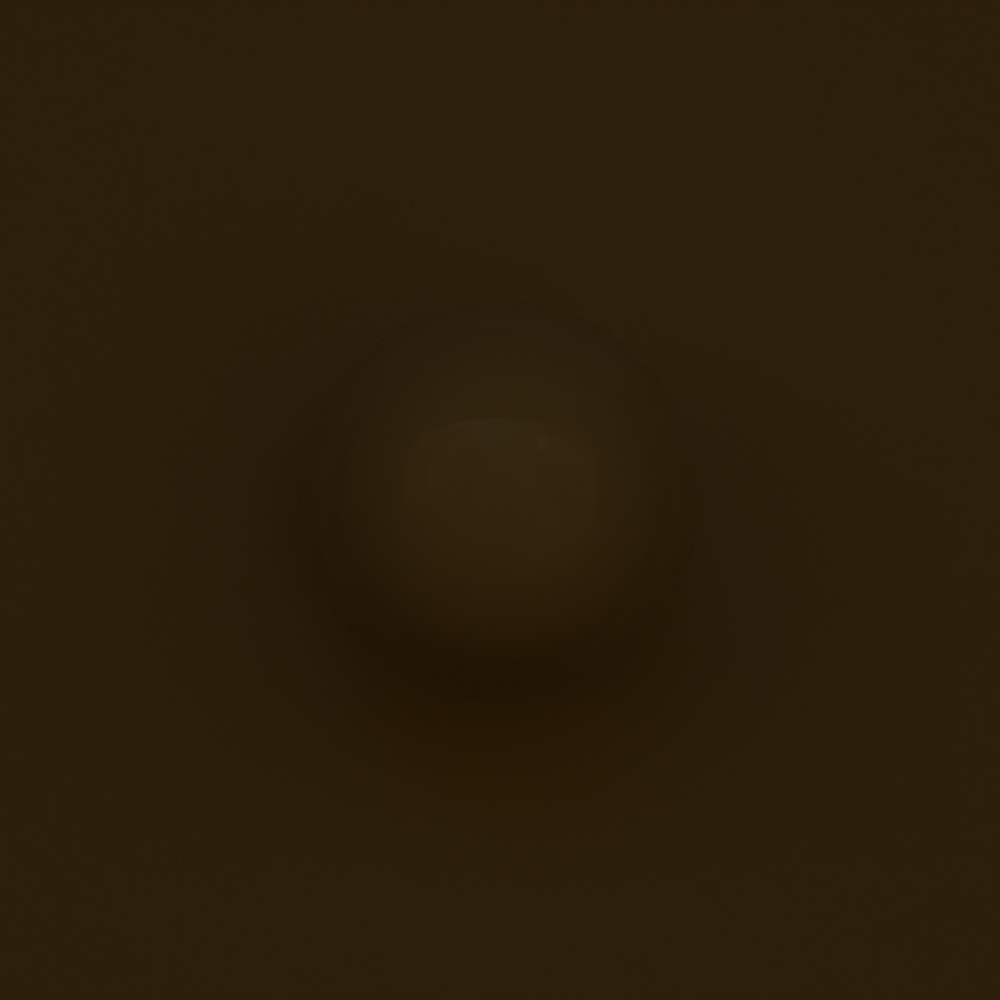}&
    \includegraphics[width=\tmplength]{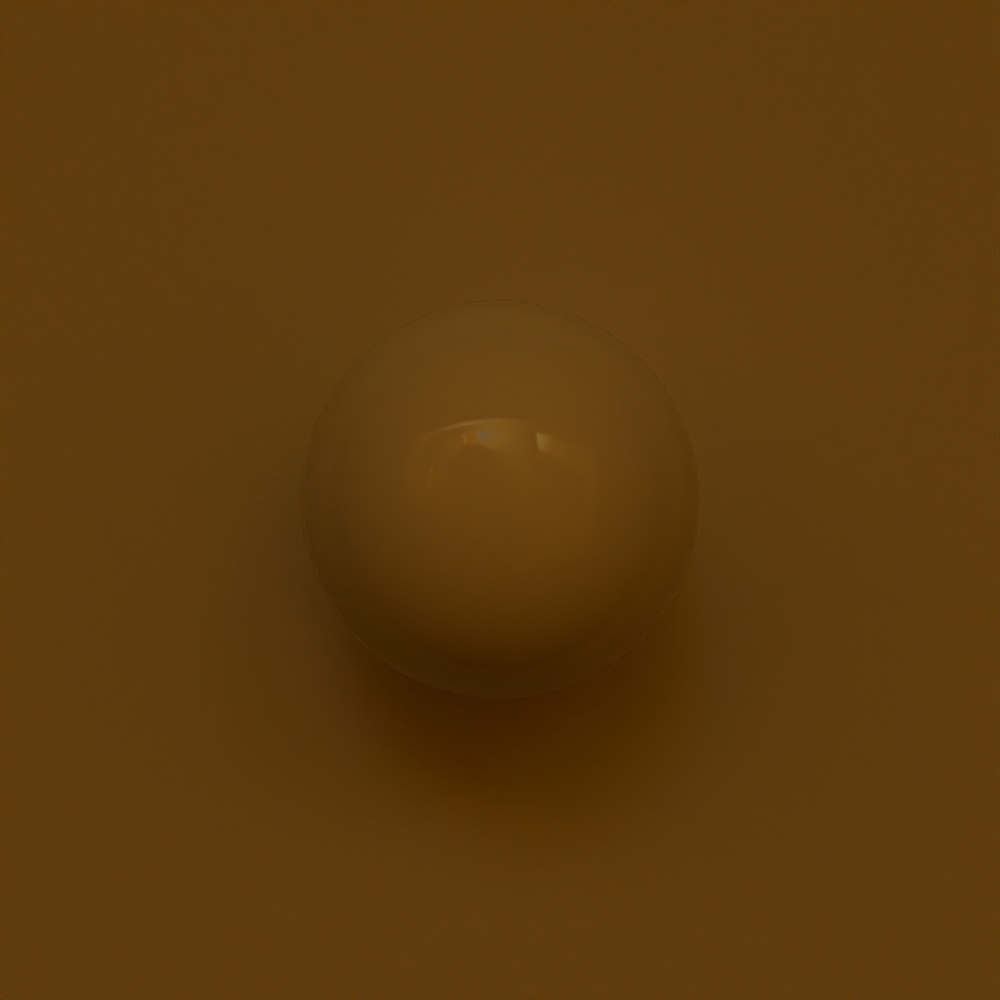}&
    \includegraphics[width=\tmplength]{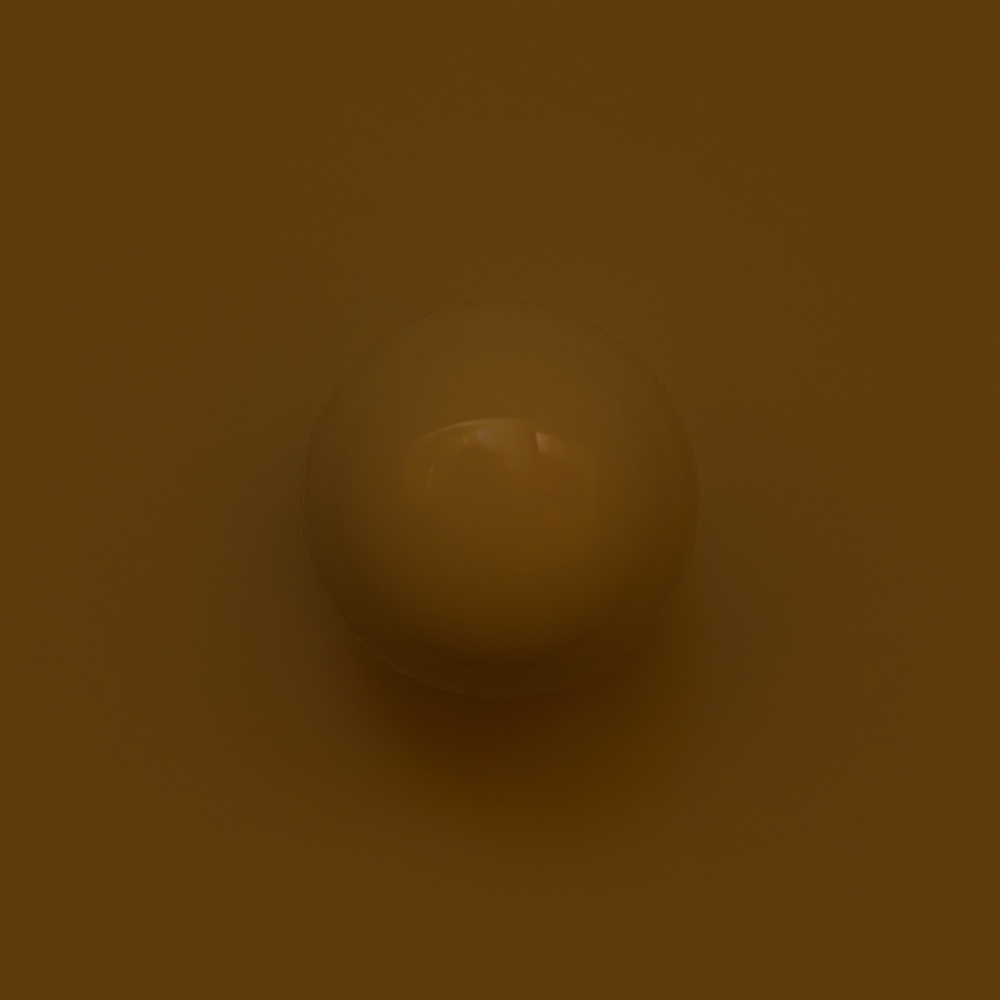}&
    \includegraphics[width=\tmplength]{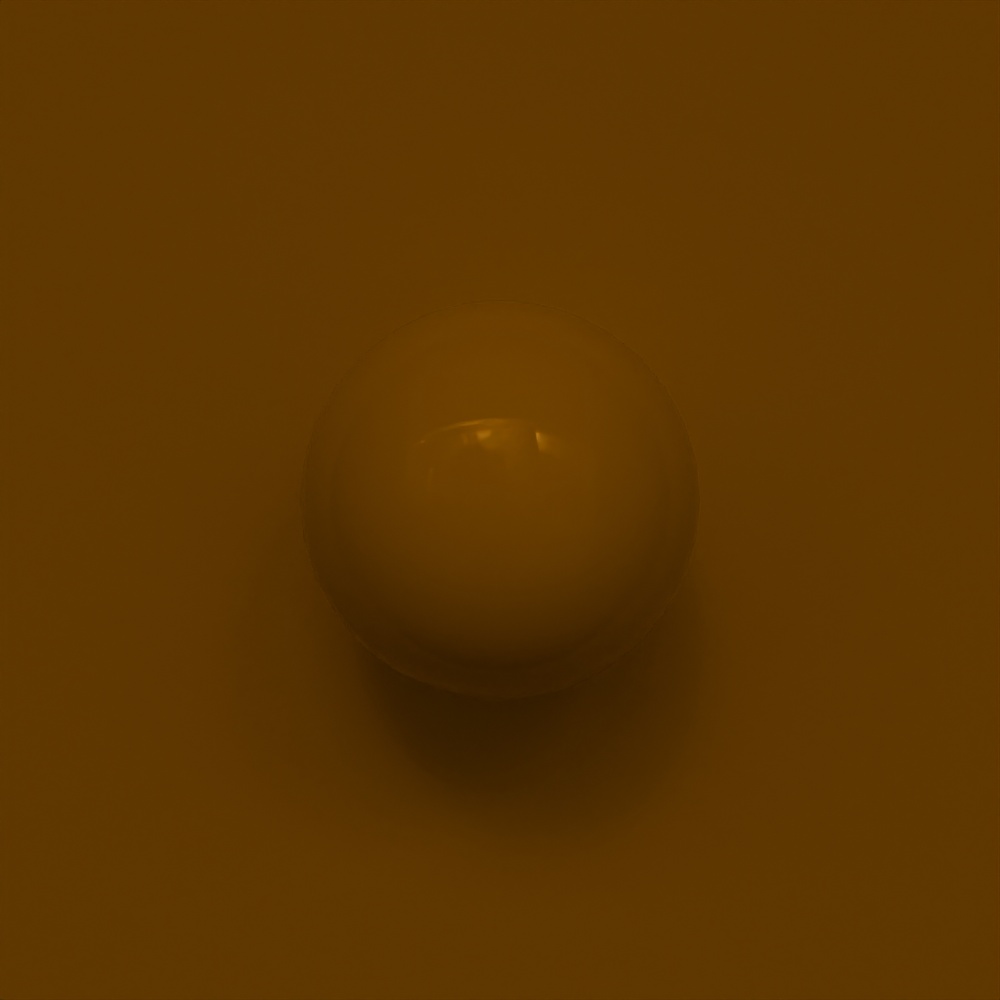}&
    \includegraphics[width=\tmplength]{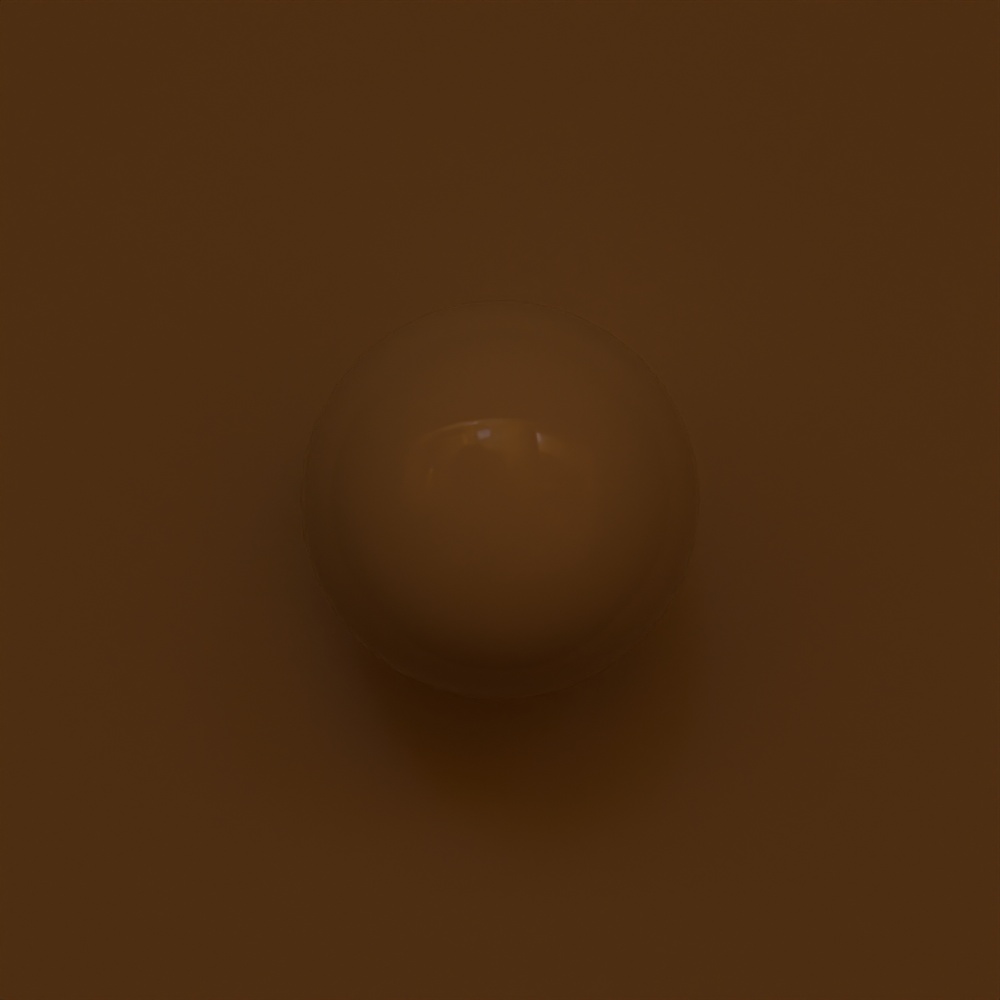}&
    \includegraphics[width=\tmplength]{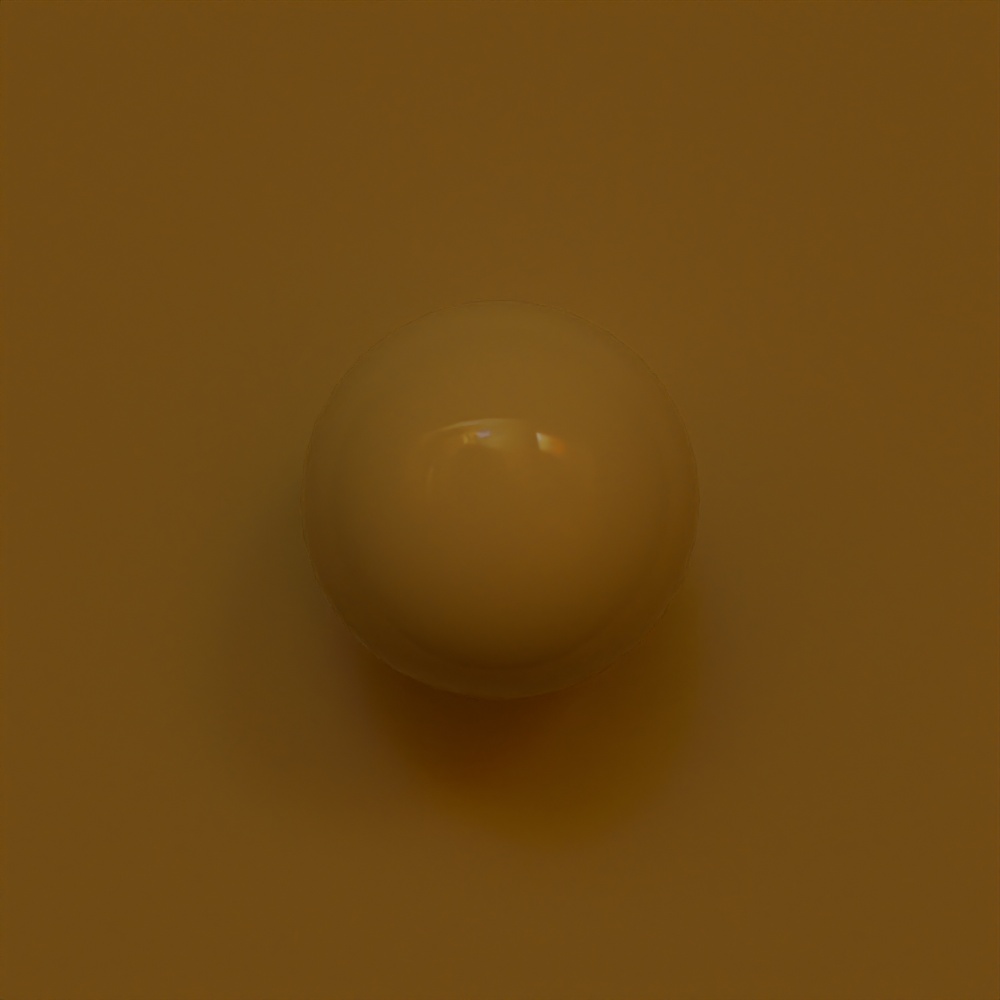}&
    \includegraphics[width=\tmplength]{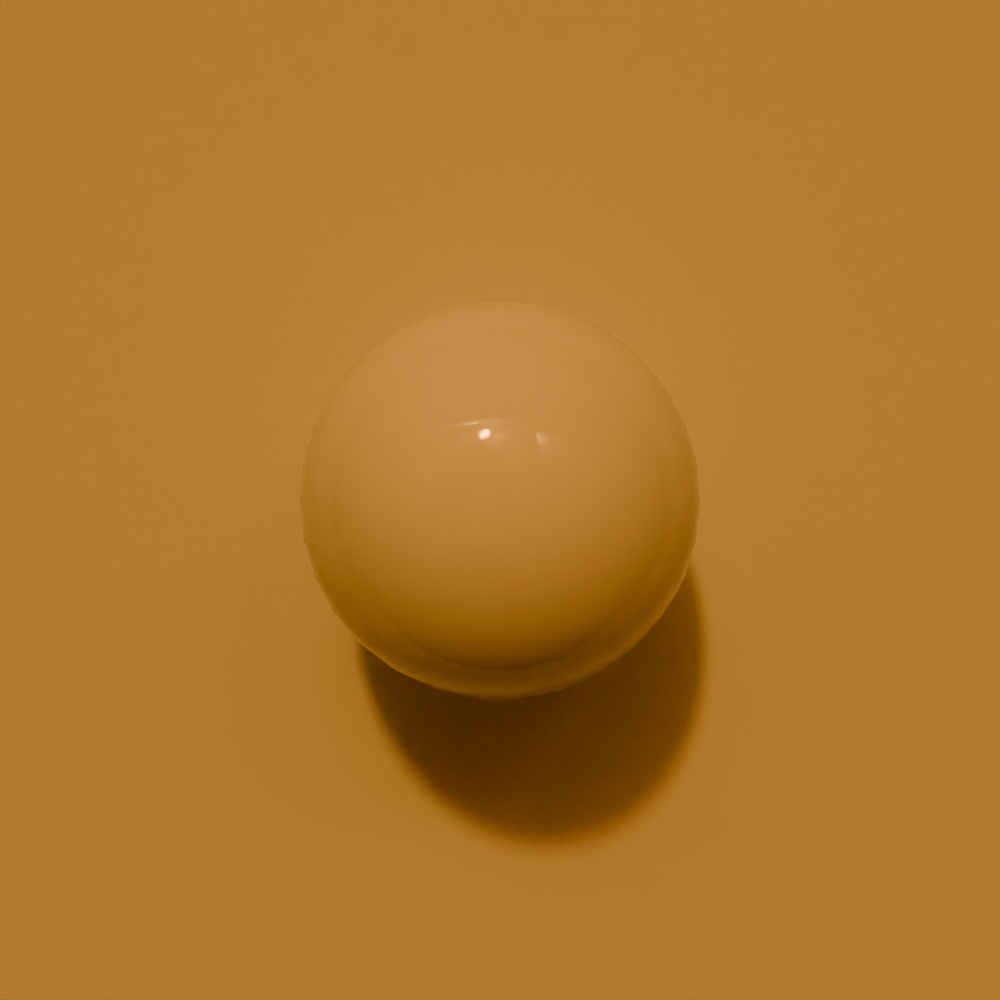}&
    \includegraphics[width=\tmplength]{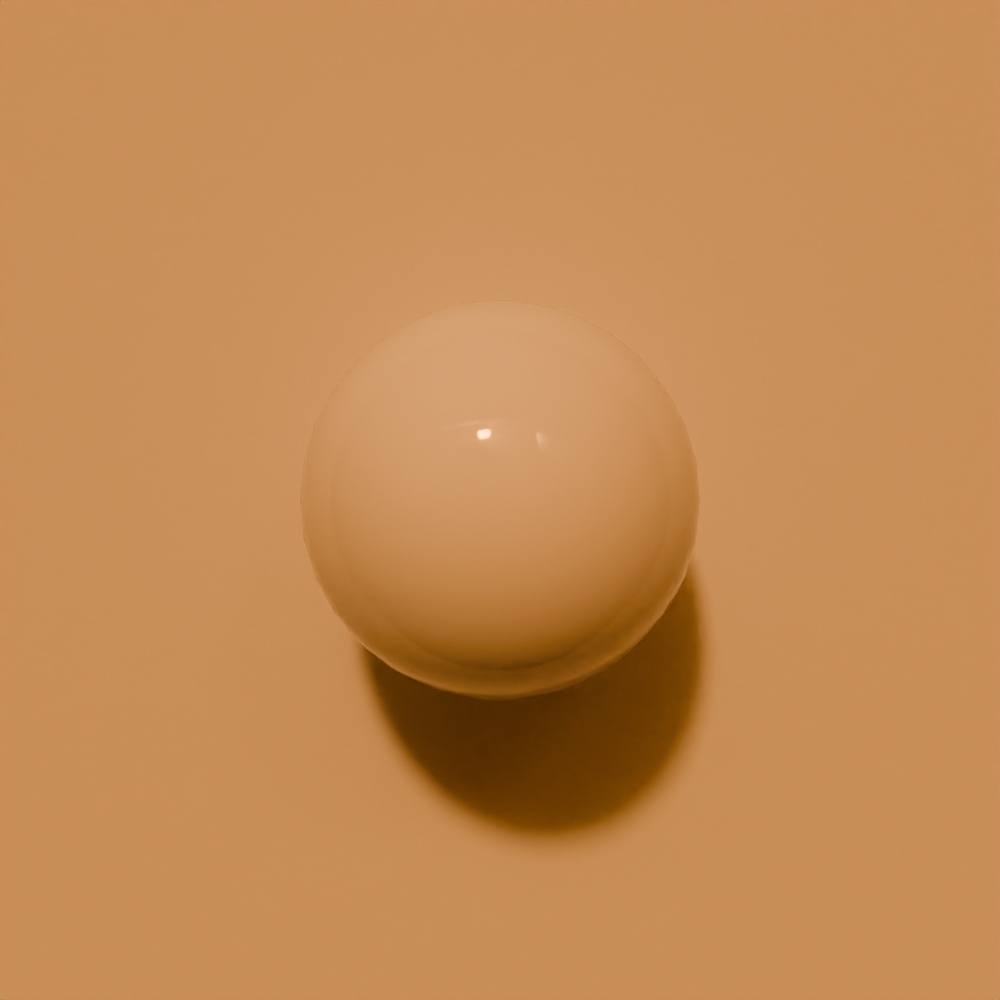}  & \\
    \end{tabular}
    \caption{Qualitative evaluation on the filtered Si-HDR dataset~\cite{hanji2022comparison}. From an LDR image (left), we compare the luminance of each method's predicted HDR (even rows) and the rendering of a glossy sphere illuminated by the prediction (odd rows). Our approach more accurately estimates light intensity, producing renders similar to those rendered using the ground truth HDR image.} 
    \label{fig:relight_glossy_sihdr}
\end{figure*}

The approach that is most closely related to ours is that of PanoHDR-NeRF~\cite{gera2022panohdrnerf}, who train a NeRF-based implicit representation on images that have been converted to HDR using the LANet~\cite{yu2021luminance_lanet} LDR-to-HDR architecture. They also provide a dataset of interior scenes captured with a 360\textdegree{} camera, in which a few (between 8--10) ground truth HDR panoramas were captured using exposure bracketing. 

We therefore compare our HDR Gaussian Splatting representation to PanoHDR-NeRF on two scenes from their dataset and show the quantitative results in \cref{tab:quant_3D}. These metrics were computed by first rendering a synthetic scene (similar to \cref{sec:eval_light_estimation}), lit by each method's prediction, and compared to the ground truth. The table shows that our method significantly outperforms the baseline. We note that the results they report in their paper also include a version of LANet fine-tuned on a small dataset of HDR images captured by the same camera that was used to capture the ground truth---we use the ``vanilla'' model to avoid unfairly advantaging the method, and assume that no method has knowledge of the actual camera used to capture the data. 



\subsection{Ablation}

\begin{figure}
    \centering
    \setlength{\tabcolsep}{1pt}
    \newlength{\mywidth}
    \setlength{\mywidth}{0.235\linewidth}
    \begin{tabular}{ccccc}
    \rot{\tiny \hspace{11pt}HDR} &
    \includegraphics[width=\mywidth]{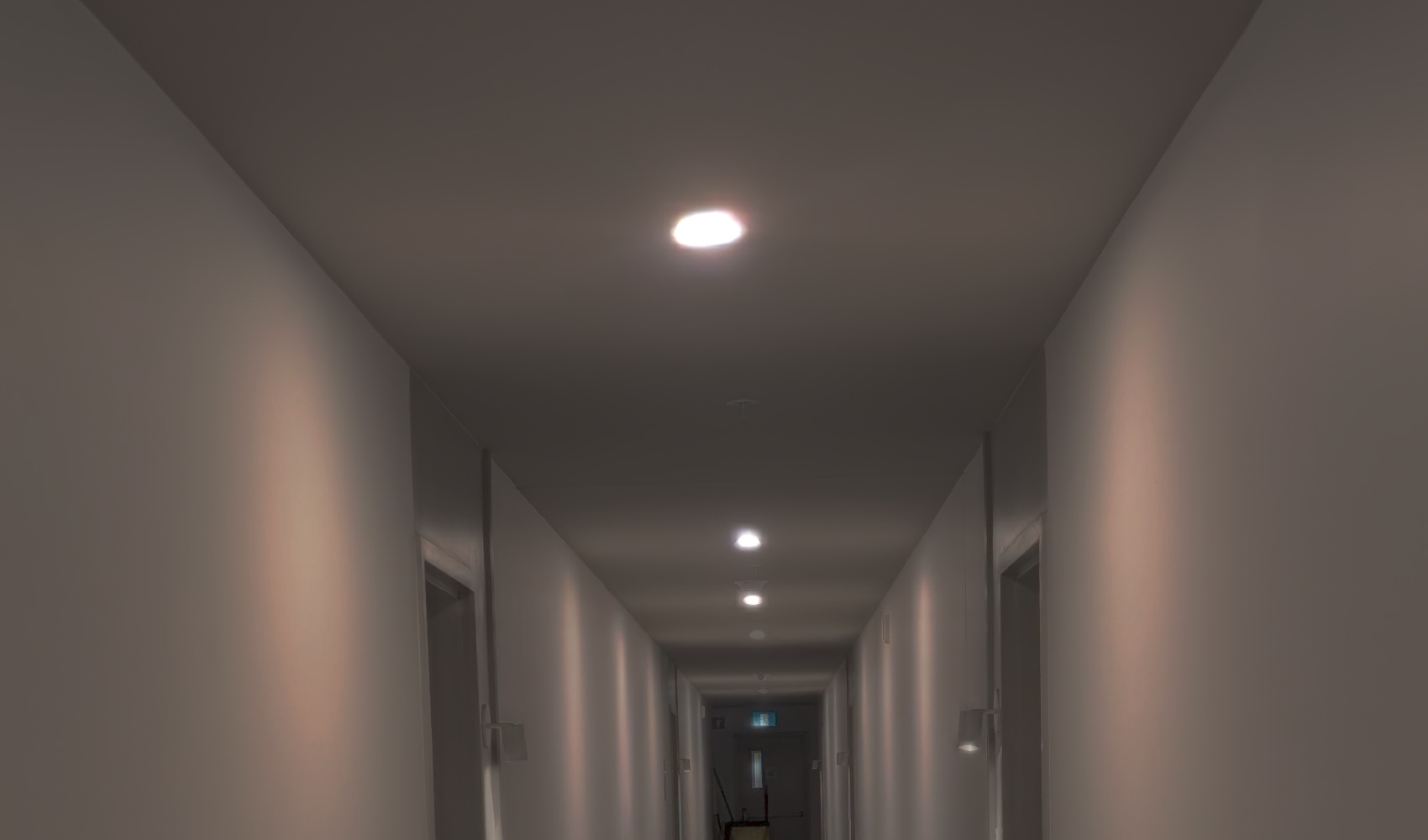} &
    \includegraphics[width=\mywidth]{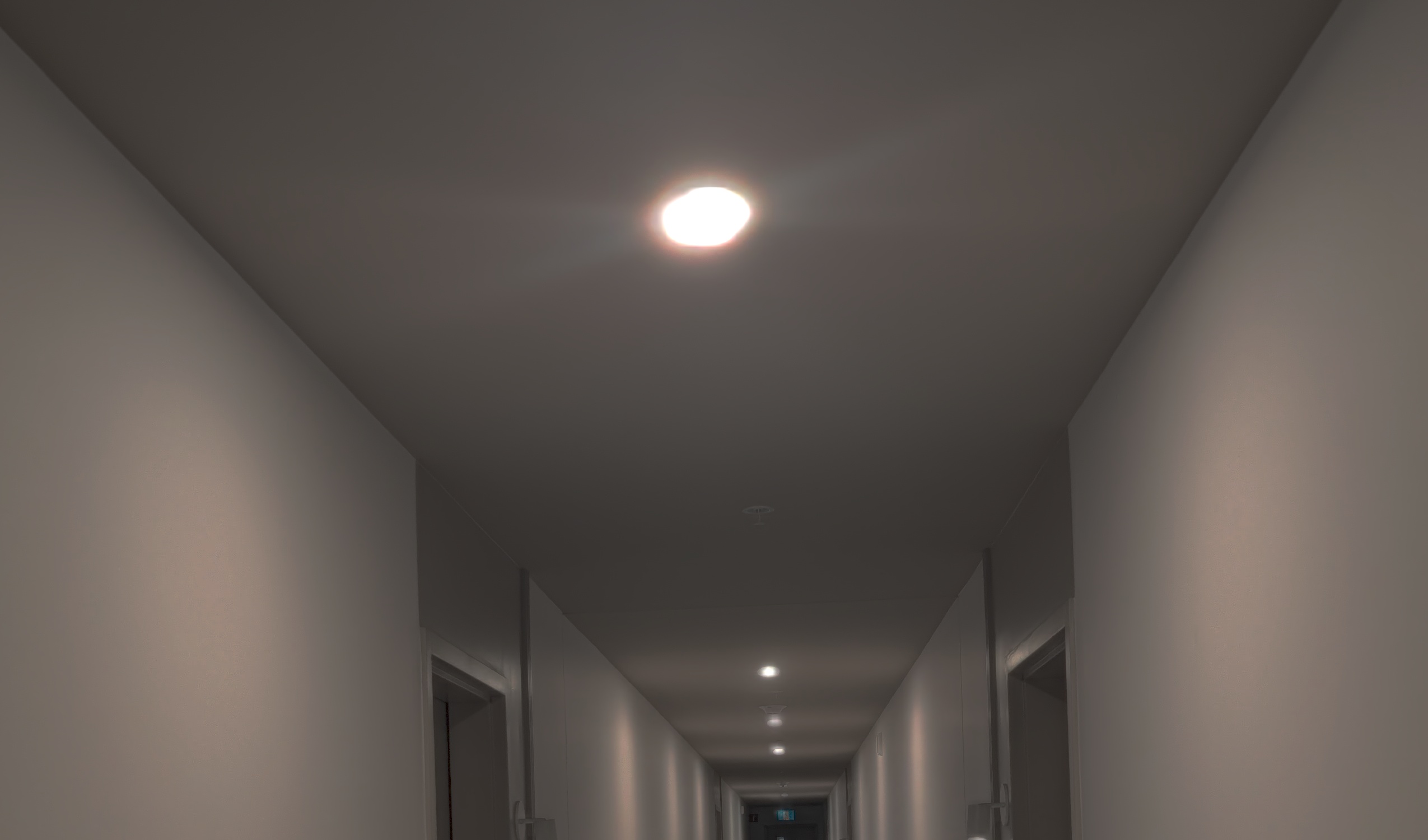} &
    \includegraphics[width=\mywidth]{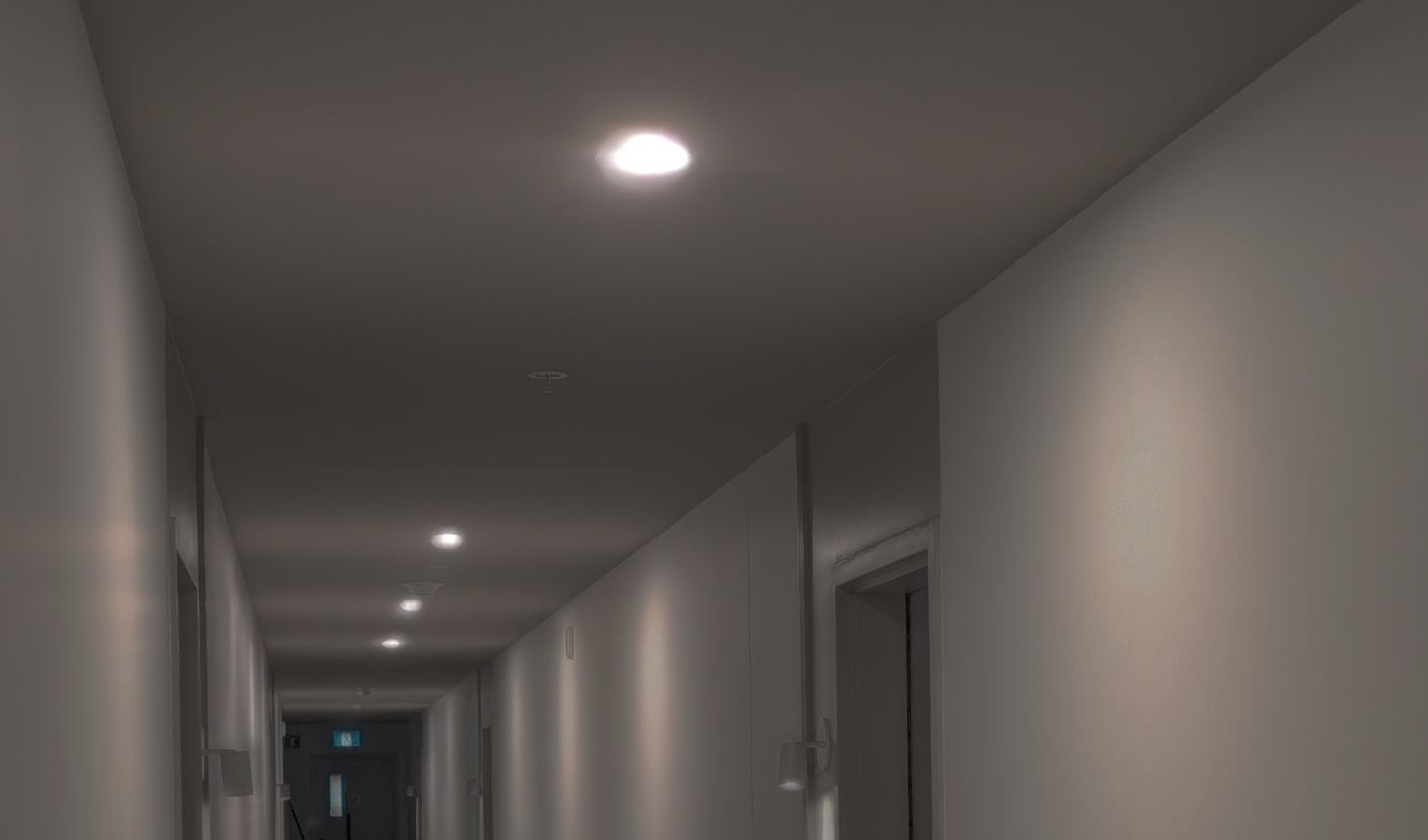} &
    \includegraphics[width=\mywidth]{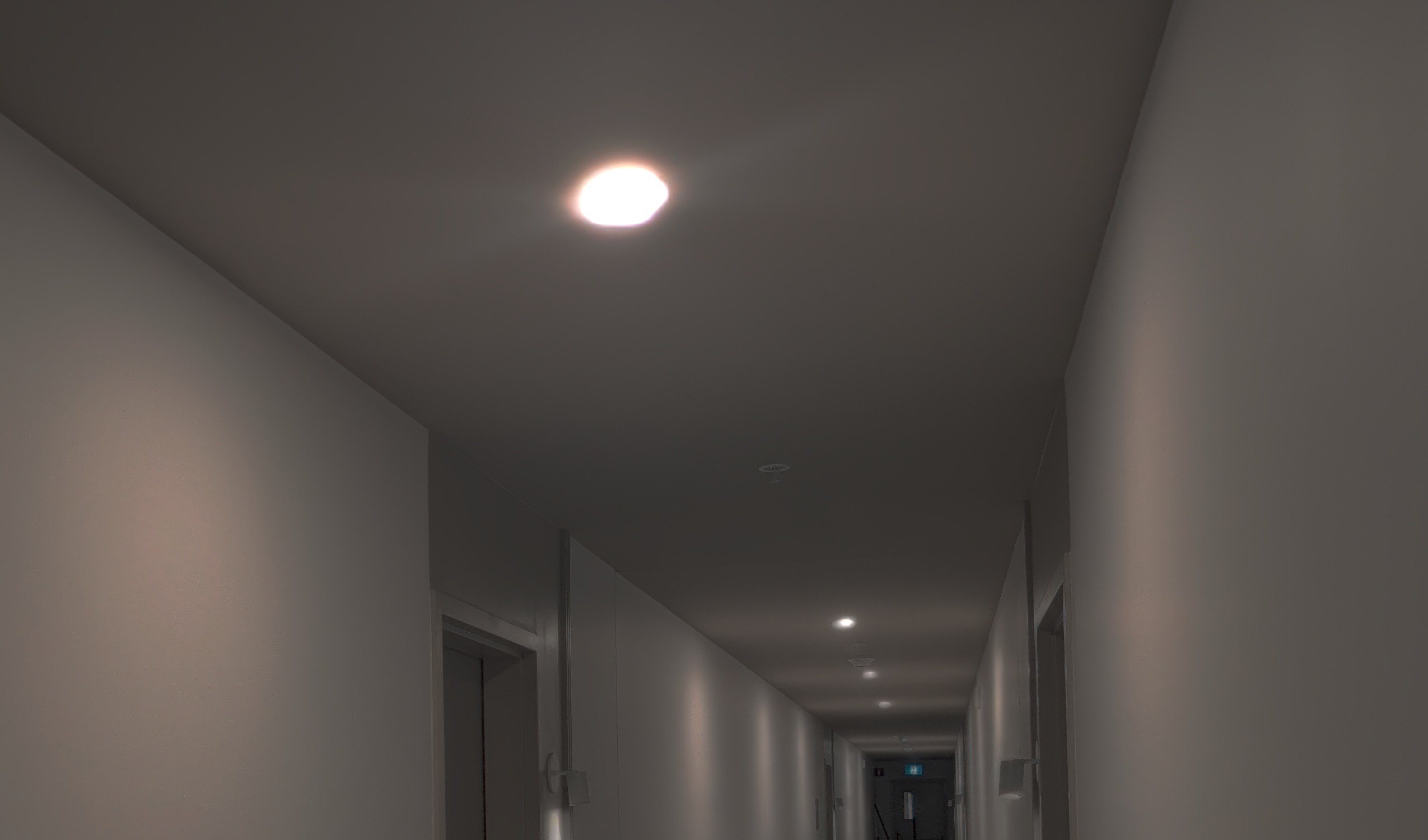} \\*[-0.25em]
     \rot{\tiny \hspace{10pt}Clamp} &
    \includegraphics[width=\mywidth]{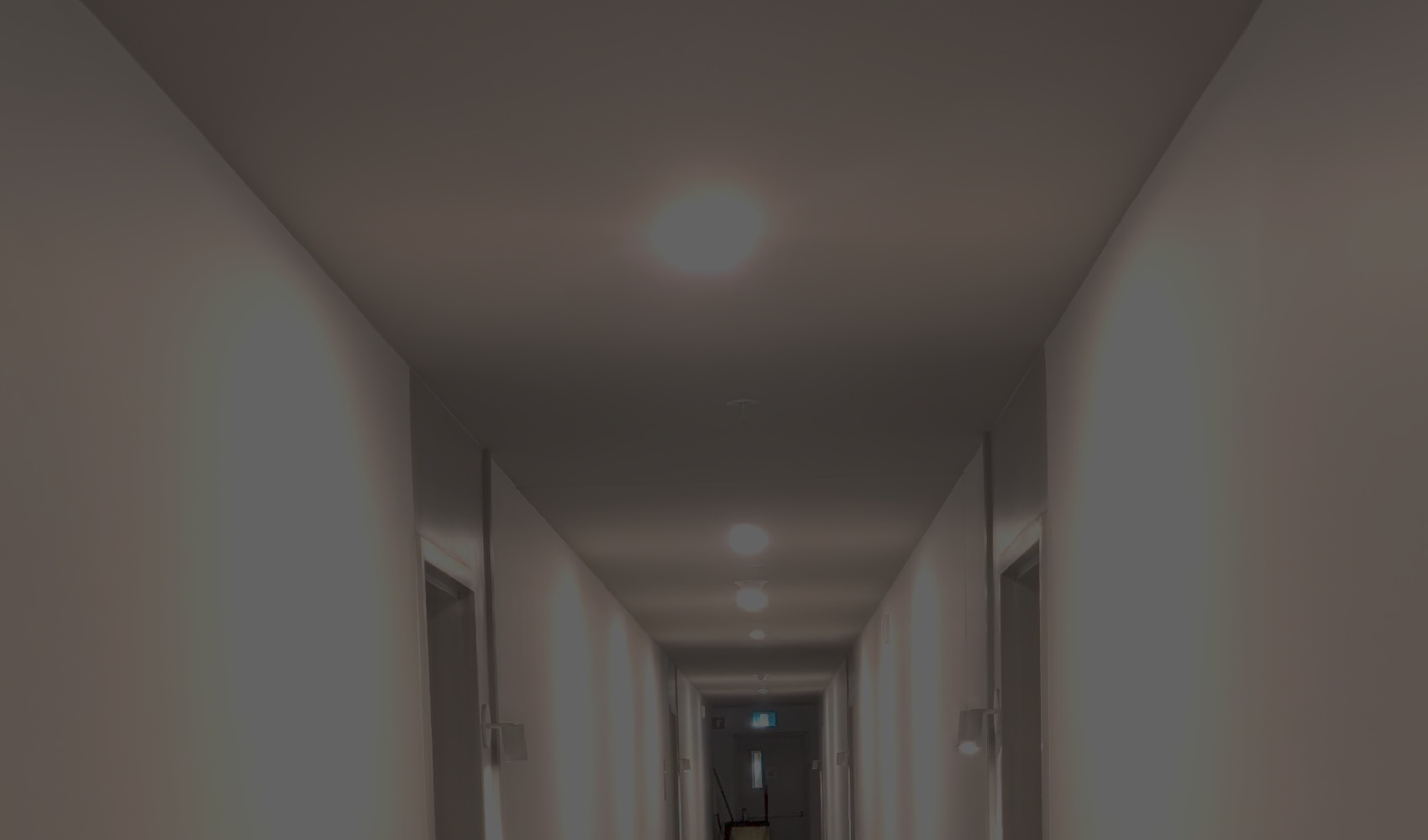} &
    \includegraphics[width=\mywidth]{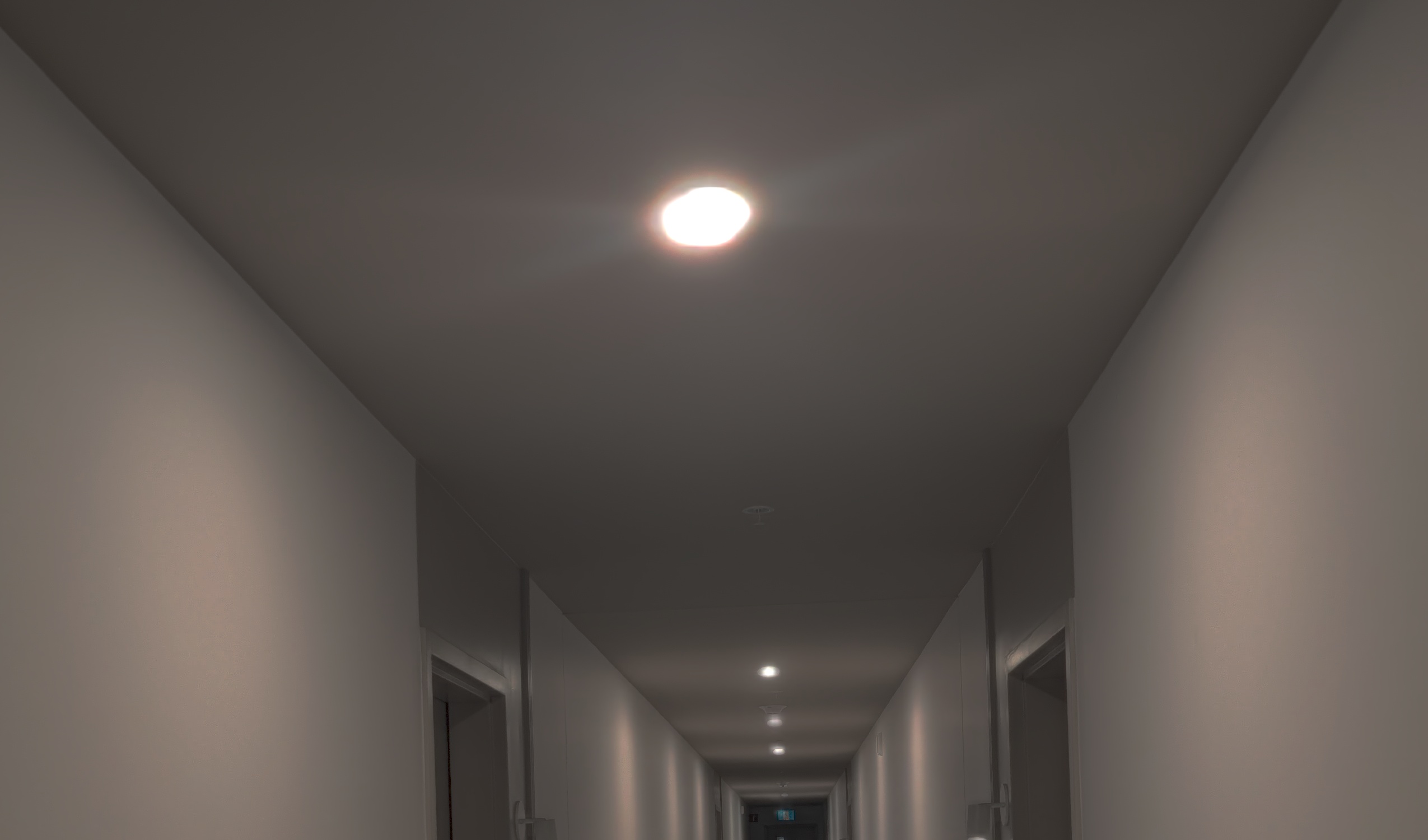} &
    \includegraphics[width=\mywidth]{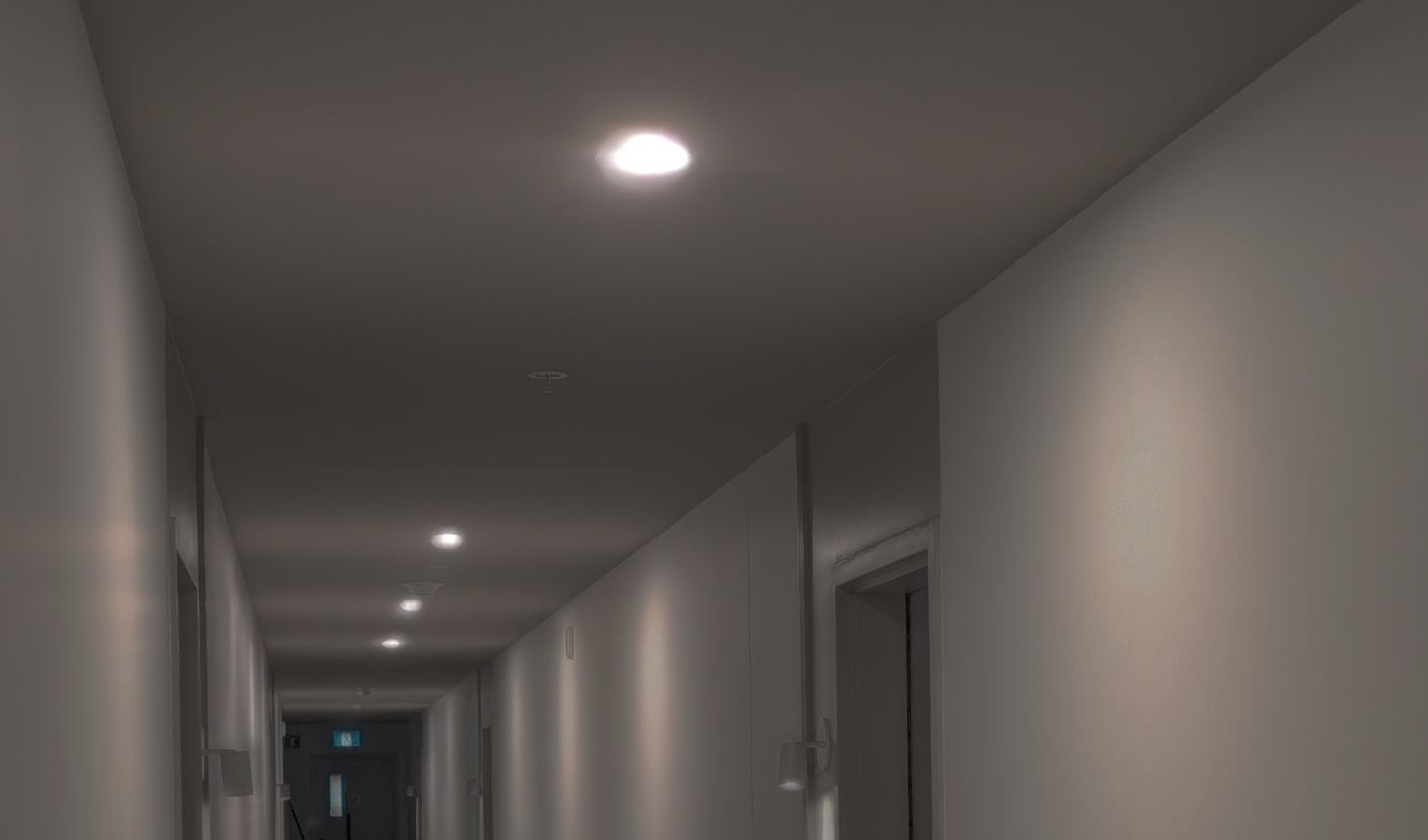} &
    \includegraphics[width=\mywidth]{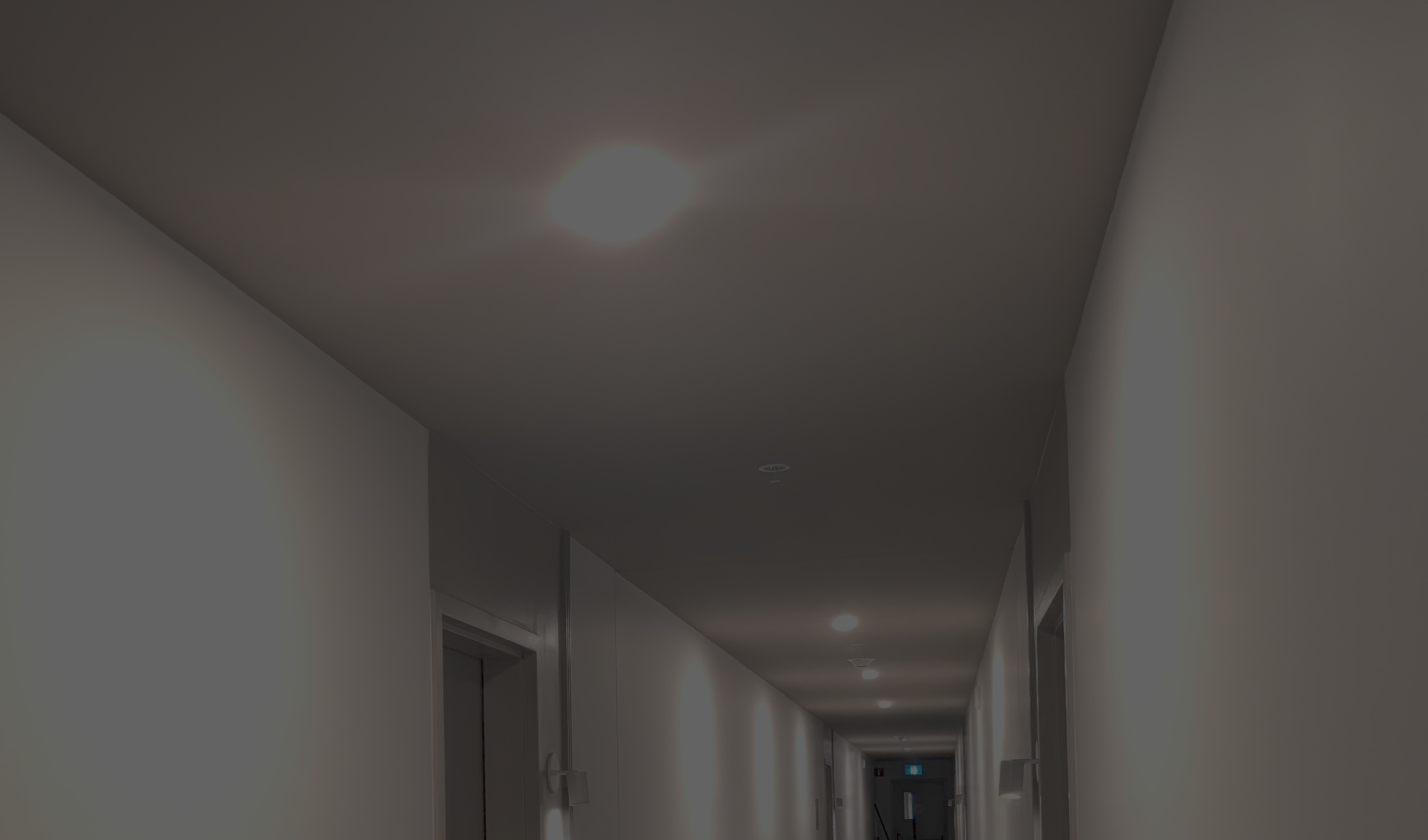} \\*[-0.25em]
    \rot{\tiny \hspace{11pt}3DGS} &
    \includegraphics[width=\mywidth]{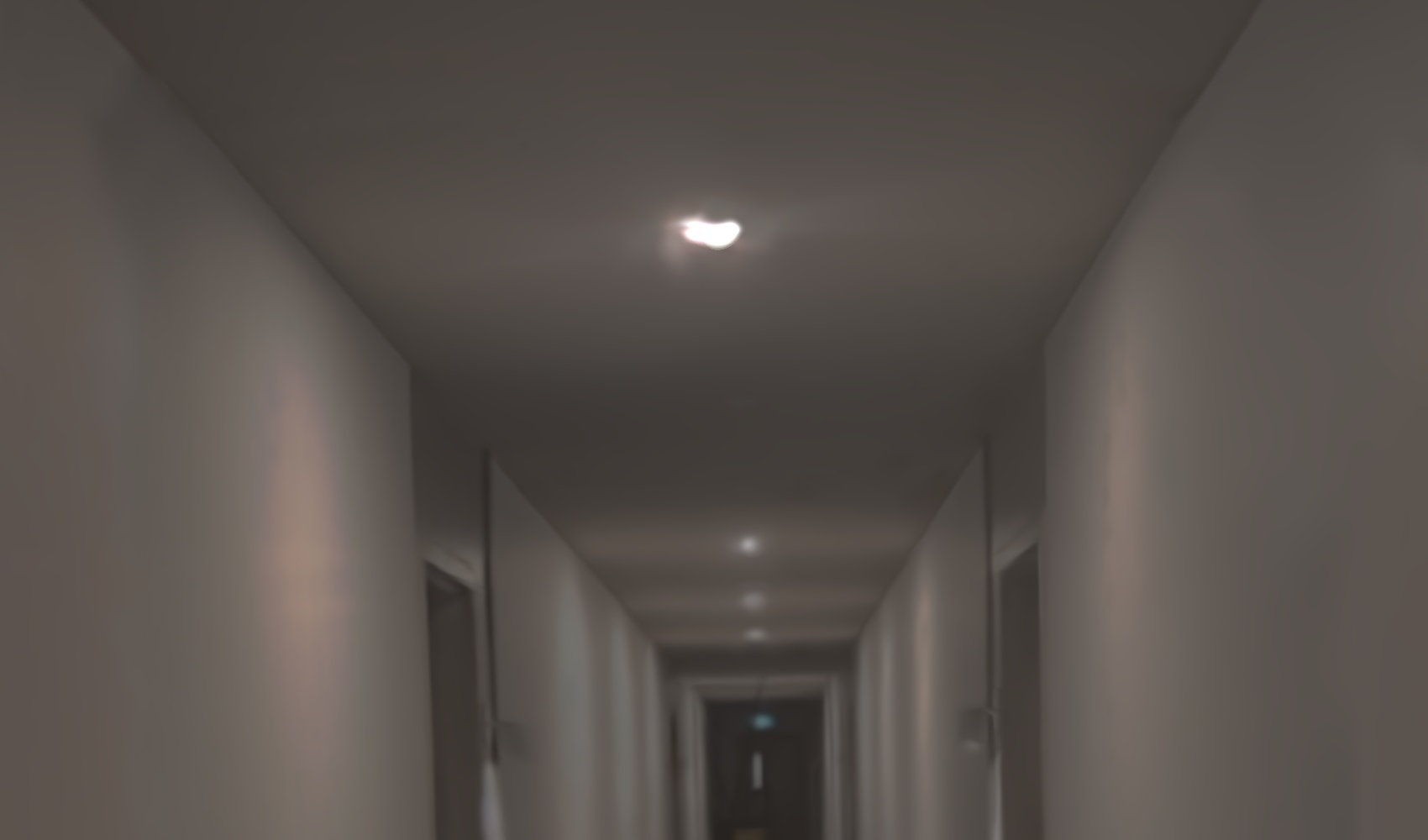} &
    \includegraphics[width=\mywidth]{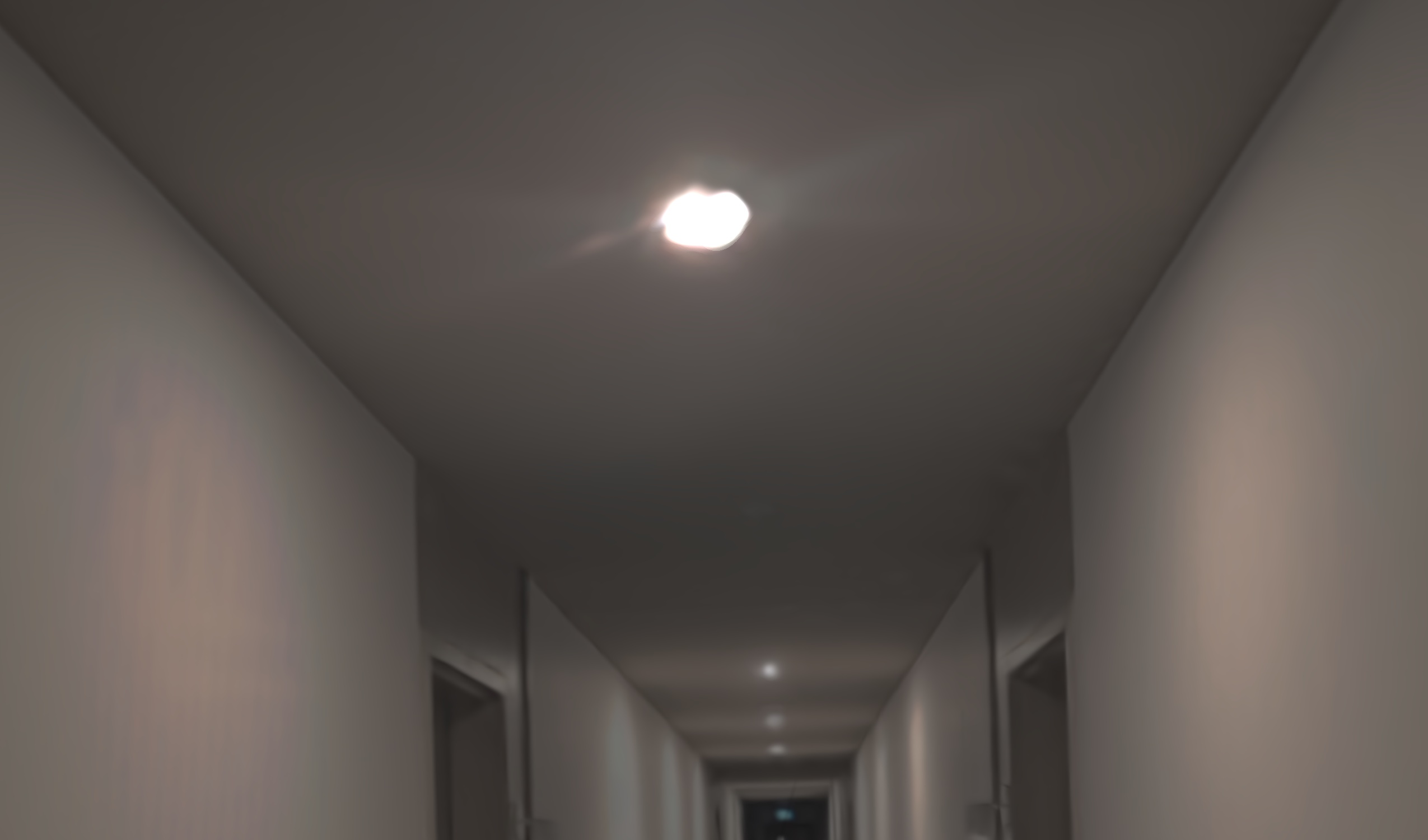} &
    \includegraphics[width=\mywidth]{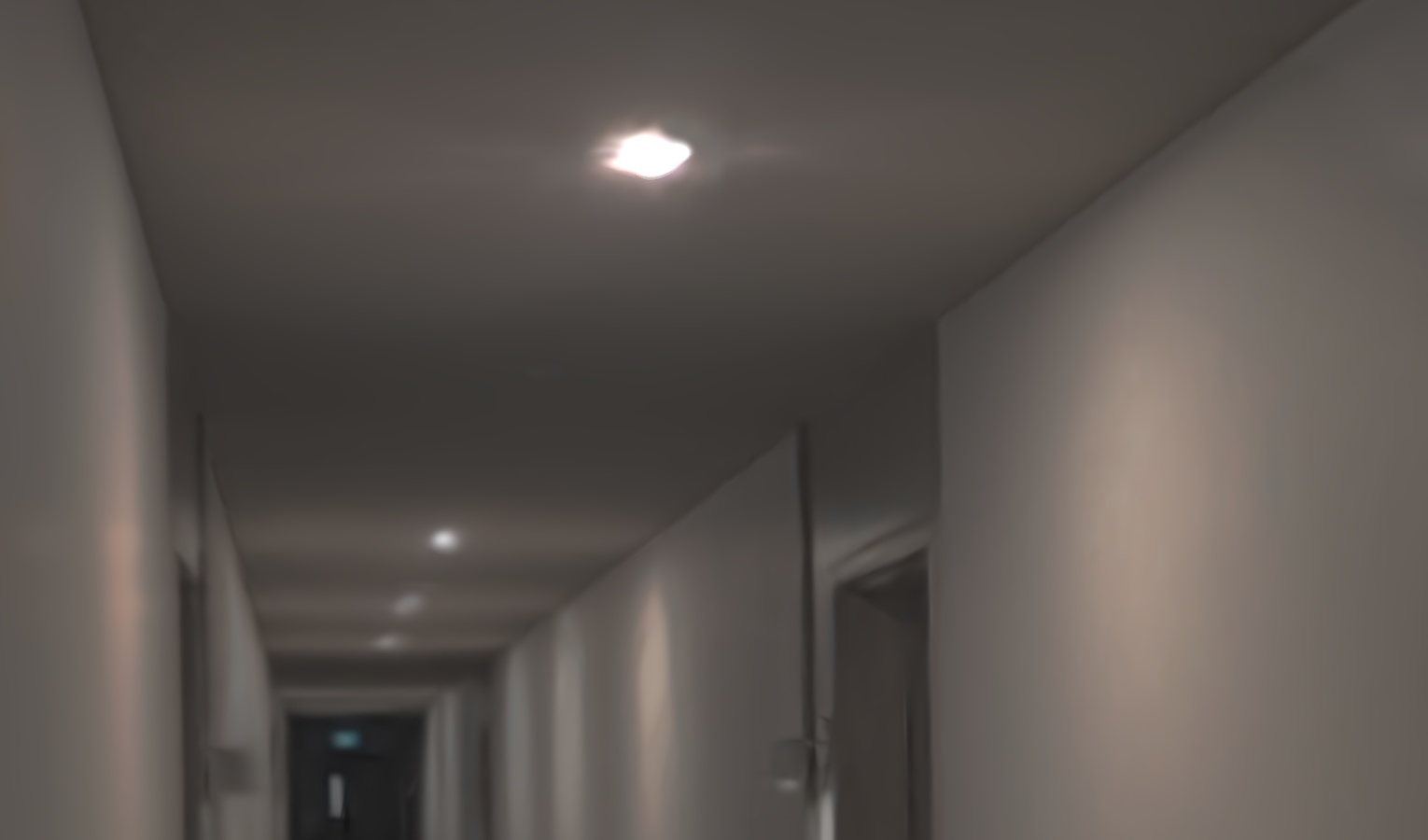} &
    \includegraphics[width=\mywidth]{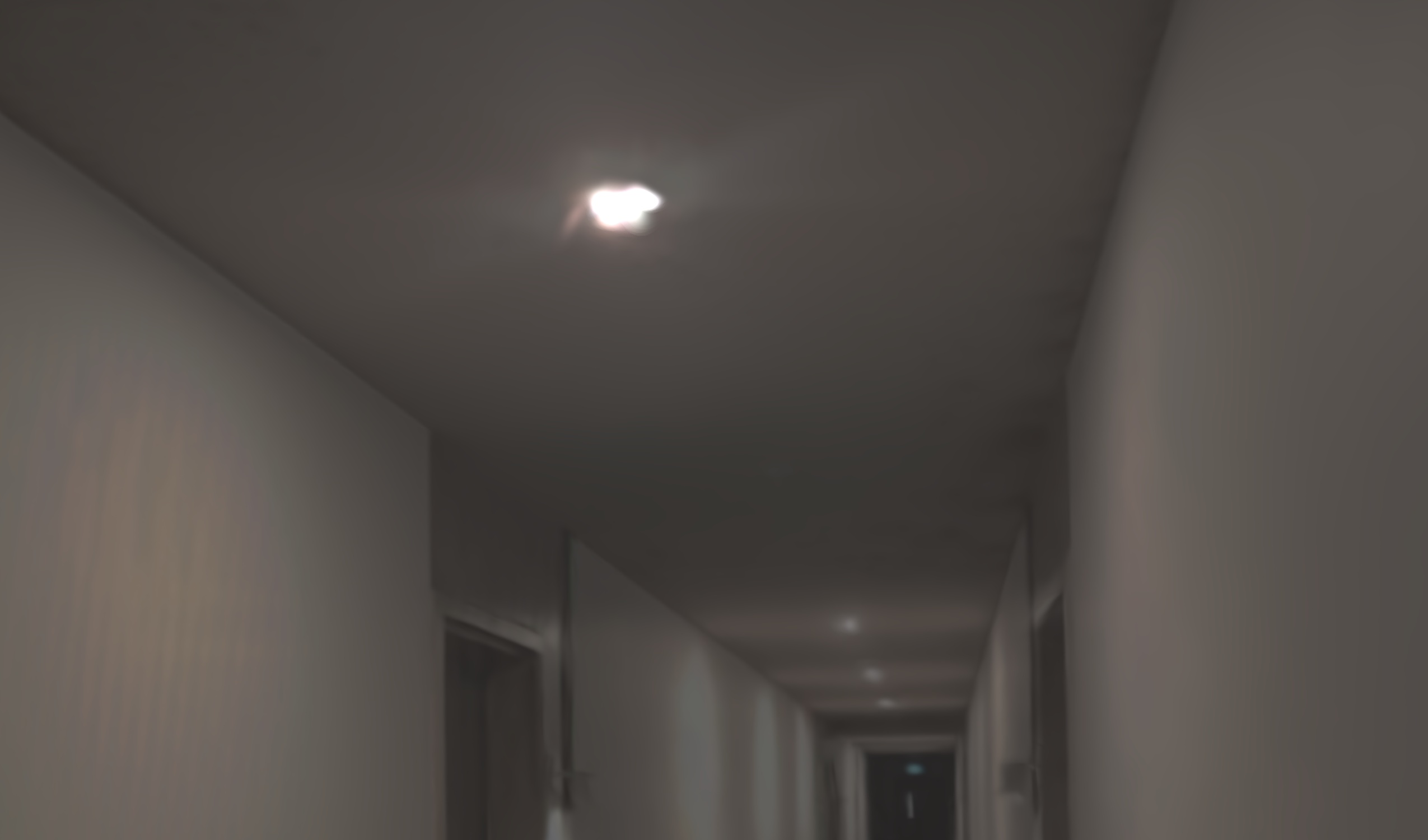} \\
    \end{tabular}
    \caption{\modif{HDR predictions (top row) are manually clamped to simulate inconsistent HDR predictions (middle row). The spherical harmonics RGB representation in the GS reconstruction (bottom row) smooths the values and improves multi-view consistency. Images are displayed in linear EV-3 for clarity.}}
    \label{fig:consistent-gs}
\end{figure}

\paragraph{HDR predictions consistency}
\modif{During HDR extrapolation, we process each image independently, which can lead to inconsistent predictions for the same light source. However, we found that the low frequencies of the spherical harmonics act as a form of regularization, ensuring consistency. As shown in \cref{fig:consistent-gs}, even in extreme cases where HDR values were manually clamped to $1$, multi-view consistency is maintained after reconstructing the GS scene. } 

\paragraph{Using a different HDR estimation approach}
\modif{To validate our Gaussian Splatting approach and HDR estimation method compared to PanoHDR-NeRF~\cite{gera2022panohdrnerf}, we replace our light intensity estimation network with LaNet~\cite{yu2021luminance_lanet}. \Cref{tab:lanetours} shows a consistent decrease in quality with this change. }

\begin{table}
\centering
\footnotesize
\begin{tabular}{lccc}
\toprule
& \multicolumn{2}{c}{Relight} & \multicolumn{1}{c}{HDR} \\
\midrule
 Scene      & MSE$_\downarrow$ & PSNR$_\uparrow$ & PU21-PSNR$_\uparrow$ \\
 \midrule
 PanoHDR-NeRF     & \num{0.0134} & \num{18.97} & \num{27.61} \\
 Ours+LaNet & \num{0.0055} & \num{23.00} & \num{28.06} \\
 Ours  & \cellcolor{red!25}\num{0.0026} & \cellcolor{red!25}\num{30.50} & \cellcolor{red!25}\num{30.29}  \\
 \bottomrule
\end{tabular}
\caption{\modif{Quantitative comparaison between PanoHDR-NeRF, our method with LaNet~\cite{yu2021luminance_lanet} as HDR estimator, and our full method.}}
\label{tab:lanetours}
\end{table}

\section{Applications}
\label{sec:applications}

\subsection{Virtual object insertion}
The 3DGS representation allows for the direct composition of virtual objects, as shown in \cref{fig:teaser,fig:object_insertion}. We achieve these results by importing our 3DGS into Blender \cite{Blender} with the Cycles renderer. \modif{We set Gaussians with intensity $>\!1$ as light emitters, and the rest as diffuse surfaces.} This yields images with cast shadows and near-field lighting effects. 

{\renewcommand{\arraystretch}{0.}
\begin{figure}
    \centering
    \footnotesize
    \newcommand{\figsevensize}{0.40}
    \begin{tabular}{@{}c@{}c@{}}
        \includegraphics[width=\figsevensize\linewidth]{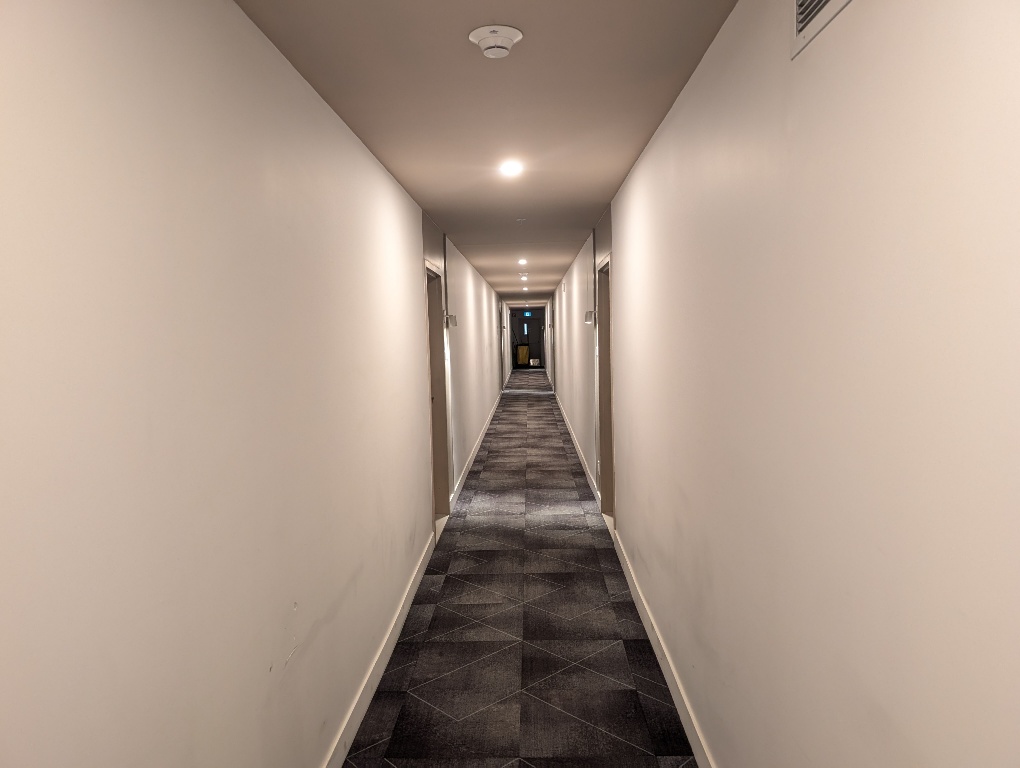} & \includegraphics[width=\figsevensize\linewidth]{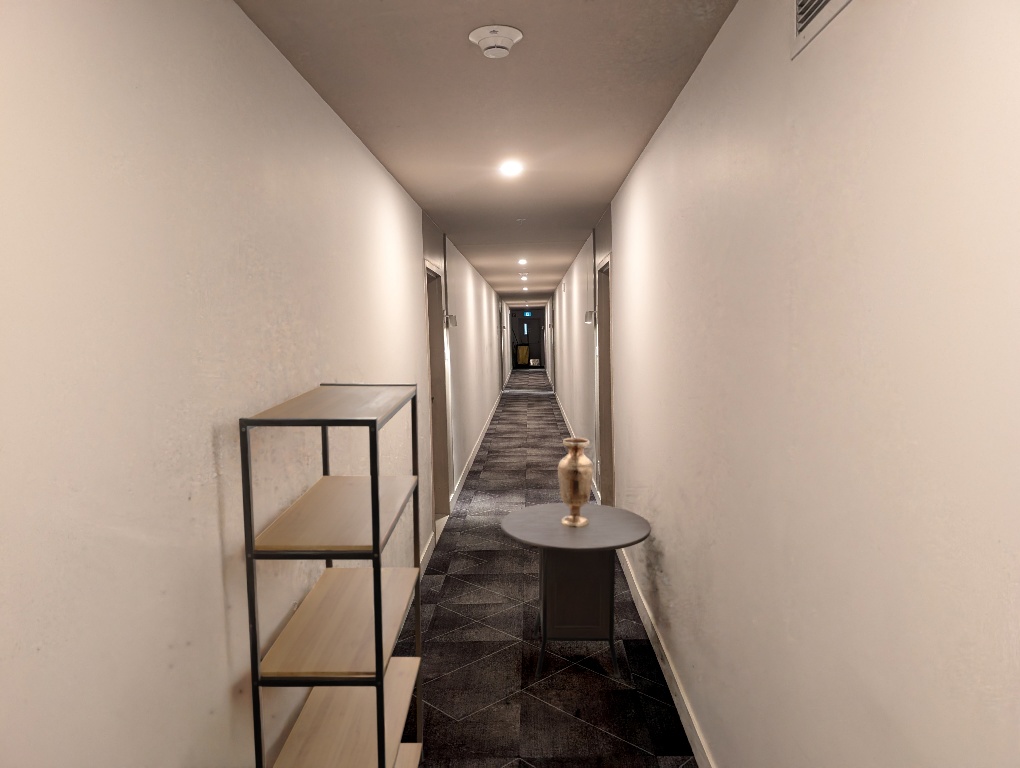} \\
        \includegraphics[width=\figsevensize\linewidth]{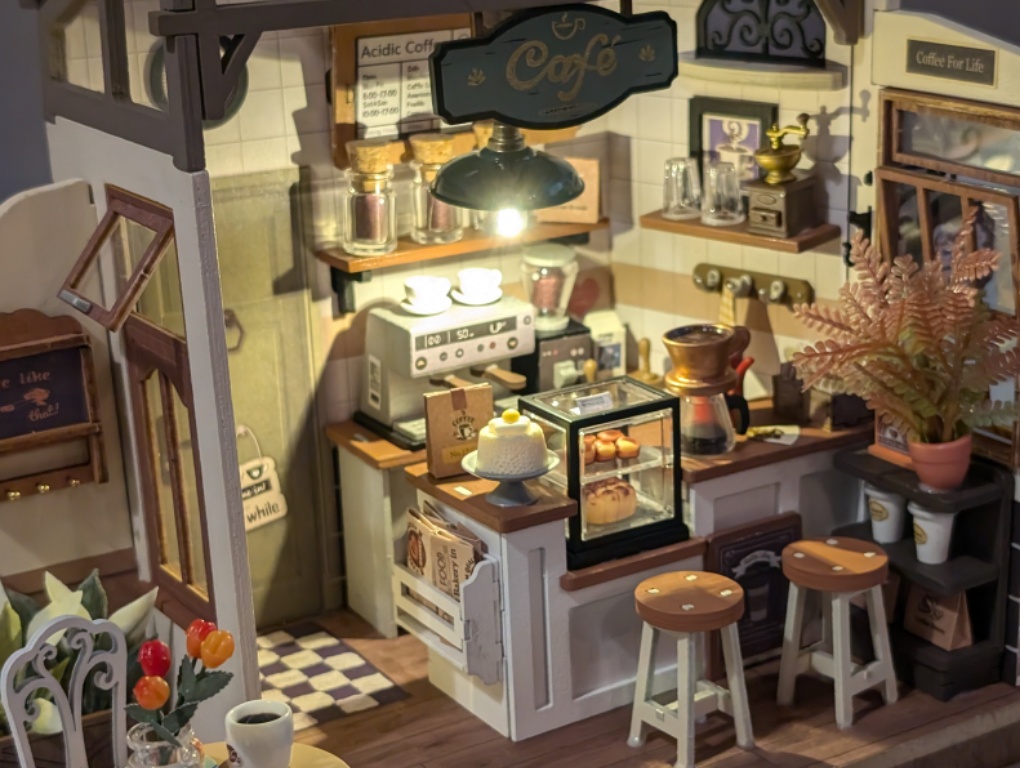} & \includegraphics[width=\figsevensize\linewidth]{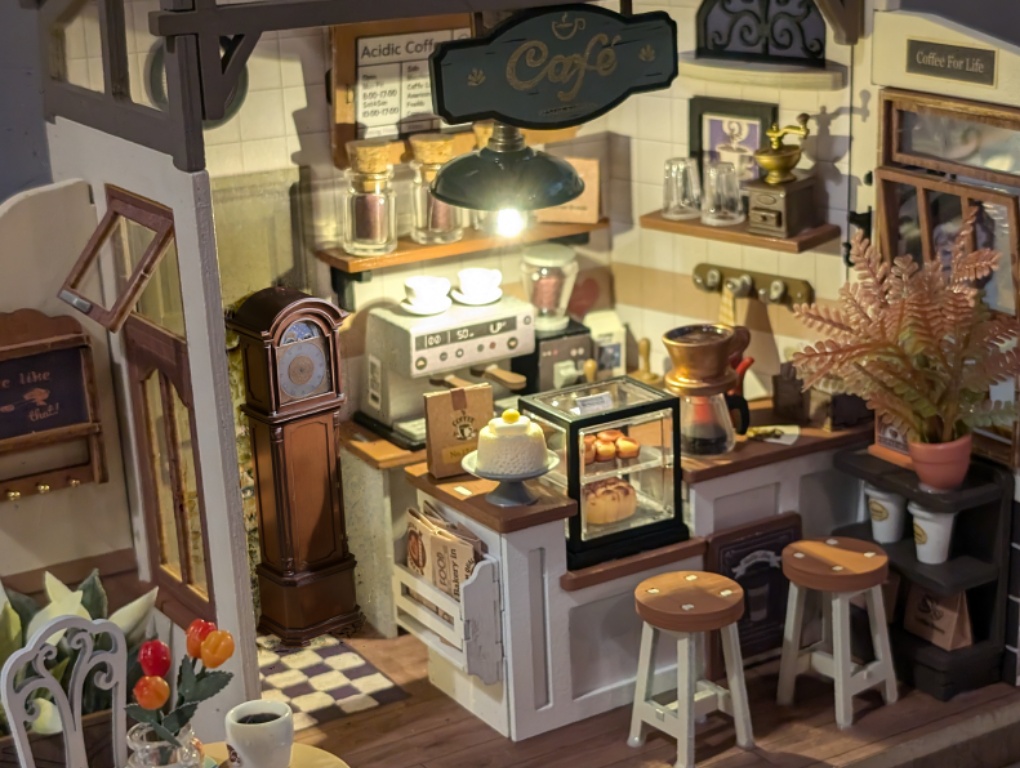} \\
        \includegraphics[width=\figsevensize\linewidth]{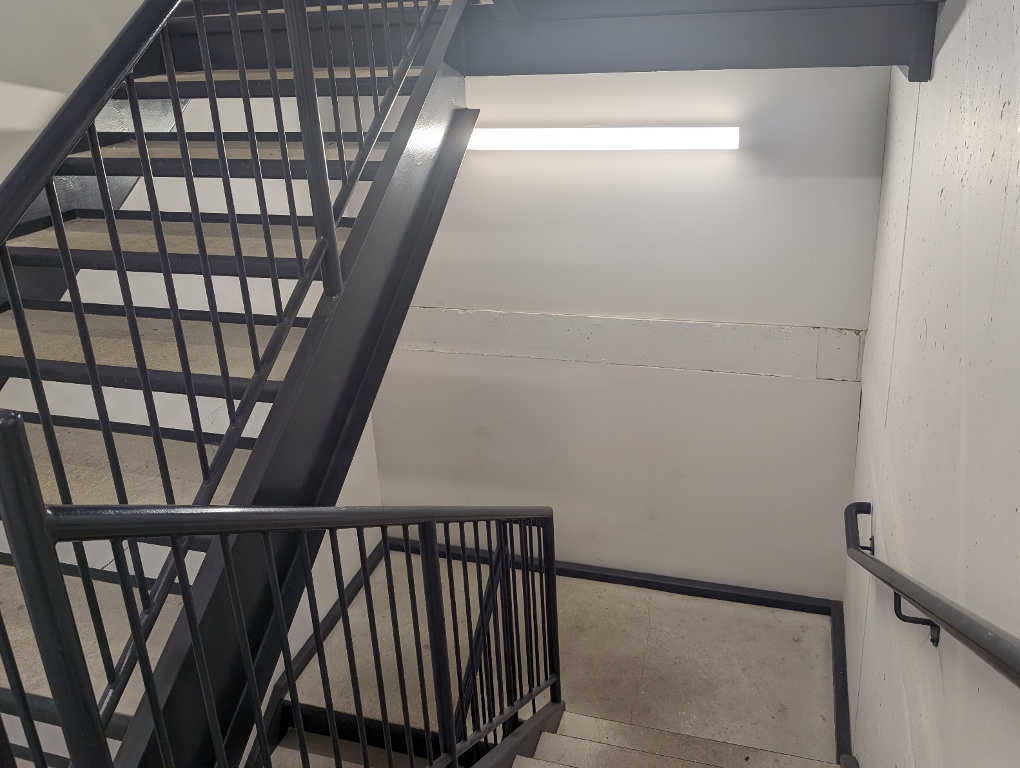} & \includegraphics[width=\figsevensize\linewidth]{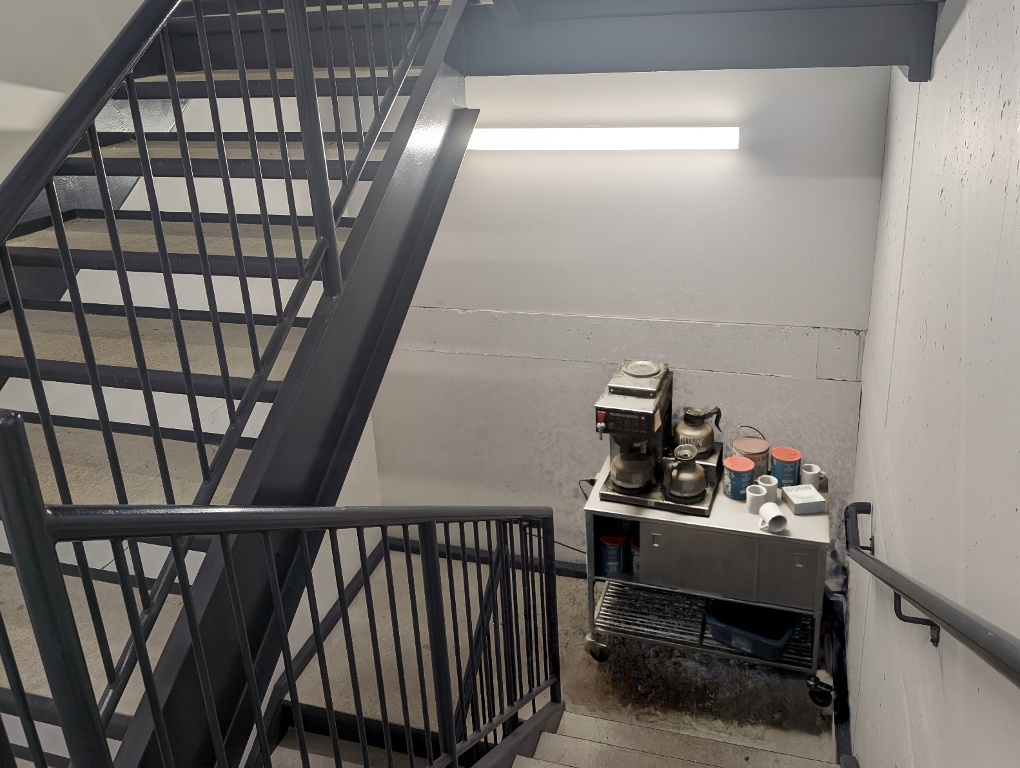} \\
        \includegraphics[width=\figsevensize\linewidth]{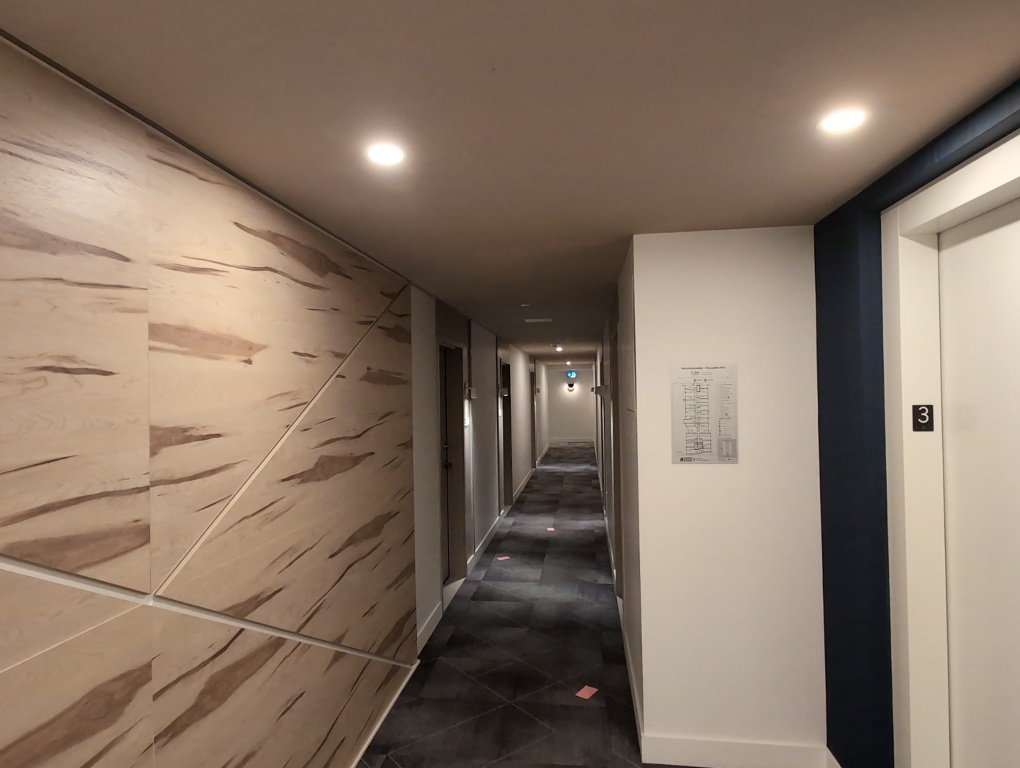} & \includegraphics[width=\figsevensize\linewidth]{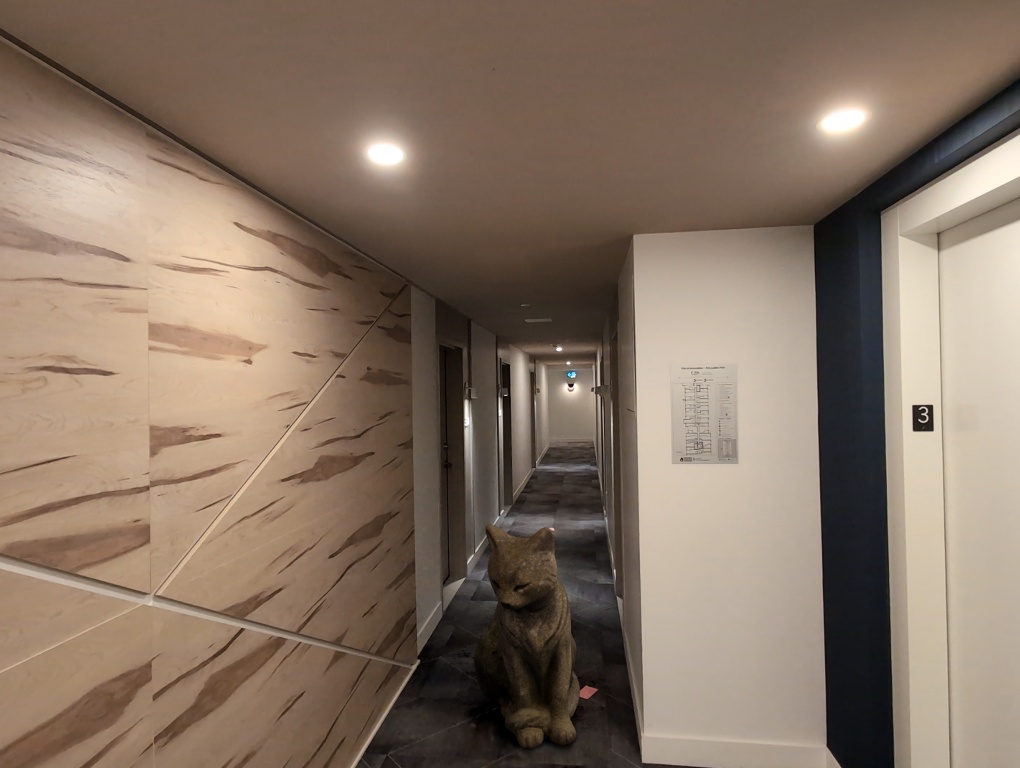} \\
        \vspace{-6pt}
    \end{tabular}
    \caption{\modif{Examples of object insertions using \themethod}}
    \label{fig:object_insertion}
\end{figure}
}


\subsection{Text-to-HDR environment}

\begin{figure}
    \centering
    \scriptsize
    \setlength{\tabcolsep}{1pt}
    \setlength{\mywidth}{0.325\linewidth}
    \begin{tabular}{ccc}
        \includegraphics[width=\mywidth]{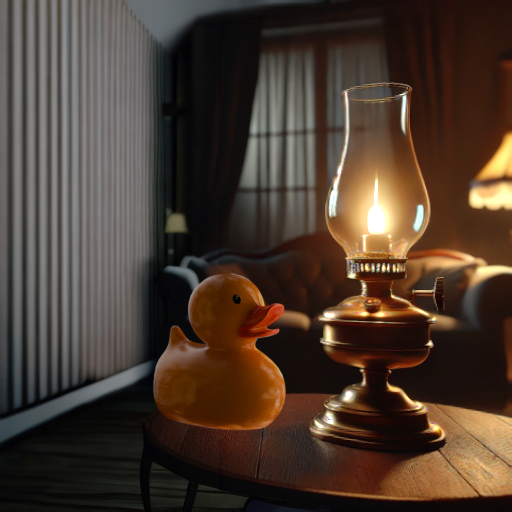} & \includegraphics[width=\mywidth]{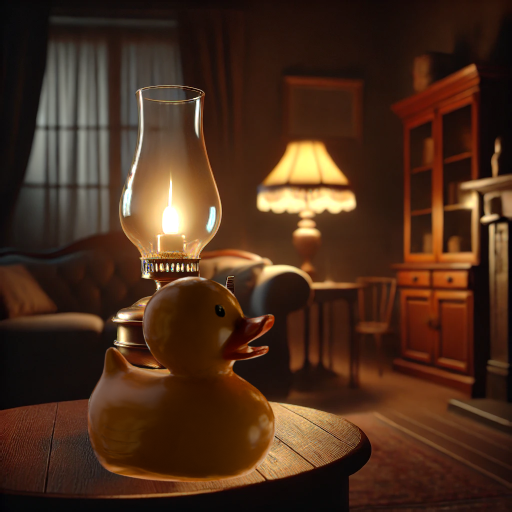} & \includegraphics[width=\mywidth]{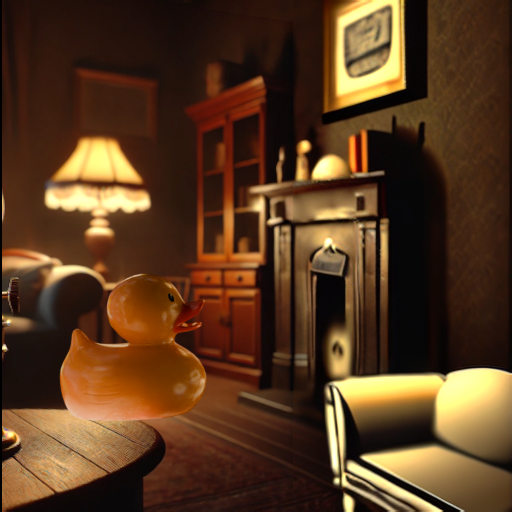} \\
        \multicolumn{3}{l}{\emph{Prompt: ``A gas lamp standing on a table in the center of a dark living room''}} \\*[0.5em]
        \includegraphics[width=\mywidth]{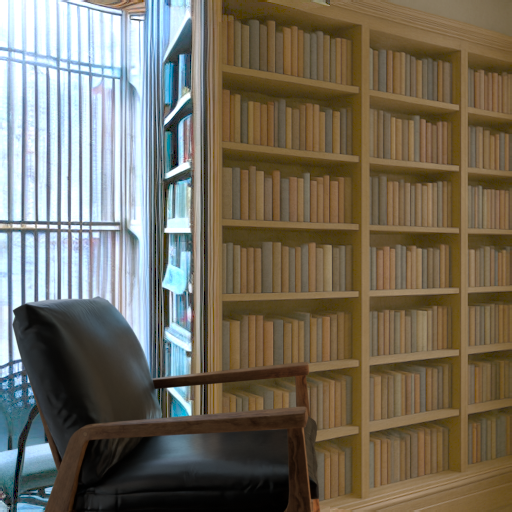} & \includegraphics[width=\mywidth]{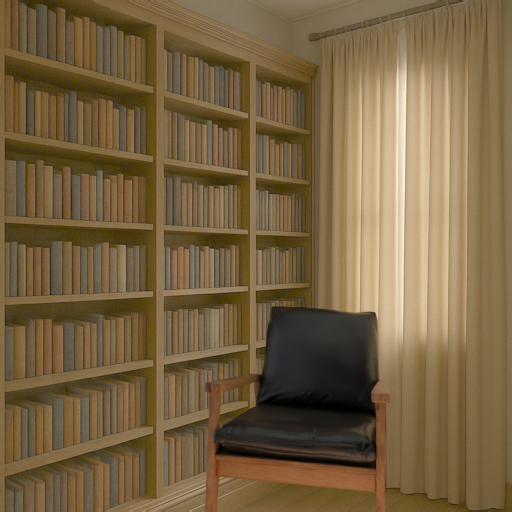} & \includegraphics[width=\mywidth]{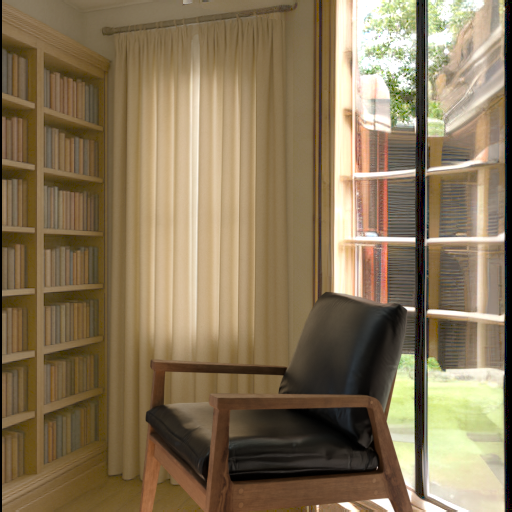} \\
        \multicolumn{3}{l}{\emph{Prompt: ``A library with a window on one side''}} \\
    \end{tabular}
    \caption{\modif{Our method can easily be combined with recent prompt-based multi-view generative methods. Here we extend the text-to-scene LucidDreamer \cite{chung2023luciddreamer} method to \themethod, resulting in text-to-lighting. 3D virtual objects (top: ducky, bottom: chair) are inserted to show the impact of the estimated HDR lighting.}}
    \label{fig:text_to_light}
\end{figure}

\modif{Instead of reconstructing the 3DGS scene on captured photographs, our method can be combined with recent generative approaches, which produce a 3DGS scene solely from text input. For example, we combine \themethod with LucidDreamer~\cite{chung2023luciddreamer}, creating a spatially-varying HDR lighting representation from text, as shown in \cref{fig:text_to_light}.}


\section{Conclusion}

\paragraph{Limitations} 
Despite encouraging results, \themethod suffers from some limitations. 
First, each Gaussian in the 3DGS representation is effectively considered to be a light source (\eg, when rendering, see \cref{sec:applications}\modif{). This hinders} the simulation of more complex lighting effects, such as interreflections or scattering. Here, it is likely that new raytracing-based approaches~\cite{moenne20243d} could help. 
Second, our HDR light intensity estimation network is not trained to achieve multi-view consistency: it is therefore possible that different content is extrapolated for two nearby views. While the spherical harmonics in the 3DGS representation alleviate this somewhat and can realistically blend inconsistent predictions, future work could enforce such consistency, for example, through score distillation sampling~\cite{poole2022dreamfusion}. We also note that, unlike most lighting estimation methods, which predict light sources outside the field of view of the image, we only recover the lighting for visible parts of the scene. Combining our method with view extrapolation approaches (\eg, \cite{chung2023luciddreamer,paliwal2024panodreamer}) is a promising future direction. 
Additionally, although our method can produce HDR environment maps for fast rendering, accurate near-field lighting effects require ray tracing due to the high density of Gaussians generated, which is impractical for real-time use. 
Finally, our current implementation, which relies on InstantSplat~\cite{fan2024instantsplat}, does not scale well to large scenes. This is somewhat orthogonal to our goal of capturing lighting---we are hopeful that new developments (\eg, \cite{kerbl2024hierarchical}) could help alleviate this while being compatible with our method. 

We present \themethod{}, a novel Gaussian Splatting-based HDR Lighting representation. 
Our method employs a diffusion-based LDR-to-HDR model that enhances the dynamic range of LDR images to an unsaturated HDR, even in the presence of very bright light sources. We then fit a 3DGS model on the predicted HDR images, resulting in a lighting representation producing spatially-varying effects, high-frequency reflections, and hard cast shadows, which can be easily integrated into existing rendering engines.


\noindent \textbf{Acknowledgements} 
This work was supported by Sentinel North, NSERC grants RGPIN 2020-04799 and ALLRP 557208-20, and the Digital Research Alliance Canada.

%

%
\clearpage
{
    \small
    \bibliographystyle{ieeenat_fullname}
    \bibliography{main}

\begin{thebibliography}{72}
\providecommand{\natexlab}[1]{#1}
\providecommand{\url}[1]{\texttt{#1}}
\expandafter\ifx\csname urlstyle\endcsname\relax
  \providecommand{\doi}[1]{doi: #1}\else
  \providecommand{\doi}{doi: \begingroup \urlstyle{rm}\Url}\fi

\bibitem[Banterle et~al.(2006)Banterle, Ledda, Debattista, and Chalmers]{banterle2006inverse}
Francesco Banterle, Patrick Ledda, Kurt Debattista, and Alan Chalmers.
\newblock Inverse tone mapping.
\newblock In \emph{Int. Conf. Comp. Graph. Int. Tech. in Australasia and Southeast Asia}, 2006.

\bibitem[Bemana et~al.(2024)Bemana, Leimk{\"u}hler, Myszkowski, Seidel, and Ritschel]{bemana2024exposurediffusion}
Mojtaba Bemana, Thomas Leimk{\"u}hler, Karol Myszkowski, Hans-Peter Seidel, and Tobias Ritschel.
\newblock Exposure diffusion: Hdr image generation by consistent ldr denoising.
\newblock \emph{arXiv preprint arXiv:2405.14304}, 2024.

\bibitem[Blinn and Newell(1976)]{blinn1976texture}
James~F Blinn and Martin~E Newell.
\newblock Texture and reflection in computer generated images.
\newblock \emph{Comm. ACM}, 19\penalty0 (10):\penalty0 542--547, 1976.

\bibitem[Bolduc et~al.(2023)Bolduc, Giroux, H{\'e}bert, Demers, and Lalonde]{bolduc2023beyond}
Christophe Bolduc, Justine Giroux, Marc H{\'e}bert, Claude Demers, and Jean-Fran{\c{c}}ois Lalonde.
\newblock Beyond the pixel: a photometrically calibrated {HDR} dataset for luminance and color prediction.
\newblock In \emph{IEEE/CVF Int. Conf. Comput. Vis.}, 2023.

\bibitem[Boss et~al.(2021)Boss, Jampani, Braun, Liu, Barron, and Lensch]{boss2021neuralpil}
Mark Boss, Varun Jampani, Raphael Braun, Ce Liu, Jonathan~T. Barron, and Hendrik~P.A. Lensch.
\newblock Neural-pil: Neural pre-integrated lighting for reflectance decomposition.
\newblock In \emph{Adv. Neural Inform. Process. Syst.}, 2021.

\bibitem[Boss et~al.(2025)Boss, Huang, Vasishta, and Jampani]{sf3d2024}
Mark Boss, Zixuan Huang, Aaryaman Vasishta, and Varun Jampani.
\newblock Sf3d: Stable fast 3d mesh reconstruction with uv-unwrapping and illumination disentanglement.
\newblock In \emph{IEEE/CVF Conf. Comput. Vis. Pattern Recog.}, 2025.

\bibitem[Cai et~al.(2025)Cai, Xiao, Liang, Qin, Zhang, Yang, Liu, and Yuille]{cai2025hdrgs}
Yuanhao Cai, Zihao Xiao, Yixun Liang, Minghan Qin, Yulun Zhang, Xiaokang Yang, Yaoyao Liu, and Alan~L Yuille.
\newblock {HDR-GS}: Efficient high dynamic range novel view synthesis at 1000x speed via gaussian splatting.
\newblock In \emph{Adv. Neural Inform. Process. Syst.}, pages 68453--68471, 2025.

\bibitem[Chung et~al.(2023)Chung, Lee, Nam, Lee, and Lee]{chung2023luciddreamer}
Jaeyoung Chung, Suyoung Lee, Hyeongjin Nam, Jaerin Lee, and Kyoung~Mu Lee.
\newblock Luciddreamer: Domain-free generation of {3D} gaussian splatting scenes.
\newblock \emph{arXiv preprint arXiv:2311.13384}, 2023.

\bibitem[Community(2018)]{Blender}
Blender~Online Community.
\newblock \emph{Blender - a 3D modelling and rendering package}.
\newblock Blender Foundation, Stichting Blender Foundation, Amsterdam, 2018.

\bibitem[Dastjerdi et~al.(2023)Dastjerdi, Eisenmann, Hold-Geoffroy, and Lalonde]{dastjerdi2023everlight}
Mohammad Reza~Karimi Dastjerdi, Jonathan Eisenmann, Yannick Hold-Geoffroy, and Jean-Fran{\c{c}}ois Lalonde.
\newblock Everlight: Indoor-outdoor editable {HDR} lighting estimation.
\newblock In \emph{IEEE/CVF Int. Conf. Comput. Vis.}, 2023.

\bibitem[Debevec and Malik(1997)]{debevec1997recovering}
Paul~E. Debevec and Jitendra Malik.
\newblock Recovering high dynamic range radiance maps from photographs.
\newblock In \emph{ACM SIGGRAPH Conf.}, 1997.

\bibitem[Eilertsen et~al.(2017)Eilertsen, Kronander, Denes, Mantiuk, and Unger]{eilertsen2017hdrcnn}
Gabriel Eilertsen, Joel Kronander, Gyorgy Denes, Rafał Mantiuk, and Jonas Unger.
\newblock Hdr image reconstruction from a single exposure using deep cnns.
\newblock \emph{ACM Trans. Graph.}, 36\penalty0 (6), 2017.

\bibitem[Endo et~al.(2017)Endo, Kanamori, and Mitani]{endo2017drtmo}
Yuki Endo, Yoshihiro Kanamori, and Jun Mitani.
\newblock Deep reverse tone mapping.
\newblock \emph{ACM Trans. Graph.}, 36\penalty0 (6):\penalty0 1--10, 2017.

\bibitem[Fan et~al.(2024)Fan, Cong, Wen, Wang, Zhang, Ding, Xu, Ivanovic, Pavone, Pavlakos, Wang, and Wang]{fan2024instantsplat}
Zhiwen Fan, Wenyan Cong, Kairun Wen, Kevin Wang, Jian Zhang, Xinghao Ding, Danfei Xu, Boris Ivanovic, Marco Pavone, Georgios Pavlakos, Zhangyang Wang, and Yue Wang.
\newblock Instantsplat: Unbounded sparse-view pose-free gaussian splatting in 40 seconds, 2024.

\bibitem[Foley(1996)]{foley1996computer}
James~D Foley.
\newblock \emph{Computer graphics: principles and practice}.
\newblock Addison-Wesley Professional, 1996.

\bibitem[Gardner et~al.(2017)Gardner, Sunkavalli, Yumer, Shen, Gambaretto, Gagn{\'e}, and Lalonde]{gardner2017learning}
Marc-Andr{\'e} Gardner, Kalyan Sunkavalli, Ersin Yumer, Xiaohui Shen, Emiliano Gambaretto, Christian Gagn{\'e}, and Jean-Fran{\c{c}}ois Lalonde.
\newblock Learning to predict indoor illumination from a single image.
\newblock \emph{ACM Trans. Graph.}, 2017.

\bibitem[Gardner et~al.(2019)Gardner, Hold-Geoffroy, Sunkavalli, Gagn{\'e}, and Lalonde]{gardner2019deepparametric}
Marc-Andr{\'e} Gardner, Yannick Hold-Geoffroy, Kalyan Sunkavalli, Christian Gagn{\'e}, and Jean-Fran{\c{c}}ois Lalonde.
\newblock Deep parametric indoor lighting estimation.
\newblock In \emph{IEEE/CVF Int. Conf. Comput. Vis.}, 2019.

\bibitem[Garon et~al.(2019)Garon, Sunkavalli, Hadap, Carr, and Lalonde]{garon2019fast}
Mathieu Garon, Kalyan Sunkavalli, Sunil Hadap, Nathan Carr, and Jean-Fran{\c{c}}ois Lalonde.
\newblock Fast spatially-varying indoor lighting estimation.
\newblock In \emph{IEEE/CVF Conf. Comput. Vis. Pattern Recog.}, 2019.

\bibitem[Gera et~al.(2022)Gera, Dastjerdi, Renaud, Narayanan, and Lalonde]{gera2022panohdrnerf}
Pulkit Gera, Mohammad Reza~Karimi Dastjerdi, Charles Renaud, PJ Narayanan, and Jean-Fran{\c{c}}ois Lalonde.
\newblock Casual indoor hdr radiance capture from omnidirectional images.
\newblock \emph{Brit. Mach. Vis. Conf.}, 33, 2022.

\bibitem[Gharbi et~al.(2017)Gharbi, Chen, Barron, Hasinoff, and Durand]{gharbi2017deep}
Micha{\"e}l Gharbi, Jiawen Chen, Jonathan~T Barron, Samuel~W Hasinoff, and Fr{\'e}do Durand.
\newblock Deep bilateral learning for real-time image enhancement.
\newblock \emph{ACM Trans. Graph.}, 36\penalty0 (4):\penalty0 1--12, 2017.

\bibitem[Green(2003)]{green2003spherical}
Robin Green.
\newblock Spherical harmonic lighting: The gritty details.
\newblock In \emph{Archives of the game developers conference}, page~4, 2003.

\bibitem[Griffiths et~al.(2022)Griffiths, Ritschel, and Philip]{griffiths2022outcast}
David Griffiths, Tobias Ritschel, and Julien Philip.
\newblock Outcast: Outdoor single-image relighting with cast shadows.
\newblock In \emph{Comput. Graph. Forum}, pages 179--193. Wiley Online Library, 2022.

\bibitem[Hanji et~al.(2022)Hanji, Mantiuk, Eilertsen, Hajisharif, and Unger]{hanji2022comparison}
Param Hanji, Rafal Mantiuk, Gabriel Eilertsen, Saghi Hajisharif, and Jonas Unger.
\newblock Comparison of single image {HDR} reconstruction methods—the caveats of quality assessment.
\newblock In \emph{ACM SIGGRAPH Conf.}, 2022.

\bibitem[Hasinoff et~al.(2016)Hasinoff, Sharlet, Geiss, Adams, Barron, Kainz, Chen, and Levoy]{hasinoff2016burst}
Samuel~W Hasinoff, Dillon Sharlet, Ryan Geiss, Andrew Adams, Jonathan~T Barron, Florian Kainz, Jiawen Chen, and Marc Levoy.
\newblock Burst photography for high dynamic range and low-light imaging on mobile cameras.
\newblock \emph{ACM Trans. Graph.}, 35\penalty0 (6):\penalty0 1--12, 2016.

\bibitem[Hayden(2002)]{hayden2002production}
Landis Hayden.
\newblock Production-ready global illumination.
\newblock In \emph{ACM SIGGRAPH Conf.}, pages 93--95, 2002.

\bibitem[Hold-Geoffroy et~al.(2019)Hold-Geoffroy, Athawale, and Lalonde]{hold2019deep}
Yannick Hold-Geoffroy, Akshaya Athawale, and Jean-Fran{\c{c}}ois Lalonde.
\newblock Deep sky modeling for single image outdoor lighting estimation.
\newblock In \emph{IEEE/CVF Conf. Comput. Vis. Pattern Recog.}, 2019.

\bibitem[Huang et~al.(2022)Huang, Zhang, Feng, Li, Wang, and Wang]{huang2022hdr}
Xin Huang, Qi Zhang, Ying Feng, Hongdong Li, Xuan Wang, and Qing Wang.
\newblock {HDR-NeRF}: High dynamic range neural radiance fields.
\newblock In \emph{IEEE/CVF Conf. Comput. Vis. Pattern Recog.}, 2022.

\bibitem[Jun-Seong et~al.(2022)Jun-Seong, Yu-Ji, Ye-Bin, and Oh]{jun2022hdrplenoxels}
Kim Jun-Seong, Kim Yu-Ji, Moon Ye-Bin, and Tae-Hyun Oh.
\newblock {HDR}-plenoxels: Self-calibrating high dynamic range radiance fields.
\newblock In \emph{Eur. Conf. Comput. Vis.}, 2022.

\bibitem[Kang~Du and Wang(2025)]{du2024gsidilluminationdecompositiongaussian}
Zhihao~Liang Kang~Du and Zeyu Wang.
\newblock {GS-ID}: Illumination decomposition on gaussian splatting via diffusion prior and parametric light source optimization.
\newblock In \emph{IEEE/CVF Int. Conf. Comput. Vis.}, 2025.

\bibitem[Kerbl et~al.(2023)Kerbl, Kopanas, Leimk{\"u}hler, and Drettakis]{kerbl2023gaussiansplatting}
Bernhard Kerbl, Georgios Kopanas, Thomas Leimk{\"u}hler, and George Drettakis.
\newblock 3d gaussian splatting for real-time radiance field rendering.
\newblock \emph{ACM Trans. Graph.}, 42\penalty0 (4):\penalty0 139--1, 2023.

\bibitem[Kerbl et~al.(2024)Kerbl, Meuleman, Kopanas, Wimmer, Lanvin, and Drettakis]{kerbl2024hierarchical}
Bernhard Kerbl, Andreas Meuleman, Georgios Kopanas, Michael Wimmer, Alexandre Lanvin, and George Drettakis.
\newblock A hierarchical {3D} gaussian representation for real-time rendering of very large datasets.
\newblock \emph{ACM Trans. Graph.}, 43\penalty0 (4):\penalty0 1--15, 2024.

\bibitem[Li et~al.(2020)Li, Shafiei, Ramamoorthi, Sunkavalli, and Chandraker]{li2020inverse}
Zhengqin Li, Mohammad Shafiei, Ravi Ramamoorthi, Kalyan Sunkavalli, and Manmohan Chandraker.
\newblock Inverse rendering for complex indoor scenes: Shape, spatially-varying lighting and svbrdf from a single image.
\newblock In \emph{IEEE/CVF Conf. Comput. Vis. Pattern Recog.}, 2020.

\bibitem[Li et~al.(2023)Li, Yu, Okunev, Chandraker, and Dong]{li2023spatiotemporally}
Zhengqin Li, Li Yu, Mikhail Okunev, Manmohan Chandraker, and Zhao Dong.
\newblock Spatiotemporally consistent {HDR} indoor lighting estimation.
\newblock \emph{ACM Trans. Graph.}, 42\penalty0 (3):\penalty0 1--15, 2023.

\bibitem[Ling et~al.(2024)Ling, Yu, Xu, Du, and Zhao]{ling2024nerf}
Jingwang Ling, Ruihan Yu, Feng Xu, Chun Du, and Shuang Zhao.
\newblock Nerf as a non-distant environment emitter in physics-based inverse rendering.
\newblock In \emph{ACM SIGGRAPH Conf.}, pages 1--12, 2024.

\bibitem[Liu et~al.(2023)Liu, Li, Wu, and Lee]{liu2023llava}
Haotian Liu, Chunyuan Li, Qingyang Wu, and Yong~Jae Lee.
\newblock Visual instruction tuning.
\newblock In \emph{Adv. Neural Inform. Process. Syst.}, 2023.

\bibitem[Liu et~al.(2020)Liu, Lai, Chen, Kao, Yang, Chuang, and Huang]{liu2020singlehdr}
Yu-Lun Liu, Wei-Sheng Lai, Yu-Sheng Chen, Yi-Lung Kao, Ming-Hsuan Yang, Yung-Yu Chuang, and Jia-Bin Huang.
\newblock Single-image hdr reconstruction by learning to reverse the camera pipeline.
\newblock In \emph{IEEE/CVF Conf. Comput. Vis. Pattern Recog.}, 2020.

\bibitem[Mantiuk and Azimi(2021)]{mantiuk2021pu}
Rafał~K. Mantiuk and Maryam Azimi.
\newblock Pu21: A novel perceptually uniform encoding for adapting existing quality metrics for {HDR}.
\newblock In \emph{Picture Coding Symp.}, 2021.

\bibitem[Mantiuk et~al.(2023)Mantiuk, Hammou, and Hanji]{mantiuk2023hdr}
Rafal~K Mantiuk, Dounia Hammou, and Param Hanji.
\newblock {HDR-VDP-3}: A multi-metric for predicting image differences, quality and contrast distortions in high dynamic range and regular content.
\newblock \emph{arXiv preprint arXiv:2304.13625}, 2023.

\bibitem[Marnerides et~al.(2018)Marnerides, Bashford-Rogers, Hatchett, and Debattista]{marnerides2018expandnet}
Demetris Marnerides, Thomas Bashford-Rogers, Jonathan Hatchett, and Kurt Debattista.
\newblock Expandnet: A deep convolutional neural network for high dynamic range expansion from low dynamic range content.
\newblock In \emph{Comput. Graph. Forum}, pages 37--49. Wiley Online Library, 2018.

\bibitem[Mildenhall et~al.(2021)Mildenhall, Srinivasan, Tancik, Barron, Ramamoorthi, and Ng]{mildenhall2021nerf}
Ben Mildenhall, Pratul~P Srinivasan, Matthew Tancik, Jonathan~T Barron, Ravi Ramamoorthi, and Ren Ng.
\newblock Nerf: Representing scenes as neural radiance fields for view synthesis.
\newblock \emph{Comm. ACM}, 65\penalty0 (1):\penalty0 99--106, 2021.

\bibitem[Mildenhall et~al.(2022)Mildenhall, Hedman, Martin-Brualla, Srinivasan, and Barron]{mildenhall2022nerf}
Ben Mildenhall, Peter Hedman, Ricardo Martin-Brualla, Pratul~P Srinivasan, and Jonathan~T Barron.
\newblock Nerf in the dark: High dynamic range view synthesis from noisy raw images.
\newblock In \emph{IEEE/CVF Conf. Comput. Vis. Pattern Recog.}, 2022.

\bibitem[Moenne-Loccoz et~al.(2024)Moenne-Loccoz, Mirzaei, Perel, de~Lutio, Martinez~Esturo, State, Fidler, Sharp, and Gojcic]{moenne20243d}
Nicolas Moenne-Loccoz, Ashkan Mirzaei, Or Perel, Riccardo de Lutio, Janick Martinez~Esturo, Gavriel State, Sanja Fidler, Nicholas Sharp, and Zan Gojcic.
\newblock 3d gaussian ray tracing: Fast tracing of particle scenes.
\newblock \emph{ACM Trans. Graph.}, 43\penalty0 (6):\penalty0 1--19, 2024.

\bibitem[Nayar and Mitsunaga(2000)]{nayar2000high}
Shree~K Nayar and Tomoo Mitsunaga.
\newblock High dynamic range imaging: Spatially varying pixel exposures.
\newblock In \emph{IEEE/CVF Conf. Comput. Vis. Pattern Recog.}, 2000.

\bibitem[{New House Internet Services B.V., Rotterdam, The Netherlands}()]{ptgui}
{New House Internet Services B.V., Rotterdam, The Netherlands}.
\newblock Ptgui.

\bibitem[Paliwal et~al.(2024)Paliwal, Zhou, Tsarov, and Kalantari]{paliwal2024panodreamer}
Avinash Paliwal, Xilong Zhou, Andrii Tsarov, and Nima~Khademi Kalantari.
\newblock Panodreamer: {3D} panorama synthesis from a single image.
\newblock \emph{arXiv preprint arXiv:2412.04827}, 2024.

\bibitem[Phongthawee et~al.(2024)Phongthawee, Chinchuthakun, Sinsunthithet, Jampani, Raj, Khungurn, and Suwajanakorn]{phongthawee2024diffusionlight}
Pakkapon Phongthawee, Worameth Chinchuthakun, Nontaphat Sinsunthithet, Varun Jampani, Amit Raj, Pramook Khungurn, and Supasorn Suwajanakorn.
\newblock Diffusionlight: Light probes for free by painting a chrome ball.
\newblock In \emph{IEEE/CVF Conf. Comput. Vis. Pattern Recog.}, 2024.

\bibitem[PolyHaven(2025)]{polyhaven}
PolyHaven.
\newblock Poly haven: The public 3d asset library.
\newblock 2025.

\bibitem[Poole et~al.(2022)Poole, Jain, Barron, and Mildenhall]{poole2022dreamfusion}
Ben Poole, Ajay Jain, Jonathan~T Barron, and Ben Mildenhall.
\newblock Dreamfusion: Text-to-3d using 2d diffusion.
\newblock \emph{arXiv preprint arXiv:2209.14988}, 2022.

\bibitem[Ramamoorthi and Hanrahan(2001)]{ramamoorthi2001efficient}
Ravi Ramamoorthi and Pat Hanrahan.
\newblock An efficient representation for irradiance environment maps.
\newblock In \emph{Proceedings of the 28th annual conference on Computer graphics and interactive techniques}, pages 497--500, 2001.

\bibitem[Reinhard(2020)]{reinhard2020high}
Erik Reinhard.
\newblock High dynamic range imaging.
\newblock In \emph{Computer Vision: A Reference Guide}, pages 1--6. Springer, 2020.

\bibitem[Reinhard et~al.(2002)Reinhard, Stark, Shirley, and Ferwerda]{reinhard2002photographic}
Erik Reinhard, Michael Stark, Peter Shirley, and James Ferwerda.
\newblock Photographic tone reproduction for digital images.
\newblock \emph{ACM Trans. Graph.}, 21\penalty0 (3):\penalty0 267--276, 2002.

\bibitem[Rombach et~al.(2022)Rombach, Blattmann, Lorenz, Esser, and Ommer]{rombach2022high}
Robin Rombach, Andreas Blattmann, Dominik Lorenz, Patrick Esser, and Bj{\"o}rn Ommer.
\newblock High-resolution image synthesis with latent diffusion models.
\newblock In \emph{IEEE/CVF Conf. Comput. Vis. Pattern Recog.}, 2022.

\bibitem[Santos et~al.(2020)Santos, Ren, and Kalantari]{santos2020maskHDR}
Marcel~Santana Santos, Tsang~Ing Ren, and Nima~Khademi Kalantari.
\newblock Single image {HDR} reconstruction using a {CNN} with masked features and perceptual loss.
\newblock \emph{ACM Trans. Graph.}, 39\penalty0 (4), 2020.

\bibitem[Sloan et~al.(2023)Sloan, Kautz, and Snyder]{sloan2023precomputed}
Peter-Pike Sloan, Jan Kautz, and John Snyder.
\newblock Precomputed radiance transfer for real-time rendering in dynamic, low-frequency lighting environments.
\newblock In \emph{Seminal Graphics Papers: Pushing the Boundaries, Volume 2}, pages 339--348. 2023.

\bibitem[Somanath and Kurz(2021)]{somanath2021hdr}
Gowri Somanath and Daniel Kurz.
\newblock {HDR} environment map estimation for real-time augmented reality.
\newblock In \emph{IEEE/CVF Conf. Comput. Vis. Pattern Recog.}, 2021.

\bibitem[Song and Funkhouser(2019)]{song2019neural}
Shuran Song and Thomas Funkhouser.
\newblock Neural illumination: Lighting prediction for indoor environments.
\newblock In \emph{IEEE/CVF Conf. Comput. Vis. Pattern Recog.}, 2019.

\bibitem[Srinivasan et~al.(2020)Srinivasan, Mildenhall, Tancik, Barron, Tucker, and Snavely]{srinivasan2020lighthouse}
Pratul~P Srinivasan, Ben Mildenhall, Matthew Tancik, Jonathan~T Barron, Richard Tucker, and Noah Snavely.
\newblock Lighthouse: Predicting lighting volumes for spatially-coherent illumination.
\newblock In \emph{IEEE/CVF Conf. Comput. Vis. Pattern Recog.}, 2020.

\bibitem[Stumpfel et~al.(2006)Stumpfel, Jones, Wenger, Tchou, Hawkins, and Debevec]{stumpfel2006direct}
Jessi Stumpfel, Andrew Jones, Andreas Wenger, Chris Tchou, Tim Hawkins, and Paul Debevec.
\newblock Direct {HDR} capture of the sun and sky.
\newblock In \emph{ACM SIGGRAPH 2006 Courses}, pages 5--es. 2006.

\bibitem[Wang et~al.(2023{\natexlab{a}})Wang, Serrano, Pan, Chen, Myszkowski, Seidel, Theobalt, and Leimk{\"u}hler]{wang2023glowgan}
Chao Wang, Ana Serrano, Xingang Pan, Bin Chen, Karol Myszkowski, Hans-Peter Seidel, Christian Theobalt, and Thomas Leimk{\"u}hler.
\newblock Glowgan: Unsupervised learning of hdr images from ldr images in the wild.
\newblock In \emph{IEEE/CVF Int. Conf. Comput. Vis.}, 2023{\natexlab{a}}.

\bibitem[Wang et~al.(2024{\natexlab{a}})Wang, Wolski, Kerbl, Serrano, Bemana, Seidel, Myszkowski, and Leimk{\"u}hler]{wang2024cinematic}
Chao Wang, Krzysztof Wolski, Bernhard Kerbl, Ana Serrano, Mojtaba Bemana, Hans-Peter Seidel, Karol Myszkowski, and Thomas Leimk{\"u}hler.
\newblock Cinematic gaussians: Real-time {HDR} radiance fields with depth of field.
\newblock In \emph{Comput. Graph. Forum}, page e15214. Wiley Online Library, 2024{\natexlab{a}}.

\bibitem[Wang et~al.(2024{\natexlab{b}})Wang, Xia, Leimk{\"u}hler, Myszkowski, and Zhang]{wang2024lediff}
Chao Wang, Zhihao Xia, Thomas Leimk{\"u}hler, Karol Myszkowski, and Xuaner Zhang.
\newblock Lediff: Latent exposure diffusion for hdr generation.
\newblock \emph{arXiv preprint arXiv:2412.14456}, 2024{\natexlab{b}}.

\bibitem[Wang et~al.(2024{\natexlab{c}})Wang, Wang, Gong, and Xue]{wang2024bilateral}
Yuehao Wang, Chaoyi Wang, Bingchen Gong, and Tianfan Xue.
\newblock Bilateral guided radiance field processing.
\newblock \emph{ACM Trans. Graph.}, 43\penalty0 (4):\penalty0 1--13, 2024{\natexlab{c}}.

\bibitem[Wang et~al.(2021)Wang, Philion, Fidler, and Kautz]{wang2021learning}
Zian Wang, Jonah Philion, Sanja Fidler, and Jan Kautz.
\newblock Learning indoor inverse rendering with {3D} spatially-varying lighting.
\newblock In \emph{IEEE/CVF Conf. Comput. Vis. Pattern Recog.}, 2021.

\bibitem[Wang et~al.(2023{\natexlab{b}})Wang, Shen, Gao, Huang, Munkberg, Hasselgren, Gojcic, Chen, and Fidler]{wang2023neural}
Zian Wang, Tianchang Shen, Jun Gao, Shengyu Huang, Jacob Munkberg, Jon Hasselgren, Zan Gojcic, Wenzheng Chen, and Sanja Fidler.
\newblock Neural fields meet explicit geometric representations for inverse rendering of urban scenes.
\newblock In \emph{IEEE/CVF Conf. Comput. Vis. Pattern Recog.}, 2023{\natexlab{b}}.

\bibitem[Wu et~al.(2024)Wu, Yi, Fang, Liu, and Wang]{wu2024fast}
Guanjun Wu, Taoran Yi, Jiemin Fang, Wenyu Liu, and Xinggang Wang.
\newblock Fast high dynamic range radiance fields for dynamic scenes.
\newblock In \emph{Int. Conf. 3D Vis.}, 2024.

\bibitem[Wu et~al.(2023)Wu, Hu, Li, Zhang, Fan, and Yu]{Wu_2023_CVPR}
Haoqian Wu, Zhipeng Hu, Lincheng Li, Yongqiang Zhang, Changjie Fan, and Xin Yu.
\newblock Nefii: Inverse rendering for reflectance decomposition with near-field indirect illumination.
\newblock In \emph{IEEE/CVF Conf. Comput. Vis. Pattern Recog.}, 2023.

\bibitem[Yu et~al.(2021)Yu, Liu, Long, Dong, Zou, and Xiao]{yu2021luminance_lanet}
Hanning Yu, Wentao Liu, Chengjiang Long, Bo Dong, Qin Zou, and Chunxia Xiao.
\newblock Luminance attentive networks for hdr image and panorama reconstruction.
\newblock In \emph{Comput. Graph. Forum}, pages 181--192. Wiley Online Library, 2021.

\bibitem[Zeng et~al.(2024)Zeng, Deschaintre, Georgiev, Hold-Geoffroy, Hu, Luan, Yan, and Ha{\v{s}}an]{zeng2024rgb}
Zheng Zeng, Valentin Deschaintre, Iliyan Georgiev, Yannick Hold-Geoffroy, Yiwei Hu, Fujun Luan, Ling-Qi Yan, and Milo{\v{s}} Ha{\v{s}}an.
\newblock {RGB} $\leftrightarrow$ {X}: Image decomposition and synthesis using material-and lighting-aware diffusion models.
\newblock In \emph{ACM SIGGRAPH Conf.}, 2024.

\bibitem[Zhan et~al.(2021)Zhan, Zhang, Yu, Chang, Lu, Ma, and Xie]{zhan2021emlight}
Fangneng Zhan, Changgong Zhang, Yingchen Yu, Yuan Chang, Shijian Lu, Feiying Ma, and Xuansong Xie.
\newblock Emlight: Lighting estimation via spherical distribution approximation.
\newblock In \emph{Assoc. Adv. of Art. Int.}, 2021.

\bibitem[Zhang and Lalonde(2017)]{zhang2017learning}
Jinsong Zhang and Jean-Fran{\c{c}}ois Lalonde.
\newblock Learning high dynamic range from outdoor panoramas.
\newblock In \emph{IEEE/CVF Int. Conf. Comput. Vis.}, 2017.

\bibitem[Zhang et~al.(2021)Zhang, Luan, Wang, Bala, and Snavely]{physg2021}
Kai Zhang, Fujun Luan, Qianqian Wang, Kavita Bala, and Noah Snavely.
\newblock {PhySG}: {I}nverse rendering with spherical gaussians for physics-based material editing and relighting.
\newblock In \emph{IEEE/CVF Conf. Comput. Vis. Pattern Recog.}, 2021.

\bibitem[Zhang et~al.(2025)Zhang, Rao, and Agrawala]{zhang2025scaling}
Lvmin Zhang, Anyi Rao, and Maneesh Agrawala.
\newblock Scaling in-the-wild training for diffusion-based illumination harmonization and editing by imposing consistent light transport.
\newblock In \emph{Int. Conf. Learn. Represent.}, 2025.

\end{thebibliography}
}

\end{document}